\documentclass[twoside,11pt]{article}

\usepackage[abbrvbib, nohyperref, preprint]{jmlr2e}

\usepackage[dvipsnames]{xcolor}
\colorlet{siaminlinkcolor}{green!50!black}
\colorlet{siamexlinkcolor}{red!50!black}
\colorlet{siamreviewcolor}{black!50}
\usepackage[colorlinks,pagebackref,bookmarksdepth=3]{hyperref}
\hypersetup{allcolors=siaminlinkcolor,urlcolor=siamexlinkcolor}
\renewcommand*{\backref}[1]{\ifx#1\relax \else Page #1 \fi}
\renewcommand*{\backrefalt}[4]{%
  \ifcase #1 \footnotesize{(not cited)}%
  \or        \footnotesize{(cited on p.~#2)}%
  \else      \footnotesize{(cited on pp.~#2)}%
  \fi
}
\usepackage{doi} % get doi hyperlinks in bib
\usepackage{amsfonts}
\usepackage{epstopdf}
\usepackage{amsopn}

\usepackage{amssymb} % for end of environment symbols and asymptotics
\usepackage{graphicx}
\usepackage{subcaption} % for multiple subfigures
\usepackage{booktabs} % for tables, if needed
\usepackage{mathtools} % to DeclarePairedDelimiter for ip, abs, norm
\usepackage{bbm} % for indicator function
\usepackage{mathrsfs} % for additional math fonts for letters
\usepackage{stmaryrd} % for ``mapsfrom''
\let\originalleft\left % Fixes \left and \right spacing issues. See discussion at http://tex.stackexchange.com/questions/2607/spacing-around-left-and-right
\let\originalright\right
\renewcommand{\left}{\mathopen{}\mathclose\bgroup\originalleft}
\renewcommand{\right}{\aftergroup\egroup\originalright}

\newcommand{\N}{\mathbb{N}}

\newcommand{\R}{\mathbb{R}}

\newcommand{\cX}{\mathcal{X}}
\newcommand{\cY}{\mathcal{Y}}

\newcommand{\cG}{\mathcal{G}}

\newcommand{\cU}{\mathcal{U}}

\newcommand{\cO}{\mathcal{O}}

\newcommand{\al}{\alpha}

\newcommand{\bbar}[1]{\overline{#1}}

\newcommand{\Kpost}{K^{(N)}} % posterior cov gp
\newcommand{\Fd}{F^{\star}}
\newcommand{\GP}{\mathsf{GP}}

\DeclarePairedDelimiterX{\iptemp}[2]{\langle}{\rangle}{#1, #2}

\DeclarePairedDelimiterX{\normtemp}[1]{\lVert}{\rVert}{#1}
\newcommand{\norm}{\normtemp}
\DeclarePairedDelimiterX{\abstemp}[1]{\lvert}{\rvert}{#1}

\DeclarePairedDelimiterX{\trtemp}[1]{(}{)}{#1}

\newcommand{\defeq}{\coloneqq} % definition equal in math mode
\newcommand{\condbar}{\, \vert \,} % conditioning RVs

\DeclarePairedDelimiterX{\floor}[1]{\lfloor}{\rfloor}{#1} % floor
\DeclarePairedDelimiterX{\ceil}[1]{\lceil}{\rceil}{#1} % ceiling
\newcommand{\tp}{\top} % tranpose
\DeclareMathOperator*{\argmax}{argmax}
\DeclareMathOperator*{\argmin}{argmin}
\def\mmm[#1]{\mathcal{#1}} % placeholder for mathcal
\def\bbb[#1]{\boldsymbol{#1}} % placeholder for bold, ideally replace with \vct and \mtx
\newcommand{\E}{\operatorname{\mathbb{E}}} % expectation
\newcommand{\comp}{\textsf{c}} % complement of set
\newcommand{\Cov}{\operatorname{Cov}} % covariance (operator)
\newcommand{\diid}{\stackrel{\mathrm{i.i.d.}}{\sim}} % distributed iid tilde symbol
\newcommand{\normal}{\mathsf{N}} % normal distribution
\def\qfa{\quad\text{for all}\quad}

\def\qa{\quad\text{and}\quad}

\def\qw{\quad\text{where}\quad}

\newcommand{\sfit}[1]{\textup{\textsf{\small{#1}}}} % item label font
\newcommand{\set}[2]{{\left\{ #1 \,\middle|\, #2 \right\}}}
\newcommand{\slot}{{\,\cdot\,}}

\definecolor{darkred}{rgb}{.7,0,0}
\definecolor{darkblue}{rgb}{0,0,0.7}
\definecolor{darkorange}{rgb}{1,0.40,0}
\definecolor{darkmagenta}{rgb}{0.55,0,0.55}
\newcommand{\nn}[1]{#1} % Remove color
\ifpdf
\DeclareGraphicsExtensions{.eps,.pdf,.png,.jpg}
\else
\DeclareGraphicsExtensions{.eps}
\fi
\graphicspath{ {figures/} }

\usepackage{enumitem}
\setlist[enumerate]{leftmargin=.5in}
\setlist[itemize]{leftmargin=.5in}

\usepackage{lastpage}
\jmlrheading{XX}{2024}{1-\pageref{LastPage}}{1/XX; Revised XX/24}{XX/24}{21-0000}{Oliver R. A. Dunbar, Nicholas H. Nelsen, and Maya Mutic}

\ShortHeadings{Hyperparameter Optimization for Randomized Algorithms}{Dunbar, Nelsen, and Mutic}
\firstpageno{1}

\ifpdf
\hypersetup{
  pdftitle={Hyperparameter Optimization for Randomized Algorithms},
  pdfauthor={O. R. A. Dunbar, N. H. Nelsen, and M. Mutic}
}
\fi

\begin{document}

\title{Hyperparameter Optimization for Randomized Algorithms: A Case Study \nn{on} Random Features}
\author{\name Oliver R.\ A.\ Dunbar \email odunbar@caltech.edu \\
	\addr Division of Geological and Planetary Sciences\\
	California Institute of Technology\\
	Pasadena, CA 91125, USA
	\AND
	\name Nicholas H.\ Nelsen \email nnelsen@caltech.edu \\
	\addr Department of Computing and Mathematical Sciences\\
	California Institute of Technology\\
	Pasadena, CA 91125, USA
        \AND
        \name Maya Mutic \email mmutic@princeton.edu \\
	\addr Andlinger Center for Energy and the Environment\\
	Princeton University\\
	Princeton, NJ 08554, USA
}

\editor{My editor}

\maketitle
\begin{abstract}% MUST BE less than or equal to 200 words. currently 200
Randomized algorithms exploit stochasticity to reduce computational complexity. One important example is random feature regression (RFR) that accelerates Gaussian process regression (GPR). RFR approximates an unknown function with a random neural network whose hidden weights and biases are sampled from a probability distribution. Only the final output layer is fit to data. In randomized algorithms like RFR, the hyperparameters that characterize the sampling distribution greatly impact performance, yet are not directly accessible from samples. This makes optimization of hyperparameters via standard (gradient-based) optimization tools inapplicable. Inspired by Bayesian ideas from GPR, this paper introduces a random objective function that is tailored for hyperparameter tuning of vector-valued random features. The objective is minimized with ensemble Kalman inversion (EKI). EKI is a gradient-free particle-based optimizer that is scalable to high-dimensions and robust to randomness in objective functions. A numerical study showcases the new black-box methodology to learn hyperparameter distributions in several problems that are sensitive to the hyperparameter selection: two global sensitivity analyses, integrating a chaotic dynamical system, and solving a Bayesian inverse problem from atmospheric dynamics. The success of the proposed EKI-based algorithm for RFR suggests its potential for automated optimization of hyperparameters arising in other randomized algorithms. 
\end{abstract}

\begin{keywords}
random features, Gaussian process regression, hyperparameter learning, ensemble Kalman inversion, Bayesian inverse problems
\end{keywords}

\section{Introduction}\label{sec:intro}
Nonconvex optimization is ubiquitous in machine learning. In deep learning, it most frequently arises when training a neural network. This involves adjusting the weights and biases of the network to minimize a loss function. Alternatively, one can greatly simplify this optimization task by replacing the inner network weights with fixed samples from a probability distribution \citep{huang2006extreme,ScaWan17} and only learning the neural network's linear output layer. With a suitable choice of loss function, such as a (possibly regularized) least squares functional, this optimization problem becomes a convex program. In the least squares example, its explicit solution involves the inversion of a linear system. Such an approach reduces approximation power, but does not have to deal with the challenges that come with nonconvexity. A large body of empirical evidence demonstrates that these convex reductions display remarkable performance in streaming large data sets at moderate processing and memory costs during training, partly because they do not require backpropagation \citep{ChaHasSub20, LeSarSmo13, SunZhaZhu15, bollt2021explaining}. Performance on (often high-dimensional) problems with little data available for training is less explored~\citep{BarBolGauGri21,lanthaler2023error}. 

Naturally, there is a hidden price attached to such randomized approximations: one must pose additional structure on the problem in the form of the probability distribution $\mu$ from which the neural network weights are sampled. Critically, approximation accuracy is strongly tied to the choice of $\mu$. Unsystematic choice of $\mu$ may misrepresent performance or decrease robustness and usability. Indeed, practitioners often resort to unsystematic approaches such as manual tuning, grid searches, or fixed samples that lead to user bias, scalability issues, and overfitting, respectively. One systematic method that the present paper advocates for is to (\emph{i}) choose a $u$-parametrized family $\{\mu_u\}_{u\in\cU}$ of probability distributions, (\emph{ii}) design a regularized stochastic optimization problem for the hyperparameter $u$, and (\emph{iii}) optimize over $u\in\cU$ to achieve consistent performance over different random realizations of the algorithm. Stage (\emph{iii}) will be referred to as \nn{\emph{tuning} or \emph{calibration}} for the remainder of this paper.
% \footnote{We will also write \emph{hyperparameter training} for stage (\emph{iii}) to distinguish it from traditional training of machine learning models with hyperparameters held fixed.} 
The simplicity of the convex optimization of final layer weights is now contrasted by a nonconvex stochastic optimization task for the optimal hyperparameters.

Motivated by scientific and engineering contexts, in this paper we instantiate the preceding systematic optimization pipeline for the specific machine learning task of function emulation from a small noisy data set. We apply the framework to one randomized algorithm, the method of \emph{random features}~\citep[RFs,][]{RahRec07}. We benchmark the RF approach with the current state-of-the-art tool for such tasks, the kernel-based method of \emph{Gaussian processes}~\citep[GPs,][]{RasWil06}. The underlying hyperparameter \nn{tuning} problem for the RF method involves optimally adapting the RF sampling distribution $\mu_u$ to data. This is equivalent to learning a data-adapted kernel or Bayesian prior distribution. Indeed, RFs, which in special cases can be viewed as neural networks with randomly sampled weights (e.g., ``extreme learning machines'' from \citet{huang2006extreme}), by construction can also be viewed as low-rank approximations to GPs. A broad chronology of the developments and performance gains from RF approximation of kernel and GP methods can be found in the work of~\citet{CheHuaLiuSuy22}.

The majority of existing studies do not systematically optimize the RF sampling distribution $\mu_u$. Of those that do, the approach of \citet{SmoSonWilZic15} is closely related to our own. The authors adopt a GP perspective on RFs by deriving an objective function from empirical Bayes arguments. The work fixes the number and specific realizations of the sampled features. Although this produces accurate results, their approach only fits to the individual sample and not its law, which prohibits future resampling and flexible computational cost adjustments. Other fixed-feature approaches use random projections \citep{DeCGitHamXia14} or discrete kernel alignment \citep{DucSin16} to create an optimal weighting of a pre-generated set of RFs. More modern methods learn a generative model for the feature distribution \citep{ChaLiMroPocYan19,falk2022implicit}; this results in more flexible RF methods. Another line of research derives explicit formulas for the ``optimal'' RF distribution based on minimization of upper bounds on the approximation error~\citep{kammonen2020adaptive,kammonen2023smaller}. The authors then approximately sample from this inaccessible distribution with Markov chain Monte Carlo (MCMC) methods. To summarize, most existing works that do adopt systematic hyperparameter calibration pipelines must exploit special properties and structure of the distribution family $\{\mu_u\}$. In contrast, the present paper introduces a widely applicable hyperparameter learning methodology that treats $\{\mu_u\}$ in a black-box manner.

The optimization workhorse of this paper is the ensemble Kalman inversion (EKI) algorithm \citep{IglLawStu13,SchStu17,CalReiStu22}, one of a family of derivative-free algorithms designed to solve nonlinear optimization problems that arise from Bayesian inversion. It can be configured to solve for both a maximum likelihood estimator or a maximum a posteriori estimator. For linear inversion with Gaussian priors and Gaussian observational noise models, the EKI algorithm converges to the true point estimator of the Bayesian posterior distribution. As a particle-based method, EKI can be derived from the ensemble Kalman filtering (EnKF) literature \citep{CheOli12, IglLawStu13} and, along with its variants, has been used successfully in a wide range of optimization and inversion tasks \citep{Bot23_pre, CleGarLanSchStu21, DunGarSchStu21, DunDunStuWol22, BacDunGarHuaLopWu22, HuaHuaReiStu22, KovStu19, XiaSunRoyWanWu16}. In part, this success is due to the method's robustness to rough optimization landscapes and the inheritance of modifiers (e.g., localization, sampling error correction, covariance inflation) and variants (e.g., square root inversion, unscented inversion, sparsity-promoting inversion) from the decades-long history of EnKF in meteorological applications. Since EKI treats the hyperparameter-to-output map as a black-box (in particular, without access to derivatives of the map), it allows for easy prototyping and modularity over different families of features, hyperparameters, and objective functions. \nn{On the other hand, algorithms that make use of higher-order derivative information tend to converge faster than those that do not. However, the derivative-free EKI algorithm is still known to approximate a gradient descent dynamic and hence acceleration methods such as momentum are readily applicable to EKI~\citep[Section 4]{KovStu19}.}

\subsection{Contributions}
The primary purpose of this paper is to build a robust optimization framework for the automated calibration of hyperparameters that appear in randomized algorithms. The approach developed in this work is derivative-free, which enables a black-box treatment of hyperparameters at the level of probability distributions instead of at the level of individual samples. Highlighting our framework in the specific context of RF algorithms, in this paper we make the following contributions.
\begin{enumerate}[label=(C\arabic*)]
    \item Building off of empirical Bayes, we introduce a stochastic objective function for RF distribution hyperparameters that is specifically tailored to the EKI algorithm.
    
    \item We demonstrate that optimized hyperparameters obtained from moderate numbers of EKI iterations are not overfit to specific feature samples. This offers new flexibility such as being able to use a different numbers of features for hyperparameter optimization than for downstream tasks such as prediction. 
    
    \item We contextualize the proposed machine learning tools with the requirements of common scientific or engineering model emulation tasks. In particular, the following constraints are imposed on the learning tasks:
    \begin{itemize}
    \item[(i)] the amount of available data is severely limited, as these must be generated by scientific computer codes that may be expensive to run;
    \item[(ii)] inputs and outputs are both multidimensional and correlated, as often such relationships are unknown a priori;
    \item[(iii)] output observations are corrupted with noise. 
    \end{itemize} 
    This is in contrast to much of the broader machine learning literature, which is more concerned with scaling performance on massive data sets and more reliant on approaches such as exact diagonalization to tackle multi-output problems.
    
    \item We test the new machine learning tools on three scientifically-minded benchmark problems. The tools are used to emulate scalar-valued functions for global sensitivity analysis, integrate the state a chaotic Lorenz 63 dynamical system, and accelerate a Bayesian parameter estimation problem from the atmospheric sciences by emulating the underlying forward map. In all three applications, we compare our approach with a current gold-standard GP-based methodology.
    
    \item We document and actively maintain all tools and examples in open-source, registered, Julia-language GitHub repositories~\citep{BacDunGarHuaLopWu22,Dun_etal24} \nn{with documentation at:
    \begin{itemize}
        \item[(i)]{\makebox[2cm][l]{\sfit{(EKI)}}\href{https://clima.github.io/EnsembleKalmanProcesses.jl/dev/} {\texttt{clima.github.io/EnsembleKalmanProcesses.jl/dev/}}\,,
        }

        \item[(ii)]{\makebox[2cm][l]{\sfit{(RF)}}\href{https://clima.github.io/RandomFeatures.jl/dev/}{\texttt{clima.github.io/RandomFeatures.jl/dev/}}\,,
        }

        \item[(iii)]{\makebox[2cm][l]{\sfit{(Examples)}}\href{https://clima.github.io/CalibrateEmulateSample.jl/dev/}{\texttt{clima.github.io/CalibrateEmulateSample.jl/dev/}}\,.
        }
    \end{itemize}
    }
\end{enumerate}

\subsection{Outline}
The remainder of this paper is organized as follows. Section~\ref{sec:rfr} introduces Bayesian regression with RFs from a GP perspective for both scalar-valued and vector-valued function learning settings. Section~\ref{sec:hyp_learning} motivates the RF hyperparameter learning problem from empirical Bayes ideas. It also presents the black-box EKI optimizer that is championed in this work and a random objective function that is amenable to EKI.
Section~\ref{sec:applications} applies the \nn{tuned} RF emulators to three scientific applications. The paper concludes with a final discussion in Section~\ref{sec:conc}. Appendix~\ref{app:details} contains additional details regarding the numerical experiments from Section~\ref{sec:applications}.

\section{Bayesian Regression with Random Features}\label{sec:rfr}
In this section, we formulate a supervised learning problem in the framework of RFs. To this end, let $\cX\subseteq \R^d$ denote the input space and $\cY$ denote the output space. Suppose that we have access to $N$ pairs of data $\{(x_n,y_n)\}_{n=1}^N\subset\cX\times\cY$. To set the notation, let $X\defeq \{x_n\}_{n=1}^N$, $Y\defeq \{y_n\}_{n=1}^N$, and $D_N\defeq (X,Y)$. Write $[N]\defeq\{1,2,\ldots, N\}$. Although our framework can handle both random or deterministic input data $X$, from here onward we assume that $X$ is fixed to ease notation and simplify the exposition. The goal is to train a model $F\colon\cX\to\cY$ to both ``fit'' the observed data $D_N$ and generalize well, that is, $F(x)\approx y$ in an appropriate sense both for $x\in X$ and $x\notin X$. With only finite data and without further assumptions on how $X$ is related to $Y$, the learning problem is ill-posed. Learning algorithms impose prior knowledge, regularization, or constraints to obtain a stable solution. Subsection~\ref{sec:rfscalar} first presents the \emph{Gaussian process regression} (GPR) solution approach and then its approximation by RFs, \emph{random feature regression} (RFR), in a scalar output setting, that is, $\cY=\R$. This setup is then generalized in Subsection~\ref{sec:rfvector} to the more realistic setting of vector-valued learning, where now $\cY=\R^p$ and output space correlations must be accounted for. Extensions to infinite-dimensional $\cX$ and $\cY$ are also possible~\citep{lanthaler2023error,nelsen2021random} but are not considered here.

\subsection{Scalar-Valued Learning}\label{sec:rfscalar}
We now consider the output space $\cY=\R$ and thus real-valued regression. This is a classical problem and various estimators exist for it, ranging from those based on parametric models or nonparametric models to those that return a single point solution or an entire family of possible solutions in the form of a probability distribution. In this paper, we focus on the latter class of so called Bayesian estimators. GPR is one canonical example from this class. 

\subsubsection{Gaussian Process Regression}\label{sec:gpr_scalar_subsub}
We begin by prescribing the following statistical model for the data $D_N$:

\vspace{-15pt}

\begin{align}\label{eqn:likelihood}
    y_n=F(x_n)+\eta_n ,\qw \eta_n\diid \normal(0,\sigma^2)\qfa n\in[N].
\end{align}
Here $E\defeq \{\eta_n\}_{n=1}^N$ represents noisy observations in the form of independent and identically distributed (i.i.d.) real-valued zero mean Gaussian random variables with common variance $\sigma^2$, and $F$ determines the input-output relationship. The ideal situation is the well-specified setting\footnote{The observed data $D_N$ may not necessarily follow the likelihood model \eqref{eqn:likelihood} in reality. For example the noise may not be i.i.d. Gaussian or may appear multiplicatively. From this perspective, $\sigma>0$ may be viewed as a tunable hyperparameter of the GPR algorithm and not fixed in advance by the data.} in which the observed $D_N$ is actually generated according to \eqref{eqn:likelihood} for some ground truth function $F=\Fd$. 

GPR proceeds by imposing additional structure on \eqref{eqn:likelihood} in the form of a prior probability distribution over the function $F$. That is, we model $F$ as a GP, independent of $E$, given by 

\vspace{-15pt}

\begin{align}\label{eqn:prior_gp}
   F\sim \GP\bigl(\bbar{F},K\bigr).
\end{align}
This means that $F\colon \cX\to\R$ is a \emph{random function} with mean function $\bbar{F}\colon\cX\to\R$ and symmetric covariance function $K\colon\cX\times\cX\to\R$ such that for any $T\in\N$ and any distinct points $Z\defeq \{z_t\}_{t=1}^T\subset \cX$, the finite-dimensional marginals $F(Z)\defeq (F(z_1), F(z_2),\ldots, F(z_T))^\tp\in\R^T$ are distributed according to

\vspace{-15pt}

\begin{align}\label{eqn:marginal}
    F(Z) \sim\normal\bigl(\bbar{F}(Z), K(Z,Z)\bigr),
\end{align}
where the mean vector and positive semidefinite covariance matrix of the multivariate Gaussian distribution in \eqref{eqn:marginal} are given by

\vspace{-15pt}

\begin{align}
    \bbar{F}(Z)\defeq \bigl(\bbar{F}(z_1), \ldots, \bbar{F}(z_T)\bigr)^\tp \in\R^T \qa K(Z,Z)\defeq \bigl[K(z_i,z_j)\bigr]_{i,j\in[N]}\in\R^{T\times T}.
\end{align}
Typically the prior mean $\bbar{F}$ is set to zero. The choice of covariance function $K$ greatly influences prediction accuracy and is informed by prior knowledge about the problem, such as smoothness, sparsity, and lengthscales.

To solve the original regression problem, GPR simply applies Bayes' rule by conditioning the prior function $F$ on the observed data $D_N$ to obtain the \emph{posterior distribution}

\vspace{-15pt}

\begin{align}\label{eqn:posterior_scalar}
    F\condbar D_N \sim \GP\bigl(\bbar{F}^{(N)}, K^{(N)}\bigr),
\end{align}
where the posterior mean function and posterior covariance function are given by\footnote{The posterior mean formula here is simply \emph{kernel ridge regression} \citep{HenKanSejSri18_pre} with a specific ridge penalty parameter depending on $\sigma^2$ and $N$.}

\vspace{-15pt}

\begin{align}\label{eqn:posterior_scalar_formula}
\begin{split}
    \bbar{F}^{(N)}(x) &= \bbar{F}(x) + K(x,X)\bigl(K(X,X) + \sigma^2I_N\bigr)^{-1}\bigl(Y - \bbar{F}(X)\bigr) \qa \\
\Kpost(x,x') &= K(x,x') - K(x,X)\bigl(K(X,X) + \sigma^2 I_N\bigr)^{-1}K(X,x')
\end{split}
\end{align}
for any $x\in\cX$ and $x'\in\cX$ \citep{RasWil06}. Here $I_N\in \R^{N\times N}$ represents the $N$-dimensional identity and 

\vspace{-15pt}

\begin{align}
K(X,x) \defeq \bigl(K(x_1,x), \ldots, K(x_N,x)\bigr)^\tp\in\R^N
\end{align}
is the vector of cross covariances between the training inputs and point $x \in \cX$. We identify $K(x,X)\defeq K(X,x)^\tp\in\R^{1\times N}$ as a row vector. The whole posterior distribution \eqref{eqn:posterior_scalar} itself is the GPR estimator. The posterior mean is a standard point predictor and can be related to deterministic kernel methods. The posterior covariance enables the quantification of uncertainty in the prediction and beyond. The underlying regression problem is actually regularized by the prescription of the prior distribution over $F$ in the sense that the Bayesian inversion $D_N\mapsto F\condbar D_N$ is a stable mapping~\citep{stuart2010inverse}. The posterior also being Gaussian is due to the fact that both the likelihood~\eqref{eqn:likelihood} and prior~\eqref{eqn:prior_gp} were assumed Gaussian.\footnote{One can also perform GPR with non-Gaussian likelihoods. However, the posterior will no longer be Gaussian in this case and the closed form expressions~\eqref{eqn:posterior_scalar_formula} for the mean and covariance will no longer be valid. More sophisticated approaches to compute the posterior, such as MCMC, would be required.}

The closed form expressions for the posterior mean and covariance functions in \eqref{eqn:posterior_scalar_formula} are what standard GPR software packages implement in practice. For $x\in\cX$, these implementations involve solving the linear systems
\begin{equation}\label{eqn:1dlinsys}
\bigl(K(X,X) + \sigma^2 I_N\bigr)\alpha  = \bigl(Y-\bbar{F}(X)\bigr) \qa \bigl(K(X,X) + \sigma^2 I_N\bigr)\alpha(x)  = K(X,x) 
\end{equation}
for coefficients $\alpha \in \R^N$ and $\alpha(x) \in \R^N$, respectively. We observe here that the linear algebraic complexity falls into three stages: \emph{space} (i.e., memory storage), \emph{offline} (i.e., operations on stored objects), and \emph{online} (i.e., operations for evaluation at new $x\in\cX$). GPR is a nonparametric method that requires access to the full input data $X$ to evaluate the posterior mean and covariance at online cost $\cO(N)$. Storing $X$ requires $\cO(dN)$ memory. Since $(K(X,X) + \sigma^2 I_N)$ is a dense matrix in general, it requires $\mathcal{O}(N^2)$ storage in space. Often applications require the solution of the linear systems \eqref{eqn:1dlinsys} many times with the same system matrix. In this case, we use a Cholesky factorization that has an offline cost of $\mathcal{O}(N^3)$ operations. Then the cost of obtaining $\alpha$ in \eqref{eqn:1dlinsys} is reduced to $\mathcal{O}(N^2)$. Similarly, for a new input $x\in\cX$, we can use the precomputed Cholesky factor to solve for $\alpha(x)$ with cost $\mathcal{O}(N^2)$. Nevertheless, for problems with large $N$, such costs make GPR infeasible as presented. This motivates randomized approximations that reduce computational complexity.

\subsubsection{Random Feature Regression}\label{sec:rfr_scalar_subsub}
Several scalable approximations to GPR have been proposed in the literature. These include, but are not limited to, sparse variational approaches based on \nn{inducing points \citep{hensman2015scalable,hensman2018variational,titsias2009variational},} Nystr\"om subsampling \citep{SunZhaZhu15}, and RFs \citep{RahRec07}. We focus on RFs and in particular the method of RFR. We interpret RFR as an approximation to GPR obtained by first replacing the GP prior covariance $K$ with a low-rank approximation $K_M$ and then performing exact GPR inference \eqref{eqn:posterior_scalar} with this low-rank covariance function. We now describe this procedure in detail.

To begin, let $\Theta$ be a set and $\varphi\colon \cX\times\Theta \to \R$ be a feature map. Let $\mu$ be a probability distribution over $\Theta$. The choice of pair $(\varphi,\mu)$ defines the particular RF method. From this pair, define a GP prior covariance function $K\colon\cX\times\cX\to\R$ for $x$ and $x'$ in $\cX$ by
\begin{equation}\label{eqn:cov_limit_rf}
K(x,x') \defeq \E_{\theta\sim\mu}\bigl[\varphi(x;\theta) \varphi(x';\theta)\bigr].
\end{equation}
A covariance defined this way is always symmetric and positive semidefinite by construction. Conversely, any given GP covariance function $K$ can be written in the form \eqref{eqn:cov_limit_rf} for some set $\Theta$ and RF pair $(\varphi,\mu)$.\footnote{Indeed, by definition $\GP(0,K)$ satisfies $K(x,x')=\E_{F\sim \GP(0,K)}[F(x)F(x')]$. To satisfy \eqref{eqn:cov_limit_rf}, take $\Theta$ to be a sufficiently smooth function space, $\mu=\GP(0,K)$, and $\varphi(x;\theta)=F(x)$ with $\theta=F\sim\mu$.} In practice, many popular covariance functions fit such a form \citep[Supplement E]{rudi2017generalization}. The functions $\varphi(\slot;\theta)$ with $\theta\sim\mu$ are also called RFs and play a role similar to that of feature maps in the kernel methods literature.

So far, the kernel $K$ is completely deterministic. RF methods apply a Monte Carlo approximation to the integral in \eqref{eqn:cov_limit_rf}, replacing the full expectation with the empirical average
\begin{equation}\label{eqn:cov_finite_rf}
K_{M}(x,x') \defeq \frac{1}{M} \sum_{m=1}^M \varphi(x;\theta_m)\varphi(x';\theta_m),\qw \theta_m\diid \mu,
\end{equation}
for $x\in\cX$ and $x'\in\cX$ and some $M\in\N$. This is a random finite rank  prior covariance function. RFR is then simply the process of applying Bayes' rule to the RF prior $\GP(\bbar{F}, K_M)$ under the likelihood model~\eqref{eqn:likelihood} to obtain the RF posterior $\GP(\bbar{F}_M^{(N)}, K_M^{(N)})$. Here, the RF posterior mean $\bbar{F}_M^{(N)}$ and RF posterior covariance $K_M^{(N)}$ are given by \eqref{eqn:posterior_scalar_formula} except with all instances of $K$ replaced by $K_M$. In terms of sample paths, we write $F_M\sim \GP(\bbar{F}, K_M)$ for the RF prior and $F_M\condbar D_N\sim \GP(\bbar{F}_M^{(N)}, K_M^{(N)})$ for the RF posterior.
Although it appears that RFR is just GPR with a special choice of prior, from an implementation and practical point of view it is provably beneficial to view RFR as an entirely distinct methodology~\citep{RahRec08,rahimi2008weighted}. Indeed, RFR can be interpreted as a finite rank Bayesian linear model in weight space \citep[Section 3.3]{bishop2006pattern}, which simplifies sampling of the posterior distribution. Moreover, in what follows we show that RFR involves different system matrices that are cheaper to invert and coefficients that have different interpretations than those in appearing in GPR. We leave a rigorous analysis of the convergence of RFR to GPR to future work.

We now follow through the linear algebra to define the system matrices that characterize the posterior mean and covariance of RFR. To this end, we adopt the notation

\vspace{-15pt}

\begin{align}\label{eqn:rf_phi_vec}
    \begin{split}
        \Phi_M(x) &\defeq \bigl(\varphi(x;\theta_1),\ldots,\varphi(x;\theta_M)\bigr) \in \R^{1\times M} \qa \\
        \Phi_M(X) &\defeq \bigl[\varphi(x_n;\theta_m)\bigr]_{n\in[N],m\in[M]} \in \R^{N\times M}
    \end{split}
\end{align}
for $x\in\cX$. Thus $K_M(x,x')=\Phi_M(x)\Phi_M(x')^\tp/M$. We observe from \eqref{eqn:posterior_scalar_formula}~and~\eqref{eqn:1dlinsys} that $\bbar{F}_M^{(N)}(x) = \bbar{F}(x) + K_M(x,X)\al = \bbar{F}(x) + M^{-1}\Phi_M(x)\beta$, where
\begin{align}
    \beta\defeq \Phi_M(X)^\tp \al \in \R^M.
\end{align}
This shows that the difference between the posterior mean and the prior mean belongs to the linear span of the RFs $\{\varphi(\slot;\theta_m)\}_{m=1}^M$. We see that the random neural network example from Section~\ref{sec:intro} holds whenever the RFs $\varphi(\slot;\theta)$ take the form of a hidden neuron with sampled weights $\theta\sim \mu$ (or combinations thereof). The coefficients $\beta$ represent the final linear output layer weights in this analogy. It remains to derive a computationally tractable equation for $\beta$. To do this, left multiply the first equation in \eqref{eqn:1dlinsys} by $\Phi_M(X)^\tp$ to obtain

\vspace{-15pt}

\begin{align}\label{eq:beta_scalar}
\biggl(\frac{1}{M}\Phi_M(X)^{\tp} \Phi_M(X) + \sigma^2 I_{M}\biggr)\beta = \Phi_M(X)^{\tp}\bigl(Y-\bbar{F}(X)\bigr).
\end{align}
Thus, obtaining $\beta$ only requires the inversion of an $M\times M$ matrix instead of an $N\times N$ one. This is advantageous whenever $M\ll N$.

We similarly seek to write the RF posterior covariance function in the span of the RFs. First, we project the second equation in \eqref{eqn:1dlinsys} to obtain the parametrized $M\times M$ system
\begin{equation}\label{eq:betax_scalar}
\biggl(\frac{1}{M}\Phi_M(X)^{\tp} \Phi_M(X) + \sigma^2 I_{M}\biggr) \beta(x') =  \biggl(\frac{1}{M}\Phi_M(X)^{\tp} \Phi_M(X)\biggr) \Phi_M(x')^{\tp}.
\end{equation}
for $\beta(x')\defeq \Phi_M(X)^\tp\al(x')\in\R^M$. From the second equation in \eqref{eqn:posterior_scalar_formula}, we deduce that
$K_M^{(N)}(x,x')=K_M(x,x') - M^{-1}\Phi_M(x)\beta(x') = M^{-1}\Phi_M(x)(\Phi_M(x')^\tp - \beta(x'))$. However, realizing that the system matrix on the left hand side of \eqref{eq:betax_scalar} is actually the precision matrix of the RF weight space posterior \citep[Section 3.3]{bishop2006pattern} reveals an even more efficient form of $K_M^{(N)}$. This form requires the solution $\widehat{\beta}(x')$ of 
\begin{equation}\label{eq:betax_scalar_weight_space}
\biggl(\frac{1}{M}\Phi_M(X)^{\tp} \Phi_M(X) + \sigma^2 I_{M}\biggr) \widehat{\beta}(x') = \Phi_M(x')^{\tp}
\end{equation}
and use of the matrix identity $I-(A+\sigma^2 I)^{-1}A=\sigma^2 (A+\sigma^2 I)^{-1}$ for symmetric positive semidefinite $A$. We summarize the closed form formulas for the RF posterior mean and covariance functions in terms of the coefficients $\beta$ from \eqref{eq:beta_scalar} and $\widehat{\beta}(x')$ from \eqref{eq:betax_scalar_weight_space} as

\vspace{-15pt}

\begin{align}\label{eqn:rf_scalar_summary}
    \begin{split}
        \bbar{F}^{(N)}_M(x) &= \bbar{F}(x) + \frac{1}{M}\Phi_M(x)\beta = \bbar{F}(x) + \frac{1}{M}\sum_{m=1}^M\beta_m\varphi(x;\theta_m) \qa  \\
    K_M^{(N)}(x,x') &= \frac{\sigma^2}{M} \Phi_M(x)\widehat{\beta}(x')= \frac{\sigma^2}{M} \sum_{m=1}^M\bigl(\widehat{\beta}(x')\bigr)_m \,\varphi(x;\theta_m)
    \end{split}
\end{align}
for any $x$ and $x'$ in $\cX$. 

We now assess the computational complexity for RFR. Since RFR is a parametric method of estimation, we can discard the data $X$ from memory once the dense matrix $\Phi_M(X)\in\R^{N\times M}$ is stored with cost $\cO(NM)$. The new system matrix in \eqref{eq:beta_scalar} leads to offline complexity $\mathcal{O}(M^3)$ for a Cholesky factorization. Subsequent solves for $\beta$ in \eqref{eq:beta_scalar} offline and $\widehat{\beta}(x)$ in \eqref{eq:betax_scalar} online both have complexity $\mathcal{O}(M^2)$. The online cost to evaluate the RF posterior mean and covariance functions is $\cO(M)$. Thus, RFR is computationally advantageous to GPR if the RF regression error is comparable to that of GPR even with $M\ll N$. This is indeed true. Theoretical analysis establishes that RFR delivers the same asymptotic error rate as GPR with only $M =\mathcal{O}(\sqrt{N}) $ RFs \citep{lanthaler2023error}.

We conclude this subsection with an example of a commonly used RF pair $(\varphi,\mu)$.
\begin{example}[Random Fourier Features for RBF Kernel]
The pair $(\varphi,\mu)$ with 

\vspace{-15pt}

\begin{align}
    x\mapsto \varphi(x;\theta)\defeq \sqrt{2}\cos(a^\tp x + b) \qa \theta=(a,b)\sim \mu\defeq \normal(0,\ell^{-2}I)\otimes \mathsf{Unif}(-\pi,\pi)
\end{align}
satisfies the identity \eqref{eqn:cov_limit_rf} for the squared exponential covariance function

\vspace{-15pt}

\begin{align}
    (x,x')\mapsto K(x,x')=\exp\biggl(-\frac{\norm{x-x'}^2_2}{2\ell^2}\biggr).
\end{align}
The lengthscale $\ell>0$ is the only hyperparameter appearing in this kernel.
\end{example}
More details about random Fourier features are provided in Section~\ref{sec:kernel_structure}.

\subsection{Vector-Valued Learning}\label{sec:rfvector}
In this subsection, we summarize the Bayesian regression framework in the setting of vector-valued outputs. Let $\cY = \R^p$ for some $p\in\N$. We prescribe the likelihood model

\vspace{-15pt}

\begin{align}\label{eqn:likelihood_vector}
    y_n=F(x_n)+\eta_n ,\qw \eta_n\diid \normal(0,\Sigma)\qfa n\in[N].
\end{align}
Here $\Sigma\in \R^{p\times p}$ is the symmetric positive definite covariance matrix of the Gaussian noise. It is common to model the noise by $\Sigma=\sigma^2 I_p$, but non-isotropic covariances are also natural depending on the application. The function $F$ in \eqref{eqn:likelihood_vector} maps $\cX$ to $\R^p$. Independent GP or RF priors on $F$ lead to GP or RF posteriors, respectively. We first describe the GP setting before moving on to the computationally advantageous RF setting.

\subsubsection{Vector-Valued Gaussian Process Regression}\label{sec:gpr_vector_subsub}
We model $F$ with an $\R^p$-valued GP prior $F\sim \GP(\bbar{F}, K)$. The mean function $\bbar{F}\colon\cX\to\R^p$ is vector-valued. More importantly, the covariance function $K\colon\cX\times\cX\to\R^{p\times p}$ is matrix-valued in order to model correlations in the output space. Several examples of matrix- or operator-valued covariance kernels may be found in the work of \citet{brault2016random,kadri2016operator,micchelli2004kernels,micchelli2005learning,nelsen2021random}.

Under this $\R^p$-valued GP prior, the resulting posterior $F\condbar D_N\sim\GP(\bbar{F}^{(N)}, \Kpost)$ under the likelihood \eqref{eqn:likelihood_vector} is also an $\R^p$-valued GP with mean and covariance functions

\vspace{-15pt}

\begin{align}\label{eqn:posterior_vector_formula}
\begin{split}
    \bbar{F}^{(N)}(x) &= \bbar{F}(x) + K(x,X)\bigl(K(X,X) + B_\Sigma)^{-1}\bigl(Y - \bbar{F}(X)\bigr) \qa \\
\Kpost(x,x') &= K(x,x') - K(x,X)\bigl(K(X,X) + B_\Sigma \bigr)^{-1}K(X,x')
\end{split}
\end{align}
for any $x\in\cX$ and $x'\in\cX$. These equations are interpreted in block form. In particular, $B_\Sigma \defeq \mathrm{blockdiag}(\Sigma,\ldots,\Sigma) \in (\R^{p\times p})^{N\times N}$, $K(X,X)\defeq \bigl[K(x_i,x_j)\bigr]_{i,j\in[N]} \in (\R^{p\times p})^{N\times N}$, and

\vspace{-15pt}

\begin{align}
    K(X,x')\defeq \bigl(K(x_1,x'), \ldots, K(x_N,x')\bigr)^\tp\in\bigl(\R^{p\times p}\bigr)^N.
\end{align}
The factor $K(x,X)=K(X,x)^\tp \in (\R^{p\times p})^{1\times N}$ is interpreted as a ``row vector'' with matrix entries. The concatenated data $Y=\{y_n\}_{n=1}^N$ and prior mean entries $\bbar{F}(X)$ are viewed as length $N$ vectors with each entry taking value in $\R^p$. The formulas~\eqref{eqn:posterior_vector_formula} follow \nn{from whitening \eqref{eqn:likelihood_vector} by pre-multiplying by $\Sigma^{-1/2}$ and then applying an existing white noise result \citep[Theorem 5.3]{owhadi2022ideas}.}

Although \eqref{eqn:posterior_vector_formula} visually mimics its scalar-valued counterpart \eqref{eqn:posterior_scalar_formula}, the computational effort is greatly magnified in the vector-valued setting. Effectively, the sample size $N$ is replaced by $Np$ in all scalar-valued output space GPR complexity estimates. Indeed, \eqref{eqn:posterior_vector_formula} requires solving $Np$ by $ Np$ linear systems of equations instead of $N$ by $N$ ones. Table~\ref{tab:complexity} summarizes upper bounds on the space, offline, and online costs of the approach. This poor scaling with output space dimension $p$ has severely limited the practical utility of vector-valued GPR to date. Indeed, existing studies usually only work with separable (or even scalar) matrix-valued kernels, which suffer from a simplistic rank-one tensor structure \citep{kadri2016operator,owhadi2022ideas}, or diagonal kernels, which ignore correlations in the output space \citep{CleGarLanSchStu21}. Vector-valued RFs enable scalable approximate GP inference without neglecting correlations in the output space.

\subsubsection{Vector-Valued Random Feature Regression}\label{sec:rfr_vector_subsub}
Vector-valued RFR greatly alleviates the cost of learning with matrix-valued covariance kernels. The method is characterized by a vector-valued feature map $\varphi\colon \cX\times\Theta\to\R^p$ and a probability distribution $\mu$ over $\Theta$. We work with RF kernels of the form
\begin{equation}\label{eqn:cov_vector_rf}
K(x,x') \defeq \E_{\theta\sim\mu}\Bigl[\varphi(x;\theta) \varphi(x';\theta)^\tp\Bigr]\qa K_{M}(x,x') \defeq \frac{1}{M} \sum_{m=1}^M \varphi(x;\theta_m)\varphi(x';\theta_m)^\tp,
\end{equation}
where $\theta_m\diid \mu$ for $m\in[M]$ and $x$ and $x'$ belong to $\cX$.
The vector-valued RFR method proceeds by assigning a RF prior distribution $F_M\sim \GP(\bbar{F},K_M)$ to $F$ in \eqref{eqn:likelihood_vector}.
The RF posterior is then given by the vector-valued Gaussian process $\GP(\bbar{F}_M^{(N)}, K_M^{(N)})$, where

\vspace{-15pt}

\begin{align}\label{eqn:rf_vector_summary}
\begin{split}
    \bbar{F}^{(N)}_M(x) &= \bbar{F}(x) + \frac{1}{M}\sum_{m=1}^M\beta_m\varphi(x;\theta_m) \qa  \\
K_M^{(N)}(x,x') &= \frac{1}{M} \sum_{m=1}^M \varphi(x;\theta_m) \bigl(\widehat{\beta}(x')\bigr)_m
\end{split}
\end{align}
for any $x$ and $x'$ in $\cX$. The coefficients $\beta\in\R^M$ and $\widehat{\beta}(x')\in\R^{M\times p}$ solve the linear systems

\vspace{-15pt}

\begin{align}\label{eq:coeff_vector_summary}
\begin{split}
    \biggl(\frac{1}{M}\Phi_M(X)^{\tp} B_\Sigma^{-1} \Phi_M(X) + I_{M}\biggr) \beta &= \Phi_M(X)^{\tp}B_\Sigma^{-1}\bigl(Y - \bbar{F}(X)\bigr) \qa \\
\biggl(\frac{1}{M}\Phi_M(X)^{\tp}B_\Sigma^{-1} \Phi_M(X) + I_{M}\biggr) \widehat{\beta}(x') &= \Phi_M(x')^{\tp},
\end{split}
\end{align}
respectively. The term $\Phi_M(x')\in \R^{p\times M}$ is analogous to the first line of \eqref{eqn:rf_phi_vec}. The formulas in \eqref{eq:coeff_vector_summary} are derived using an approach similar to that in Subsection~\ref{sec:rfr_scalar_subsub}, except now with an additional left multiplication by $B_\Sigma^{-1/2}$. More precisely, we define \eqref{eq:coeff_vector_summary} by

\vspace{-15pt}

\begin{align}\label{eq:coeff_vector_detailed}
\begin{split}
    &\sum_{m=1}^M\biggl(\frac{1}{M}\sum_{n=1}^N\varphi(x_n;\theta_\ell)^{\tp}\Sigma^{-1}\varphi(x_n;\theta_m) + \delta_{\ell m}\biggr) \beta_m = \sum_{n=1}^N\varphi(x_n;\theta_\ell)^{\tp}\Sigma^{-1}\bigl(y_n - \bbar{F}(x_n)\bigr) \qa \\
&\sum_{m=1}^M\biggl(\frac{1}{M}\sum_{n=1}^N\varphi(x_n;\theta_\ell)^{\tp}\Sigma^{-1}\varphi(x_n;\theta_m) + \delta_{\ell m}\biggr)\bigl(\widehat{\beta}(x')\bigr)_m = \varphi(x';\theta_\ell)^{\tp},
\end{split}
\end{align}
where $\ell\in[M]$ \nn{and $\delta_{\ell m}=1$ if $\ell=m$ and equals zero otherwise}. The $\Sigma$-weighted system Gram matrices in \eqref{eq:coeff_vector_detailed} are still only of size $M$ by $M$. These are much cheaper to invert than the vector-valued GPR system matrices whenever $M\ll Np$; this is often the case in practice. Table~\ref{tab:complexity} summarizes the computational tradeoffs. We observe that the complexity of vector-valued GPR depends directly on $Np$, with cubic scaling in $Np$ offline and quadratic scaling online for storage. On the other hand, the complexity of RFR depends instead on $M$ instead of $Np$, with similar scaling in this variable. Thus, RFR is computationally advantageous whenever $M \ll Np$, while still maintaining the same average prediction accuracy as vector-valued GPR~\citep{lanthaler2023error}.

\begin{table}[tb]
    \centering
    \caption{Computational complexity of important operations in GPR and RFR. The integers $N$, $M$, $d$, and $p$ denote the number of data points, number of random features, dimension of $\cX$, and dimension of $\cY$, respectively. ``Space'' represents a memory cost, ``Offline'' represents a one time precomputation, and ``Online'' represents a cost at every new input $x$ and $x'$.}
    \label{tab:complexity}
    \renewcommand{\arraystretch}{1.2}
    \begin{tabular}{l@{\hspace{10.25mm}}ccc}
        \toprule
        Operation & Cost Type & GPR & RFR \\
        \midrule
        Store dense matrix & Space & $\mathcal{O}\bigl((Np)^2\bigr)$ & $\mathcal{O}\bigl(M^2\bigr)$ \\
        Cholesky factorization & Offline & $\mathcal{O}\bigl((Np)^3\bigr)$ & $\mathcal{O}\bigl(M^3\bigr)$ \\
        Evaluate $\bbar{F}^{(N)}(x)$ & Online & $\mathcal{O}\bigl((Np)pd\bigr)$ &  $\mathcal{O}\bigl(Mpd\bigr)$ \\
        Evaluate $\bbar{K}^{(N)}(x,x')$ & Online & $\mathcal{O}\bigl(Np(Np+pd)\bigr)$ & $\mathcal{O}\bigl(Mp(M+p+d)\bigr)$ \\
        \bottomrule
    \end{tabular}
\end{table}

\section{Hyperparameter Learning for Random Feature Regression}\label{sec:hyp_learning}
Recall that in the development of Section~\ref{sec:rfr}, the RF pair $(\varphi,\mu)$ is assumed given. An often overlooked practical consideration of RFR---or other algorithms based on RFs---is how to choose the pair $(\varphi,\mu)$. This is the natural analog to choosing the covariance kernel $K$ in GPR. To address this kernel selection problem, we begin in Subsection~\ref{sec:ebayes} by adapting the popular empirical Bayes approach for learning priors in GPR to the stochastic setting of RFR. This approach comes with several unique challenges that we address with the derivative-free optimization method known as EKI. We describe the EKI algorithm in Subsection~\ref{sec:eki}. In Subsection~\ref{sec:opt_to_inv}, we cast the covariance kernel learning problem as a statistical inverse problem so that it may be solved with EKI. Subsection~\ref{sec:kernel_structure} instantiates the methodology with an expressive family of parametrized distributions $\{\mu_u\}$ that also performs well in practice. \nn{Subsection~\ref{sec:limitations} acknowledges the limitations of the proposed framework.}

\subsection{Empirical Bayes Motivation}\label{sec:ebayes}
Suppose that $\{\mu_u\}_{u\in\cU}$ is a parametrized family of RF distributions over some hyperparameter set $\cU$. Our framework is still valid if $\varphi$ is also a parametrized family, but for simplicity we assume that the feature map $\varphi$ is fixed. Let $K^{(u)}\colon \cX\times\cX\to\R^{p\times p}$ be the matrix-valued kernel defined by

\vspace{-15pt}

\begin{align}\label{eqn:cov_vec_rf_hyp}
    K^{(u)}(x,x') \defeq \E_{\theta\sim\mu_u}\Bigl[\varphi(x;\theta) \varphi(x';\theta)^\tp\Bigr]
\end{align}
for $x$ and $x'$ in $\cX$ and parametrized by $u\in\cU$. Under the vector-valued likelihood~\eqref{eqn:likelihood_vector}, a fully Bayesian approach to performing GPR with the kernel~\eqref{eqn:cov_vec_rf_hyp} is to build a hierarchical prior model by first assigning a prior distribution to the hyperparameter $u$ and then taking the distribution of $F\condbar u$ to be a centered Gaussian process with covariance function $K^{(u)}$. This defines a prior distribution for the joint vector $(F,u)$. The posterior is then obtained as usual by conditioning $(F,u)$ on the observed data $Y$. In general, however, this posterior is non-Gaussian and requires expensive sampling algorithms to access it.

To overcome this, the empirical Bayes methodology takes an optimization approach \citep{chen2021consistency,naslidnyk2023comparing}. Assuming that $u$ is finite-dimensional and the prior on $u$ is uninformative, such as a uniform distribution, empirical Bayes maximizes the probability density of $u\condbar Y$, which is the joint posterior $(F,u)\condbar Y$ marginalized over $F$. By Bayes' rule, the density of $u\condbar Y$ is proportional to the marginal likelihood $Y\condbar u$. Under our Gaussian data likelihood \eqref{eqn:likelihood_vector}, $Y\condbar u \sim \normal(0, K^{(u)}(X,X) + B_\Sigma)$ so that empirical Bayes minimizes the negative log marginal likelihood

\vspace{-15pt}

\begin{align}\label{eqn:eb_cost}
    L^{(\mathrm{EB})}(u)\defeq Y^\tp\bigl(K^{(u)}(X,X) + B_\Sigma\bigr)^{-1}Y + \log\det\bigl(K^{(u)}(X,X) + B_\Sigma\bigr)
\end{align}
over $u\in\cU$. This inspires an empirical RF approximation to the true covariance matrices appearing in \eqref{eqn:eb_cost}. Let $K_M^{(u)}$ be as in \eqref{eqn:cov_vector_rf} with $\mu=\mu_u$. Then define the function

\vspace{-15pt}

\begin{align}\label{eqn:eb_cost_rf_temp}
    \widetilde{L}^{(\mathrm{EB})}_M(u)\defeq Y^\tp\bigl(K_M^{(u)}(X,X) + B_\Sigma\bigr)^{-1}Y + \log\det\bigl(K_M^{(u)}(X,X) + B_\Sigma\bigr).
\end{align}
For sufficiently large $M$, we expect $\widetilde{L}^{(\mathrm{EB})}_M$ to be close to $L^{(\mathrm{EB})}$ in some reasonable sense.

Unfortunately, \eqref{eqn:eb_cost_rf_temp} as written is not efficiently computable. We would like to write it in terms of the smaller RF gram matrix $G_M^{(u)}\in\R^{M\times M}$ defined by

\vspace{-15pt}

\begin{align}\label{eq:rf_gram}
    G_M^{(u)}\defeq \biggl[\frac{1}{M}\sum_{n=1}^N\varphi\bigl(x_n;\theta_\ell^{(u)}\bigr)^{\tp}\Sigma^{-1}\varphi\bigl(x_n;\theta_m^{(u)}\bigr)\biggr]_{\ell\in[M],m\in[M]},
\end{align}
where $\theta_m^{(u)}\sim \mu_{u}$ are i.i.d. samples. To do this, notice that

\vspace{-15pt}

\begin{align}
    \det\bigl(K_M^{(u)}(X,X) + B_\Sigma\bigr)&=\bigl(\det B_\Sigma^{1/2}\bigr)^2\det\bigl(B_\Sigma^{-1/2} K_M^{(u)}(X,X)B_\Sigma^{-1/2}+I_{Np}\bigr)\\
    &=(\det \Sigma)^N\det\Bigl(\frac{1}{M}B_\Sigma^{-1/2} \Phi_M^{(u)}(X)\Phi_M^{(u)}(X)^\tp B_\Sigma^{-1/2}+I_{Np}\Bigr).
\end{align}
This and application of the Weinstein--Aronszajn identity $\det(AB+I)=\det(BA+I)$ for $AB$ and $BA$ square and $I$ of appropriate dimension yield

\vspace{-15pt}

\begin{align}\label{eq:logdet_identity}
    \log\det\bigl(K_M^{(u)}(X,X) + B_\Sigma\bigr) = N\log \det\Sigma + \log \det\bigl(G_M^{(u)} + I_M\bigr). 
\end{align}
To simplify the first term in \eqref{eqn:eb_cost_rf_temp}, let $\alpha^{(u)}\defeq (K_M^{(u)}(X,X) + B_\Sigma)^{-1}Y\in(\R^p)^N$. Then

\vspace{-15pt}

\begin{align}\label{eq:first_term_loss_eb}
     Y^\tp\bigl(K_M^{(u)}(X,X) + B_\Sigma\bigr)^{-1}Y = (\alpha^{(u)})^\tp K_M^{(u)}(X,X) \alpha^{(u)} + (\alpha^{(u)})^\tp B_\Sigma \alpha^{(u)}.
\end{align}
The calculations in Subsection~\ref{sec:rfr_vector_subsub} show that the RFR posterior mean coefficients $\beta=\beta^{(u)}\in\R^M$ from~\eqref{eq:coeff_vector_detailed} satisfy $\beta_m^{(u)}=\sum_{n=1}^N\varphi(x_n;\theta_m^{(u)})^\tp\al_n^{(u)}$. \nn{This implies that the first term on the right hand side of \eqref{eq:first_term_loss_eb} equals $\norm{\beta^{(u)}}^2_{\R^M}/M$. The second term may be written as $(\alpha^{(u)})^\tp B_\Sigma^{\phantom{-1}} B_\Sigma^{-1}\, B_\Sigma^{\phantom{-1}} \alpha^{(u)}=\norm{B_\Sigma^{-1/2}(B_\Sigma \alpha^{(u)})}^2_{(\R^p)^N}$. Using the fact that $B_\Sigma\al^{(u)} = Y-K_M^{(u)}(X,X)\al^{(u)}$, we thus define our practically computable random empirical Bayes objective function to be ${L}^{(\mathrm{EB})}_M(u)\defeq \widetilde{L}^{(\mathrm{EB})}_M - N\log\det(\Sigma)$, which equals}

\vspace{-15pt}

\begin{align}\label{eqn:eb_cost_rf}
    {L}^{(\mathrm{EB})}_M(u) = \frac{1}{M}\norm[\big]{\beta^{(u)}}^2_{\R^M} + \norm[\Big]{B_\Sigma^{-1/2}\bigl(Y-\bbar{F}^{(N)}_M(X;u)\bigr)}^2_{(\R^p)^N} +  \log \det\bigl(G_M^{(u)} + I_M\bigr).
\end{align}
\nn{We subtracted off the term $N\log\det(\Sigma)$~\eqref{eq:logdet_identity}} because it is independent of $u$. The map  $\bbar{F}^{(N)}_M(\slot;u)\nn{=K_M^{(u)}(\slot,X) \alpha^{(u)}}$ in \eqref{eqn:eb_cost_rf} is the RF posterior mean corresponding to feature pair $(\varphi,\mu_u)$.

Although now written in terms of efficiently computable quantities, the actual optimization of the objective function ${L}^{(\mathrm{EB})}_M$ is non-trivial. This is because we view the map $u\mapsto \mu_u$ as a black-box and hence each evaluation of the objective ${L}^{(\mathrm{EB})}_M(u)$ for new $u$ requires resampling the $M$ random variables $\{\theta_m^{(u)}\}$ i.i.d. from $\mu_u$. Although one could attempt to reduce the effect of randomness by drawing enormous numbers of RFs, this will likely require $M\gg Np$ and be computationally infeasible to optimize. Instead, our approach is to use a statistical optimization tool which can handle stochastic model evaluations more naturally.

\subsection{Ensemble Kalman Inversion}\label{sec:eki}
EKI~\citep{IglLawStu13,SchStu17,CalReiStu22}---and its variants~\citep{HuaHuaReiStu22,GarHofLiStu20,GarNusRei20}---is a family of tools for stochastic optimization. More explicitly, EKI is a method for solving Bayesian inverse problems. In the following exposition, we first describe the basics of the algorithm to solve such problems. \nn{Then the following subsection proposes} a new inverse problem whose solution is related to minimization of \eqref{eqn:eb_cost_rf}.

The underlying \nn{generic inverse problem setting} is as follows. Assume a system produces observable output under a parameter-to-observable map $\cG$ that is random with Gaussian marginals.
\nn{That is, the relationship between the parameters $u$ and observable $z$ is given by}

\vspace{-15pt}

\begin{align} \label{eq:ip_sample}
    z &= \cG(u), \qw \cG(u) \sim \normal\bigl(\overline{\cG}(u),\Gamma(u)\bigr).
\end{align}
\nn{In the preceding display, $\Gamma = \Gamma(u)$ is some (possibly $u$-dependent) covariance matrix, and the ``overbar'' notation $\overline{\cG}$ represents the mean of the distribution of $\cG$. Equivalently,}

\vspace{-15pt}

\begin{align}\label{eq:ip_mean}
    z&=\overline{\cG}(u) + \eta, \qw  \eta \sim \normal\bigl(0,\Gamma(u)\bigr).
\end{align}
The form \eqref{eq:ip_mean} emphasizes that the relationship can be written in terms of a deterministic map \nn{$\overline{\cG}$} plus additive noise. By endowing the input $u\in\cU$ with a Bayesian prior distribution, and given a particular observation $z=z^\star$, we can informally state two optimization tasks (i.e., inverse problems): find an optimal $u^\star= u^{(\mathrm{MLE})}$ or $u^\star = u^{(\mathrm{MAP})}$, where

\vspace{-15pt}

\begin{align}\label{eqn:opt_mle_and_map}
   u^{(\mathrm{MLE})} \defeq \argmax_{u\in\cU} \log P(z^\star\condbar u)  \qa  u^{(\mathrm{MAP})} \defeq \argmax_{u\in\cU} \bigl\{
\log P(z^\star\condbar u) + \log P(u)\bigr\}. 
\end{align}
Here \nn{$w\mapsto P(w)$} denotes the probability density function of a finite-dimensional random variable $w$. In \eqref{eqn:opt_mle_and_map}, $P(z^\star\condbar u)$ is the data likelihood for the observation $z^\star$ arising from any input $u$ and $P(u)$ represents the prior on $u$.  Commonly, optimization frameworks evaluate \eqref{eq:ip_mean} to compute $P(z^\star\condbar u)$ and so require approximation of $\overline{\cG}(u)$, \nn{which is a computationally demanding task in many applications} (e.g., via sampling). On the other hand, EKI solves for $u^\star$ from \eqref{eq:ip_sample} and requires only evaluations of \nn{the parameter-to-observable map} $u\mapsto \cG(u)$.

\nn{We now describe EKI.} The basic EKI algorithm makes \nn{a modeling assumption that is} typical of the ensemble Kalman literature by insisting that the prior on parameter $u$ is a Gaussian \nn{distribution} $\normal(m, C)$. Under this condition, we note that

\vspace{-15pt}

\begin{subequations}\label{eqn:opt_mle_and_map_gauss}
\begin{align}
   u^{(\mathrm{MLE})} &= \argmin_{u\in\cU} \norm[\big]{\Gamma(u)^{{-\frac{1}{2}}}\bigl(z^\star - \cG(u)\bigr)}^2  \qa \\
   u^{(\mathrm{MAP})} &= \argmin_{u\in\cU} \Bigl\{\norm[\big]{\Gamma(u)^{{-\frac{1}{2}}}\bigl(z^\star - \cG(u)\bigr)}^2 + \norm[\big]{C^{{-\frac{1}{2}}}(m - u)}^2\Bigr\},
\end{align}
\end{subequations}
where the norms are induced by the Euclidean inner product. EKI represents the initial Gaussian distribution empirically from samples; \nn{these are called ensemble members. Let the size of the ensemble be denoted by $J$, which is a critical hyperparameter in EKI.} The initial ensemble can be written for each member $j\in[J]$ as \nn{i.i.d. samples} $u^{(j)}(t_0) \sim \normal(m, C)$. Given a sequence $\{t_n\}$ of artificial time values, the \nn{Kalman-type} update rule for EKI at iteration $n\in[n_{\mathrm{iter}}-1]$ is given \nn{for each $j\in[J]$} by

\vspace{-15pt}

\begin{subequations}\label{eq:eki_update}
    \begin{align}
    u^{(j)}(t_{n+1}) &= u^{(j)}(t_n) + C^{(u\cG)}(t_n)\bigl(C^{(\cG\cG)}(t_n)+(\Delta t_{n+1})^{-1}\Gamma\bigr)^{-1}
    \bigl(z^{(j)}(t_{n+1}) - \cG(u^{(j)}(t_n))\bigr), \\
    z^{(j)}(t_{n+1}) &= z^\star + \xi^{(j)}_{n+1},\qw \xi^{(j)}_{n+1} \sim \normal(0, (\Delta t_{n+1})^{-1}\Gamma).
    \end{align}
\end{subequations}
\nn{The first equation updates the parameter and the second equation represents a noisy perturbation of the observation $z^\star$. For further interpretations and derivations of the EKI update formulas, see \citet[Chapter 13]{sanz2023inverse} and references therein.}
The time step $\Delta t_{n+1}$ \nn{in the preceding display} is usually constant, e.g., $\Delta t_{n+1} \equiv \Delta t = 1$.
% The term $u^{(j)}(t_n)$ is often also denoted by $u^{(j)}_n$. 
Writing $\overline{u(t)} \defeq \frac{1}{J}\sum_{j=1}^J u^{(j)}(t)$ and similarly\footnote{Note the difference in notation: $\overline{\cG}(u)$, the mean of the distribution of $\cG$ evaluated at $u$, versus $\overline{\cG(u(t))}$, the finite ensemble average of evaluations of $\cG$ over $u^{(j)}(t)$ for $j\in [J]$.} for $\overline{\cG(u(t))}$, for any $t$ the empirical covariance matrices in \eqref{eq:eki_update} are given by

\vspace{-15pt}

\begin{subequations}
    \begin{align}
        C^{(u\cG)}(t) &= \frac{1}{J}\sum_{j=1}^J \Big(u^{(j)}(t) - \overline{u(t)}\Big)\Big(\cG(u^{(j)}(t)) - \overline{\cG(u(t))}\Big)^\tp \qa  \\  
        C^{(\cG\cG)}(t) &= \frac{1}{J}\sum_{j=1}^J \Big(\cG(u^{(j)}(t)) - \overline{\cG(u(t))}\Big)\Big(\cG(u^{(j)}(t)) - \overline{\cG(u(t))}\Big)^\tp. 
    \end{align}
\end{subequations}
\nn{This completes the description of the basic EKI algorithm.}

In the setting of a linear parameter-to-observable map $\cG$, iterating \eqref{eq:eki_update} until \nn{time $t_{n_{\mathrm{iter}}}=T=1$} will approximate the posterior distribution of the Bayesian inverse problem \citep{CalReiStu22}. That is, the mean and covariance of the final ensemble will accurately represent the posterior mean and covariance. In the nonlinear case, iterating until $T=1$ (just one step if $\Delta t\equiv 1$) will approximate $u^{(\mathrm{MAP})}$ in \eqref{eqn:opt_mle_and_map_gauss}. In contrast, upon sending $t\to\infty$, the dynamic will instead approach $u^{(\mathrm{MLE})}$---up to constraints imposed by the span of the initial ensemble. The ensemble is known to greatly underestimate posterior covariances (even at $T=1$), and so these will be disregarded in \nn{this paper except in Section~\ref{sec:ex_cloud}}. Other variants of the ensemble Kalman methodology have been developed to overcome this limitation~\citep{GarHofLiStu20, GarNusRei20}.

The algorithm \eqref{eq:eki_update} is the basic form of EKI. As well as \nn{previously mentioned} variants, there are several features that can be added for additional computational advantages. Many of these may be found in the open-source software package by~\citet{BacDunGarHuaLopWu22} and are used later on in the numerical experiments of the present paper. Detailed  EKI configurations are provided for each \nn{upcoming} experiment in Appendix~\ref{app:details}.

\subsection{Recasting Optimization as Inversion}\label{sec:opt_to_inv}
We construct a suitable inverse problem in the form \eqref{eq:ip_sample} so that the stochastic optimization of the RF empirical Bayes objective function \eqref{eqn:eb_cost_rf} becomes amenable to EKI. \nn{Take $\Omega\subseteq [N]$ and $ \Omega^\comp \subseteq [N]$ to be \emph{training} and \emph{validation} index sets, respectively.} In what follows, the notation $X_\Omega$ denotes the set $\set{x_i\in X}{i\in\Omega}$ and similarly for $Y_\Omega$.
The following items define the constituents of the proposed inversion framework:
\begin{enumerate}[label=(\roman*)]%, ,topsep=1.67ex,itemsep=0.5ex,partopsep=1ex,parsep=1ex]
    \item \sfit{(prior)} A prior probability distribution is taken over the hyperparameters $u\in\cU$.\footnote{The exact details may be found in Appendix \ref{sec:app_hypprior}.} As a canonical example, consider a finite-dimensional vector hyperparameter $u$ where the prior is given by independent probability distributions in each coordinate. These univariate distributions are chosen to ensure a broad spread of values, prevent blow-up during optimization, and enforce hard constraints (e.g., positivity) where needed.
    
    \item \sfit{(forward Map)} Let $\bbar{F}^{(N)}_{M,\Omega}(\slot;u)$ be the RF posterior mean as in \eqref{eqn:eb_cost} except with the posterior obtained by conditioning on the smaller data set $(X_\Omega, Y_\Omega)\subset (X,Y)$. Similarly, let $\beta^{(u)}_\Omega\in\R^M$ be the coefficients (recall Equation~\ref{eq:coeff_vector_detailed}) of the mean $\bbar{F}^{(N)}_{M,\Omega}(\slot;u)$. Finally, write $G_{M,\Omega}^{(u)}$ for the RF Gram matrix~\eqref{eq:rf_gram} except with the sum now over the set $\Omega$ instead of $[N]$.
    The forward map of the proposed inverse problem framework is then defined for each $u\in\cU$ to be
    
\vspace{-15pt}

\begin{align}\label{eq:pto_rf_hyp}
    \cG(u) \defeq 
    \begin{pmatrix} 
    \bbar{F}^{(N)}_{M,\Omega}(X_{\Omega^\comp};u) \\[0.25em]
    \frac{1}{\sqrt{M}}\norm{\beta^{(u)}_\Omega}\\[0.5em]\sqrt{\log\det\bigl(G_{M,\Omega}^{(u)} + I_M\bigr)}
    \end{pmatrix}.
\end{align}
    \item \sfit{(observable)} The corresponding ``observable'' quantity $z$~\eqref{eq:ip_sample} is
    
\vspace{-15pt}

    \begin{align}
     z \defeq
        \begin{pmatrix}
        Y_{\Omega^\comp} \\ 0 \\ 0
        \end{pmatrix},\
    \end{align}
    which leads to the equation $z=\cG(u)$. The zero entry in $z$ for the log determinant component in $\cG(u)$ is due to the Minkowski determinant theorem. Indeed, this theorem yields the lower bound
    
\vspace{-15pt}

    \begin{align*}
        \det\bigl(G_{M,\Omega}^{(u)} + I_M\bigr)^{1/M}&\geq \det\bigl(G_{M,\Omega}^{(u)}\bigr)^{1/M} + \det(I_M)^{1/M}\\
        &\geq \det(I_M)^{1/M}\\
        & = 1
    \end{align*}
    because $G_{M,\Omega}^{(u)} $ and $I_M$ are positive-semidefinite. This implies that zero is a uniform lower bound for the log determinant term, i.e.,
    
\vspace{-15pt}

    \begin{align}
        \inf_{u\in\cU}\left\{\log \det\bigl(G_{M,\Omega}^{(u)} + I_M\bigr)\right\}\geq 0.
    \end{align}
    \nn{In particular, the square root in the third component of \eqref{eq:pto_rf_hyp} is well-defined.}
    
    \item \sfit{(correlations)} Let $B_{\Sigma}^\comp$ denote the block diagonal matrix with $N - |\Omega|$ identical diagonal blocks containing the noise covariance matrix $\Sigma\in\R^{p\times p}$. The variability $\eta$ of the artificial forward map evaluation $\cG(u)$~\eqref{eq:pto_rf_hyp} about its mean is assumed to be a Gaussian $\normal(0,\Gamma(u))$ with covariance matrix\footnote{The zeros in \eqref{eq:rf_ip_noise} denote rectangular zero matrices of the appropriate size.}
    
\vspace{-15pt}

        \begin{align}\label{eq:rf_ip_noise}
        \Gamma(u) \defeq \Cov_{\mu_u^{\otimes M}}\bigl(\cG(u)\bigr) +  
        \begin{pmatrix}
          B_{\Sigma}^\comp & 0  \\ 0 & I_2
        \end{pmatrix}, 
        \end{align}
    namely, a contribution from the variability within $\cG$ and from the noise perturbing the \nn{labeled} training data. In \eqref{eq:rf_ip_noise}, the covariance operation on $\cG(u)$ is taken with respect to the random features and is conditional on the realization of the noisy training data that underlie the regression problem. The $2$ by $2$ identity matrix in the lower right block is needed to maintain consistency with the ideal empirical Bayes objective~\eqref{eqn:eb_cost_rf} in the large feature limit, as discussed next.
\end{enumerate}

With these four ingredients in hand, we may now directly apply the EKI framework from Subsection~\ref{sec:eki}. Iterating the EKI algorithm with small enough $\Delta t$ towards termination time $T=1$ will minimize the functional \nn{${L}^{(\mathrm{EKI})}_M$ defined by}

\vspace{-15pt}

\begin{align}\label{eq:ekiloss}
    {L}^{(\mathrm{EKI})}_M(u)\defeq \norm[\big]{\Gamma(u)^{-\frac{1}{2}}\bigl(z-\cG(u)\bigr)}^2 - 2\log P(u),
\end{align}
where $P(u)$ is the prior probability density of $u$; \nn{cf.~\eqref{eqn:opt_mle_and_map_gauss}.} This is the weighted observable misfit added to a diminishing contribution from the prior on $u$ as the number of iterations increases. The prior contribution ensures that desired constraints such as positivity hold. The implied EKI objective function \eqref{eq:ekiloss} is related to the ideal empirical Bayes cost~\eqref{eqn:eb_cost_rf} in the infinite feature limit. Indeed, in this limit, the term $\Cov_{\mu_u^{\otimes M}}(\cG(u))$ tends to zero so that for large enough $M$, it holds that

\vspace{-15pt}

\begin{align}\label{eq:ekiloss_largeM}
\begin{split}
{L}^{(\mathrm{EKI})}_M(u)
 &\approx
 \frac{1}{M}\norm[\big]{\beta_\Omega^{(u)}}^2_{\R^M} + \norm[\Big]{(B_\Sigma^\comp)^{-\frac{1}{2}}\bigl(Y_{\Omega^\comp}-\bbar{F}^{(N)}_{M,\Omega} (X_{\Omega^\comp};u)\bigr)}^2 \\
& \qquad\qquad +  \log \det\bigl(G_{M,\Omega}^{(u)} + I_M\bigr) - 2\log P(u).
\end{split}
\end{align}
The \nn{terms on the right hand side of \eqref{eq:ekiloss_largeM} in this large feature limit are structurally similar to those in the cost ${L}^{(\mathrm{EB})}_M$~\eqref{eqn:eb_cost_rf} \nn{when $\Omega=\Omega^\comp = [N]$. In practice, we prefer instead a construction similar to cross-validation: take a fixed disjoint partition $\bigcup_{k=1}^\mathsf{K} \Omega_k = [N]$, where $|\Omega_k| = N/\mathsf{K}$ and $\mathsf{K}$ divides $N$. We define a validation partition for an index $j\in[\mathsf{K}]$ as $\Omega^\comp\defeq \Omega_j$; the remaining indices form the training partition $\Omega \defeq \bigcup_{k\neq j}\Omega_k$}. Repeating this procedure by cycling through $j\in[\mathsf{K}]$, we may incorporate up to $\mathsf{K}$ partitions by stacking for each $j$ the corresponding data $z$ for each validation partition and comparing this to stacked forward map evaluations $\mathcal{G}(u)$ built from the associated training partition, plus additive Gaussian noise with block diagonal covariance matrix having $\Gamma$ as each block. The introduction of these validation partitions improves computational efficiency, which scales with $|\Omega^\comp|\leq N$ instead of $N$. With even two such partitions, we also observe improved convergence of the optimizer and improved accuracy in experiments. We hypothesize that cross-validation can improve the landscape of the objective function but leave further investigation to future work.} Since~\eqref{eq:ekiloss_largeM} arises from a Bayesian inverse problem, it also requires a prior to regularize the inversion, and hence optimization; this contribution appears in ${L}^{(\mathrm{EKI})}_M$ \nn{and in \eqref{eq:ekiloss_largeM}} but not in ${L}^{(\mathrm{EB})}_M$. Although \eqref{eq:ekiloss_largeM} is approximately valid for large numbers of random features, for efficiency purposes we \nn{minimize it} with relatively few features in practice.\footnote{\nn{The effect of varying the number of features during optimization is explored in Appendix \ref{app:n_features}.}}

\subsection{A Practical Feature Distribution Structure}\label{sec:kernel_structure}
In this subsection, we instantiate our general framework with an implementable parametrization of the RF pair $(\varphi,\mu_u)$. 
In all numerical experiments to follow, we employ real vector-valued random Fourier features~\citep{RahRec07} with Gaussian samples and an additional scaling parameter $\varsigma>0$. Specifically, define for each $x\in\cX\subset \R^d$ the $\R^p$-valued feature map 
\begin{equation}\label{eq:rff}
\varphi(x;\theta) \defeq \sqrt{\varsigma}\cos(\Xi x + B),\qw \theta = (\Xi,B) \sim \mathscr{D}
\end{equation}
and $\mathscr{D} \defeq \normal(0,C) \otimes \mathsf{Unif}([0,2\pi]^p)$. The RF distribution $\mathscr{D}$ has high-dimensional hyperparameter $C \in \R^{dp\times dp}$, the covariance operator over the Gaussian-distributed matrix $\Xi$. We always reshape samples of $\Xi$ from $\normal(0,C)$ into a $p \times d$ matrix. To reduce the number of tunable parameters (which is $\cO(d^2p^2))$ and computational effort, one can impose structure on $C$ in various forms, for example, through Cholesky factorizations, diagonal approximations, low-rank parametrizations, or tensor products. We choose a low-rank perturbation representation~\citep{AmbONeSin16_pre} as follows. For some $r\in\N$, take $U\in \R^{dp\times r}$ and symmetric positive definite $S\in \R^{r\times r}$. Then, define

\vspace{-15pt}

\begin{align}\label{eq:rff_cov}
    C \defeq \bigl(I_{dp}+USU^\tp\bigr)\bigl(I_{dp}+USU^\tp\bigr)^\tp.
\end{align}
The task is then to jointly learn the hyperparameters $(\varsigma, U, S)$. We enforce that $S$ be diagonal, leading to a moderate number of total tunable parameters $r(dp+r)+1$. This is advantageous whenever $r\ll dp$. \nn{We refer to the preceding construction as a \emph{nonseparable feature distribution} with rank $r$. We also consider $\Xi \sim \mathsf{MatrixNormal}(0,C_{\mathrm{in}},C_{\mathrm{out}})$~\citep{dutilleul1999mle} and tune the pair $(C_{\mathrm{in}},C_{\mathrm{out}})\in \mathbb{R}^{d\times d}\times \mathbb{R}^{p\times p}$. This pair of covariances can be identified with a larger covariance $C\in \R^{dp\times dp}$ with Kronecker structure. This construction is referred to as a \emph{separable feature distribution} with rank $(r_\mathrm{in},r_\mathrm{out})$ and leads to $r_\mathrm{in}(d+r_\mathrm{in})+r_\mathrm{out}(p+r_\mathrm{out})+1$ total tuneable parameters. A suitable rank is selected to produce emulators with minimal $L^2$ errors on a held-out test set in experiments.}

There are many other options for parametrization beyond (vector-valued) random Fourier features of the form \eqref{eq:rff} and \eqref{eq:rff_cov}. For instance, one could parametrize a possibly sparse precision matrix instead of the Gaussian covariance $C$, or use other efficient representations such as FastFood kernels~\citep{LeSarSmo13,SmoSonWilZic15}. Going beyond Gaussian samples, one can take inspiration from the scalar output $p=1$ case. Here, sampling $\Xi$ from a Student's $t$ distribution leads to a limiting Mat\'ern kernel; this follows from the fact that such densities are Fourier transform pairs. Similar results hold for Cauchy and Laplace distributions. In the $p>1$ case, one can still sample $\Xi$ from such non-Gaussian distributions on the reshaped $\R^{dp}$ space. The main hyperparameters that characterize these distributions are generalized lengthscale matrices $C\in\R^{dp\times dp}$, just as in the Gaussian case. Going beyond the Fourier setting, one approach is to work instead with random neural networks, where the feature map is a hidden neuron of the form $x\mapsto \varrho(\Xi x+B)$ for some activation function $\varrho(\cdot)$. These directions are left to future work.

\subsection{Current Limitations of the Methodology}\label{sec:limitations}
\nn{
We now detail additional practical considerations of the proposed stochastic inversion framework as well as some of its limitations.
First, we highlight that the law of $\cG(u)$ and the prior distributions are assumed to be Gaussian in our formulation. This assumption is not strictly necessary, but it enables straightforward application of ensemble Kalman methods.

Though RFR has admirable scaling, the objective function \eqref{eqn:eb_cost_rf} scales na\"ively as $(Np)^2$. This significantly slows down workflows involving high-dimensional spaces or large data sets. Therefore, controlling the computational cost necessitates further approximations at implementation time. One example, already discussed in Section \ref{sec:opt_to_inv}, is to evaluate the objective on a validation partition of cardinality less than or equal to the total sample size $N$. Another approximation is to avoid computing $u\mapsto \Gamma(u)$ in \eqref{eq:rf_ip_noise} at all $u$ (requiring the covariance of \eqref{eq:pto_rf_hyp} under the random features) by using an offline constant approximation $\Gamma(u) \approx \Gamma(u^\dagger)$ at a fixed $u^\dagger$. In the following numerical results, $u^\dagger$ is taken as the mean of the hyperparameter prior. The effect of improved approximation of $\Gamma(u)$ is not considered in the current paper but remains an interesting direction for future work. 

Finding local optima naturally has dependence on the initial condition. For EKI, this arises from the prior used to create the initial ensemble. There is much debate on prior selection~\citep{BetGelSim17}, particularly in high-dimensional and nonphysical systems. The heuristic adopted in the present work is to choose weakly informative priors over the hyperparameters. We use the same prior structure across all of the forthcoming numerical experiments. Appendix \ref{sec:app_hypprior} contains the details. Moreover, it is tempting to directly optimize the likelihood, i.e., define instead $\cG(u)$ to be equal to the scalar value $ -\log P(z\condbar u)$ and take $z=0$ as the ``observable''. This results in a one-dimensional optimization problem. Unfortunately, we report that this approach incurs large approximation errors in numerical experiments with input or output dimensions of even moderate size. The more costly vector-valued formulation in Subsection~\ref{sec:opt_to_inv} is preferred.
}

\section{Applications}\label{sec:applications}
To illustrate the role of the hyperparameter \nn{tuning}, this section presents different applications  that require the \nn{machine learning} emulator to represent nonlocal features away from the training data. Global sensitivity analysis requires the posterior mean function to be accurate at quasi-random \nn{samples from the} $d$-dimensional cube (Subsection~\ref{sec:ex_sensitivity}); to integrate a Lorenz 63 system, the posterior mean function must progress along, and attract towards, the Lorenz attractor (Subsection~\ref{sec:ex_lorenz}); for accelerated uncertainty quantification, the emulator mean and covariance predictions are used within a sampling method to explore the support of a five-dimensional posterior distribution (Subsection~\ref{sec:ex_cloud}).
In all three applications, \nn{the data consist of a moderate number of observations} that are perturbed by additive Gaussian noise with known covariance \nn{matrix}.

For comparison to the approach developed in the present paper, two other emulators are presented, an \nn{untuned} RF model using hyperparameters from the mean of the EKI initialization and a \nn{tuned} GP from a robust software package. \nn{To emulate smooth functions $\mathbb{R}^d\to \mathbb{R}$ (i.e., $p=1$) with GPs, we use} a radial basis function kernel with scaling parameter, nugget parameter, and learnable lengthscales in each of the $d$ input dimensions. To handle $p>1$ output dimensions, \nn{we use a singular value decomposition (SVD)} in the output space to decorrelate \nn{the coordinates, and then fit and tune} $p$ decorrelated GP regressors in each dimension. \nn{This corresponds to a single vector-valued GP regressor with a diagonal matrix-valued kernel.} Therefore, a modest $d(p+2)$ parameters are found with gradient-based optimizers built into the chosen software packages. In our experiments we report results from \texttt{SciKitLearn.jl v0.7}. We also tried \texttt{GaussianProcesses.jl v0.12.5}, but found hyperparameter optimization to be too brittle to provide consistent comparisons.

\nn{
\paragraph*{Design Choices.}
To support many of the heuristics in the upcoming experiments, we provide supplementary investigations in Appendix~\ref{app:details} for each application. These include details about the EKI configuration, objective function configuration, and convergence plots over different random initializations. Some other specific studies we include are run times and accuracy with respect to input dimension size (Appendix \ref{sec:app_app1}) and numbers of features (Appendix \ref{app:n_features}), and accuracy over different feature distribution ranks and covariance structures in the multi-output setting (Appendix \ref{app:rank} and \ref{sec:app_app3}).
}

\subsection{Global Sensitivity Analysis}\label{sec:ex_sensitivity}
Global sensitivity analysis (GSA) seeks to attribute the variance of model output to its input variables over a given domain \citep{Sal02}. Practitioners are interested in either ranking variables by their direct contribution to the variance, computing $\mathsf{V}_i = \mathbb{V}(\mathbb{E}(f(x) \condbar x_i))/\mathbb{V}(f(x))$, or residuals of the total variance when conditioning on all other variables, $\mathsf{TV}_i = (\mathbb{V}(f(x)) -\mathbb{V}(\mathbb{E}(f(x) \condbar x_{j\neq i})))/\mathbb{V}(f(x))$. Here $x_{j\neq i}$ indicates all variables except $x_i$ and $\mathbb{V}$ \nn{represents the variance operator}.

Two standard functions from the sensitivity analysis literature are used in the demonstration. These are chosen for their analytic representation of $\mathsf{V}_i$ and $\mathsf{TV}_i$. Furthermore, accurate empirical estimates of $\mathsf{V}_i$ and $\mathsf{TV}_i$ can be calculated on the cube using quasi-Monte Carlo methods. Functions are evaluated over points from a low discrepancy Sobol sequence \citep{LevSob99} and are \nn{referred to} as \emph{empirical estimates} in our results.

\subsubsection{Ishigami Function}
First, the Ishigami function \citep{HomIsh90,LevSob99}
\begin{align}
    x\mapsto f(x; a, b) \defeq (1 + bx_3^4)\sin(x_1) + a \sin(x_2)
\end{align}
on $[-\pi,\pi]^3$ with $a=7$ and $ b=0.1$ is considered.
Data is taken as 300 evaluations of the Ishigami function from a length 16000 Sobol sequence, and Gaussian noise is added with variance $0.1^2$. After their hyperparameters are \nn{tuned}, the emulators are fit to the data and used to predict values over the entire Sobol sequence. This is how $\mathsf{V}_i$ and $\mathsf{TV}_i$ are calculated empirically. 

\begin{figure}[tb]
\centering
\includegraphics[width=0.9\textwidth]{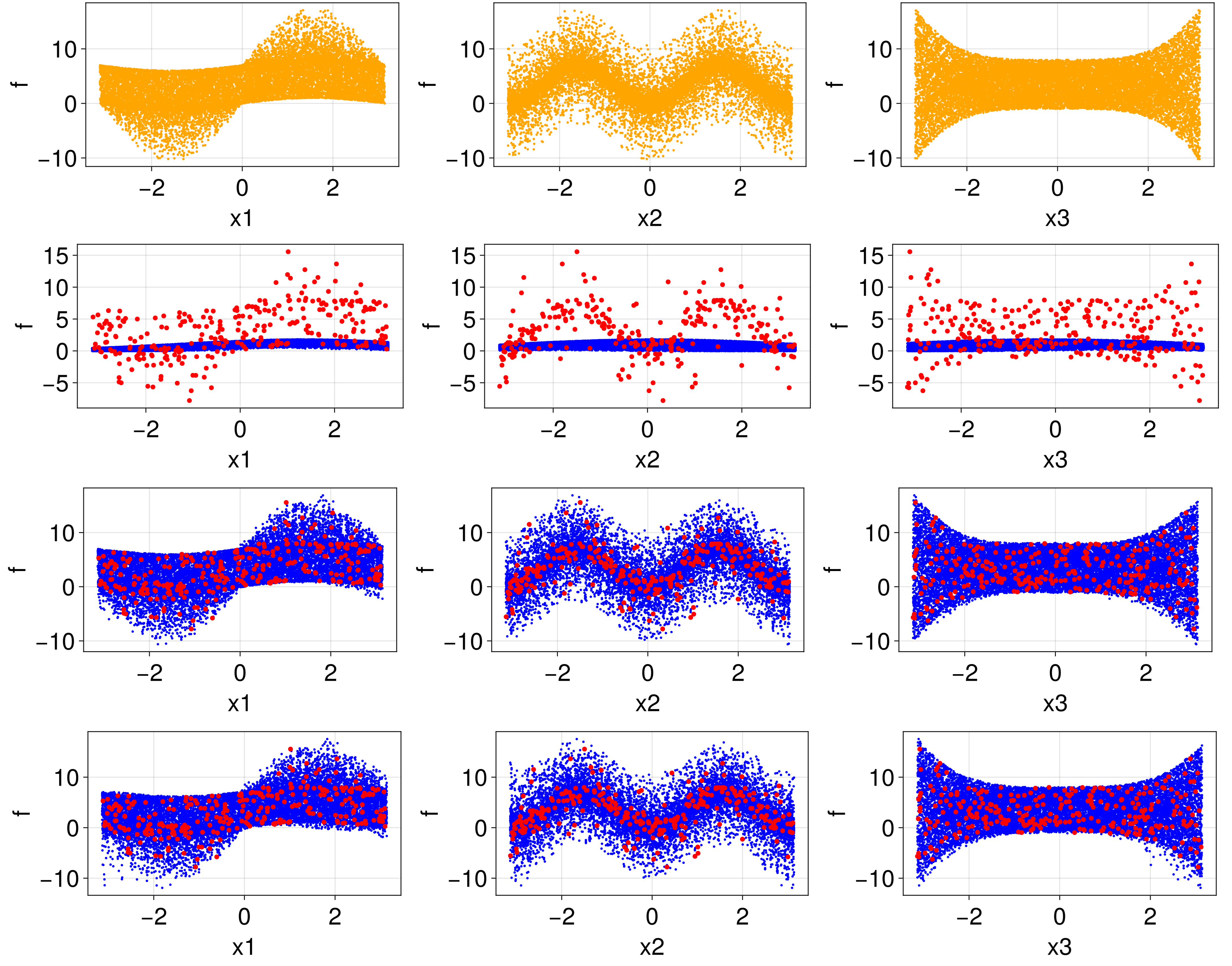}
\caption{Learning the $\mathbb{R}^3 \to \mathbb{R}$ Ishigami function from $300$ noisy samples. In orange, Sobol samples of the true function; in red, $300$ noisy observations; in blue, Sobol samples of the emulated functions. Row 2 is sampled from an RF emulator without hyperparameter \nn{tuning}, Row 3 is sampled from a \nn{tuned} GP emulator, and Row 4 is sampled from the RF approximation \nn{(using a nonseparable feature distribution with full rank covariance)} using our hyperparameter optimizer.} 
\label{fig:ishigami}
\end{figure}

Figure \ref{fig:ishigami} shows the results from a single trial. Row 1 shows slices of the true Ishigami function against each variable evaluated on the Sobol sequence. Rows 2--4 show the noisy data (red) used for learning and the emulator predictions over the Sobol sequence (blue). Row 2 corresponds to the \nn{untuned} RF, Row 3 is the prediction from a \nn{tuned} GP, and Row 4 shows results of the EKI-tuned RF with $500$ features. The successful \nn{trials show that the tuned models fit the true Ishigami function well}.

\begin{table}[tb]
    \centering
    \caption{First order global sensitivity analysis of the Ishigami function. Displayed are the analytic values, the empirical approximation from evaluating the true function at $16000$ Sobol indices, and the empirical approximations obtained from evaluating the \nn{tuned} GP and RF emulators at the Sobol indices. The final column estimates a mean and standard deviation \nn{(in parentheses)} of the emulators from repeating the \nn{stochastic tuning} procedure $20$ times on the same data.}
    \label{tab:ishigami}
    \renewcommand{\arraystretch}{1.2}
    \begin{tabular}{l@{\hspace{10.25mm}}cccc}
        \toprule
        Sobol Index & Analytic & Empirical & \nn{Tuned} GP & \nn{Tuned} RF \\
        \midrule
         $\mathsf{V}_1$ & $0.314$ & $\phantom{-}0.313\phantom{-}$    &  $\phantom{-}0.309\phantom{-}$     & $\phantom{-}0.308\phantom{-} (0.024)$ \\ 
         $\mathsf{V}_2$ & $0.442$ & $\phantom{-}0.442\phantom{-}$    &  $\phantom{-}0.451\phantom{-}$     & $\phantom{-}0.453\phantom{-} (0.023)$ \\ 
         $\mathsf{V}_3$ & $0$     & $-0.006\phantom{-}$ &  $-0.007\phantom{-}$  & $-0.007\phantom{-} (0.005)$ \\
        \midrule
        $\mathsf{TV}_1$ & $0.557$ & $0.562$    &  $0.550$     & $\phantom{-}0.537\phantom{-} (0.026)$ \\
        $\mathsf{TV}_2$ & $0.442$ & $0.442$    &  $0.459$     & $\phantom{-}0.498\phantom{-} (0.038)$ \\ 
        $\mathsf{TV}_3$ & $0.244$ & $0.245$    &  $0.235$     & $\phantom{-}0.223\phantom{-} (0.022)$ \\
        \bottomrule
    \end{tabular}
\end{table}

\nn{To quantify the accuracy of GSA}, Table \ref{tab:ishigami} displays the calculated Sobol indices of the \nn{tuned} emulators. The \nn{tuning} was repeated over $20$ different initial samples to gauge the robustness of randomized \nn{hyperparameter learning} algorithms. Table~\ref{tab:ishigami} shows the mean and standard deviation of these trials. For both $\mathsf{V}_i$ and $\mathsf{TV}_i$, both GP and RF produce close values. \nn{In particular,} RF has variations of at most 5\% of the true sensitivity and the true values always lie within two standard deviations of the RF emulator variability.

Further details about the implementation of this experiment relating to EKI configuration and convergence are provided in Appendix \ref{sec:app_app1}. The algorithm is robust to the data seeding, the number of features used for optimization and prediction, and to various modifiers of the optimization algorithm.

\subsubsection{Sobol G Function}
The Sobol G function \citep{ArcSalSob97,AnnAzzCamRatSalTar10}, defined by
\begin{align}
    x\mapsto G(x;a) = \prod_{i=1}^d \frac{|4x_i - 2|+a_i}{1+x_i},
\end{align}
is another GSA test function that additionally allows for varying the input dimension $d$ and for tuneable interaction between different inputs via the choice of coefficients $\{a_i\}_{i=1}^d$. Here we take $a_i \defeq (i-1)/2 \geq 0$, where small $i$ (and thus small $a_i$) imply larger first-order effects; interactions are primarily present between these variables. 

For this experiment, the results are only presented with a \nn{tuned} RF. \nn{We remark that a tuned GP was accurate in offline tests for dimensions $d=3,6$, and $10$, but timed out at $d=20$ dimensions}. The \nn{untuned} RF performed very poorly and was removed for plot readability. Over input dimensions $d=3,6,10$, and $20$, approximately $2000\cdot d$ Sobol points are used to produce empirical estimates of first-order Sobol indices. \nn{The exact number is determined by the GSA package \href{https://github.com/lrennels/GlobalSensitivityAnalysis.jl}{\texttt{GlobalSensitivityAnalysis.jl v1.2.0}}}. Of these points, $250\cdot d$ (approximately $10\%$ to $15\%$) were used for \nn{tuning} the emulator. The posterior mean evaluation at all Sobol points was then used to calculate empirical Sobol indices. 

\begin{figure}[tb]
    \centering
    \includegraphics[width=0.9\textwidth]{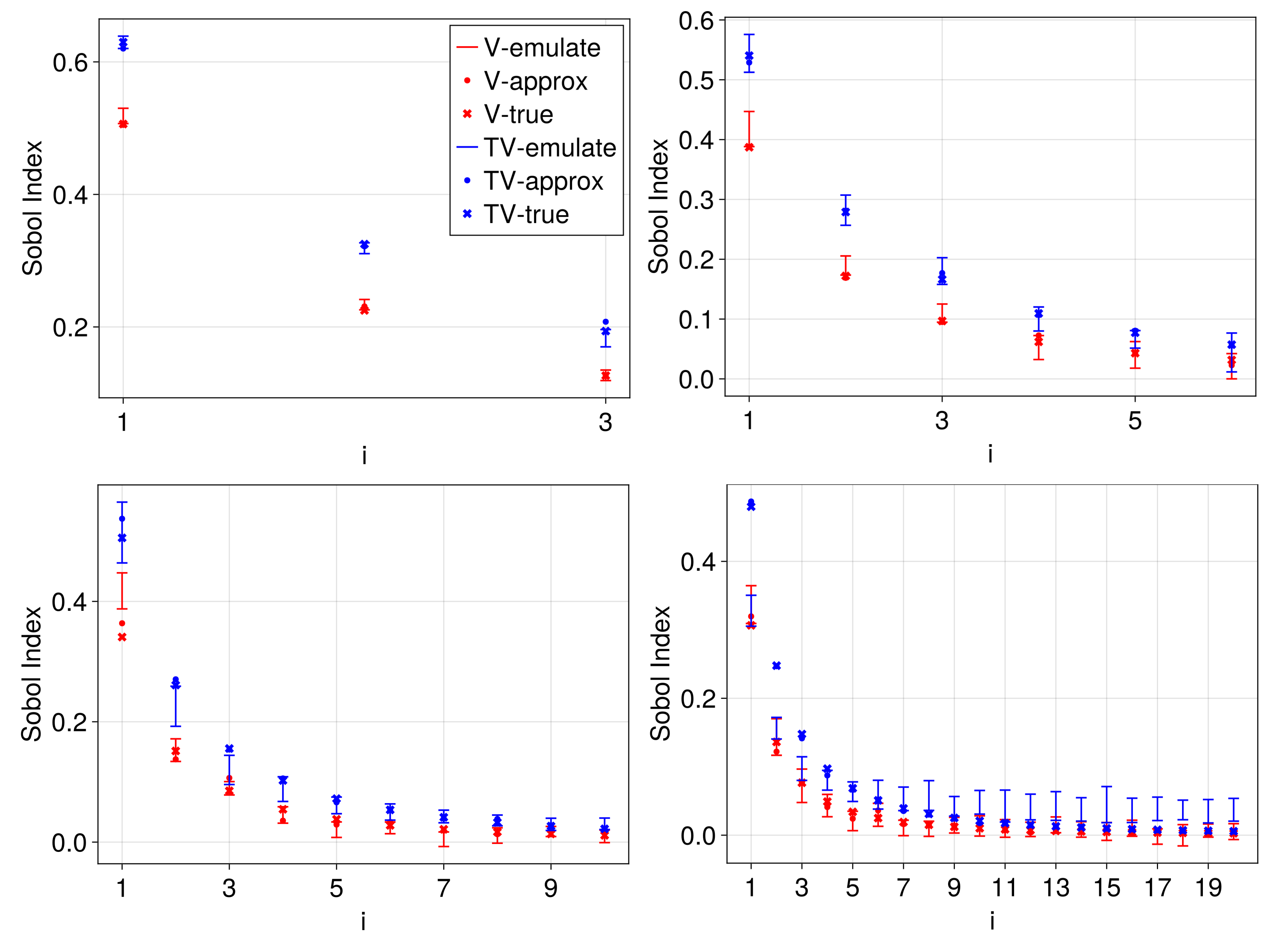} 
    \caption{First order global sensitivity analysis of the Sobol G function with input dimensions $d=3,6,10,20$. \nn{The vertical axis represents the dimensionless values and the horizontal axis represents the index.} Crosses ($\times$) denote the analytic Sobol indices, circles ($\circ$) denote the empirical indices calculated at $1600\cdot d$ points, error bars denote the $(0.05,0.95)$ percentile range of $30$ random feature trials with $250 \cdot d$ \nn{training data} points.}
 \label{fig:sobolG-GSA}
\end{figure}

\begin{figure}[tb]
    \centering
     \includegraphics[width=0.9\textwidth]{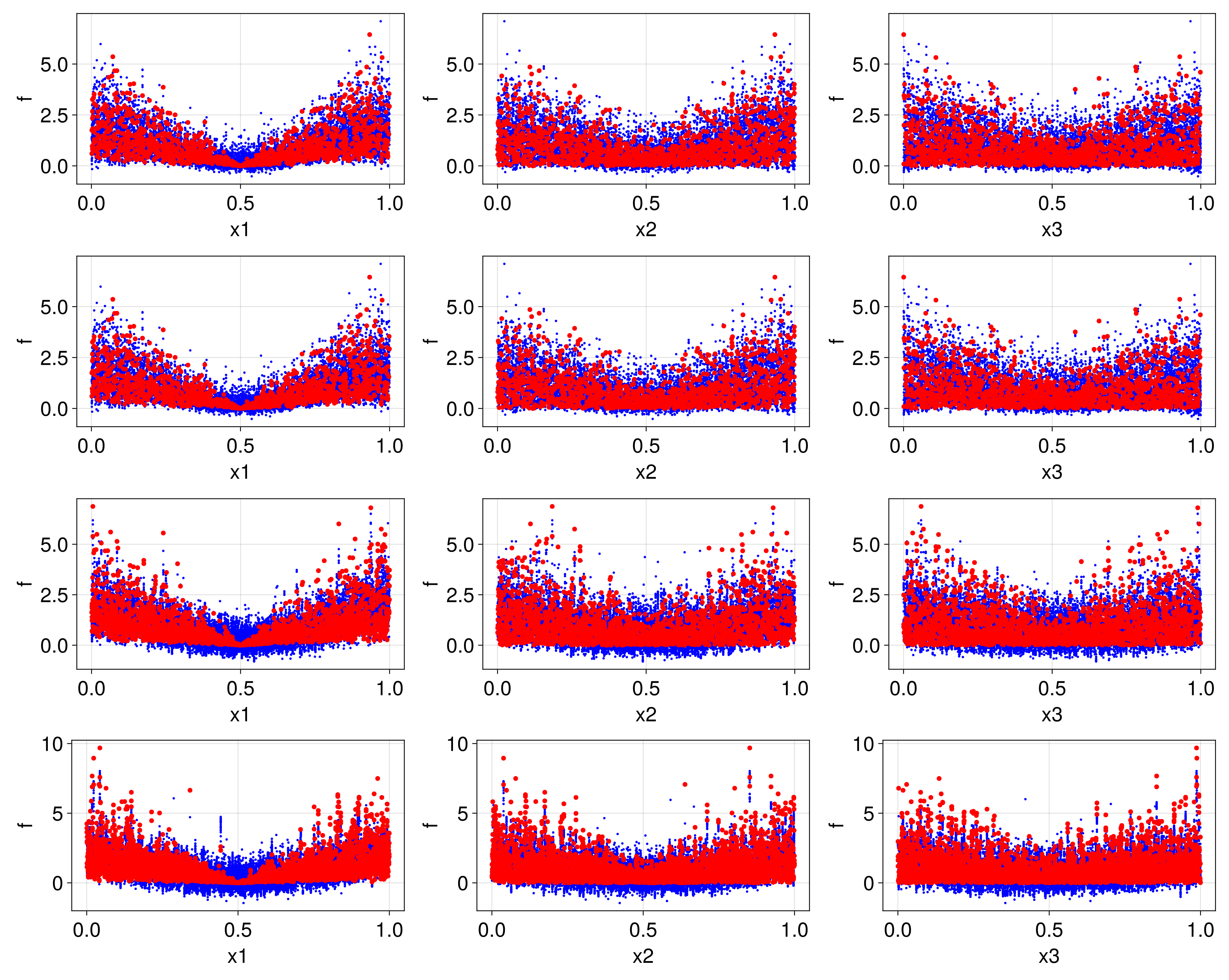}
      \caption{Views of the Sobol G function plotted against the first three variables for $d=3$, $6$, $10$, and $20$ (from top to bottom row). The red points denote the noisy observed data, and the blue represents the tuned RF emulator prediction at all other Sobol points.}
    \label{fig:sobolG-slices}
\end{figure}

Figure \ref{fig:sobolG-GSA} shows the indices $\mathsf{V}_i$ and $\mathsf{TV}_i$ for $i=1,\dots d$ with one panel for each experiment. The emulation was repeated 30 times to produce the spread. Qualitatively, the RF is always able to capture the decaying sensitivity as $i$ increases. Quantitatively, for $d= 3, 6$, and $ 10$, the RF spread well captures the analytic indices, while for $d=20$, only the $\mathsf{V}_i$ index is captured well. Most noticeably for the $\mathsf{TV}_i$ indices, there is a dimensional effect: as more insensitive dimensions are added, the RF emulation begins to overpredict the $\mathsf{TV}_i$ of the least sensitive dimensions and correspondingly overpredicts $\mathsf{TV}_i$ in the most sensitive dimensions. 
Figure \ref{fig:sobolG-slices} provide views in the three most sensitive dimensions for just one of the RF realizations. The emulator has more of a challenge to capture the sharp transition at $x_i = 0.5$ in the most sensitive dimensions as $d$ increases. \nn{Appendix~\ref{sec:app_app1} contains additional results pertaining to emulator performance as the input dimension $d$ and the number of features $M$ are varied.}

\subsection{Lorenz 63 Integrator}\label{sec:ex_lorenz}
The next task is to learn an integrator, or $\mathbb{R}^3 \to \mathbb{R}^3$ next-step map, of the Lorenz 63 dynamical system \citep{Lor63}. The data is produced from observations of a trajectory of the full state that is perturbed with additive Gaussian noise. There is growing use of the Lorenz system (without noise) in the recurrent network literature to validate the learning of deep or recurrent structure from full-state observations \citep{CheFlaSha22,Dud05,FabHerOua18,SchMes19}; here, learning is demonstrated efficiently without deep networks. Since the Lorenz 63 system gives rise to the archetypal chaotic dynamics, the measure of a successful integrator is an accurate representation of the chaotic attractor over long integration times rather than \nn{pointwise accurate fits} to a test trajectory.

The classical parameter choices $(\sigma,\rho,\beta) \defeq (10,28,8/3)$ are chosen for the Lorenz system

\vspace{-15pt}
\nn{
\begin{subequations}
\begin{align}
        \frac{dx}{dt}&=\sigma(y-x)\,,\\
        \frac{dy}{dt}&=x(\rho-z) - y\,,\\
        \frac{dz}{dt}&=xy-\beta z
\end{align}
\end{subequations}}
to produce chaotic dynamics. The state is denoted by $(x(t),y(t),z(t))$ at each time $t$. The true dynamics were integrated with an Euler method and time step of $10^{-2}$. For the sample size, $500$ training data pairs were sourced randomly\footnote{In experiments, sourcing data sequentially from a trajectory gives similar performance and introducing a spin-up time (e.g., learn with data from \nn{time interval} $[10,20]$) decreases performance slightly.} from $[0,20]$. The data pairs therefore comprise inputs $\{(x(t_n),y(t_n),z(t_n)\}$ and (noisy) outputs $\{(x(t_{n+1}),y(t_{n+1}),z(t_{n+1})) + \eta\}$, where $\eta \sim \normal(0,\Sigma)$ \nn{and $\{t_n\}$ is the Euler time discretization}. We emulate these maps with both scalar GPs (requiring decorrelation of the output space) and with vector-valued RFs (native output space).

\begin{figure}[tb]
\centering
\includegraphics[width=\textwidth]{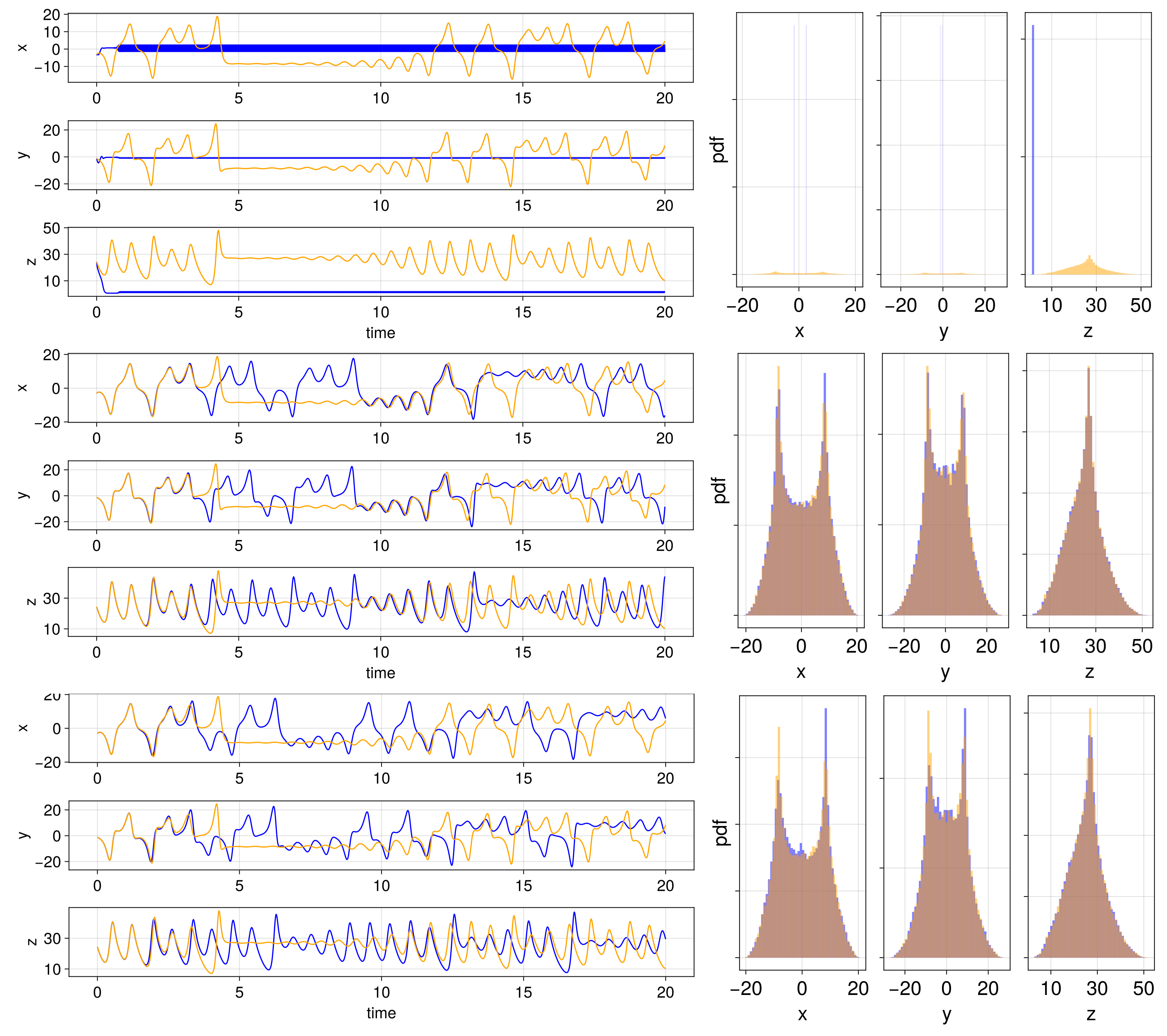}
\caption{Learning a Lorenz 63 integrator from noisy data: \nn{emulated (blue) vs. truth (orange). The first column displays the time evolution of the three state variables $x(t)$, $y(t)$, and $z(t)$ (vertical axis) as a function of time $t$ (horizontal axis) on initial conditions not seen during training. The second column visualizes the marginal probability density functions for each state variable.} Row 1 shows integration with the \nn{untuned} RF without hyperparameter learning. Row 2 shows integration with the GP emulator with 12 total hyperparameters learned, and Row 3 shows integration with RF using a rank-$4$ \nn{nonseparable feature distribution} with 31 total hyperparameters learned. The $500$ data pairs were subjected to observational noise with covariance $\Sigma = 10^{-4}I$.}
\label{fig:L63}
\end{figure}

For validation, we recursively apply the emulated integrators on time interval $[20,1020]$, plotting the trace and emulated attractor over $[20,\nn{40}]$ and the marginal \nn{probability density functions (PDFs)} of the individual \nn{state} variables. The results for $\Sigma = 10^{-4}I$ are shown in Figure \ref{fig:L63}, with the true integrated system in orange and the emulated system in blue. Row 1 shows the fit of an \nn{untuned RF emulator with $M=600$ features}. Row 2 shows the fit of the \nn{tuned} GP emulator, and Row 3 shows the fit of the \nn{tuned} vector-valued RF emulation. Without hyperparameter learning, the integrator failed to reproduce the attractor, while both \nn{the tuned} vector-valued RF and GP methods demonstrate stable integration for long times. The trace plot illustrates a \nn{trajectory} forecast skill until \nn{time} $T=4$ for RF and for GP, and after these \nn{time} horizons the trajectories diverge as expected. Importantly, the emulated integrators continue to remain on the attractor, fitting the long-time PDFs closely. The approach is relatively robust to noise in the observations, as seen in the supplementary Figure~\ref{fig:L63_highnoise}, where the noise of the data is increased to $\Sigma=10^{-2}I$. Here the short term forecasts are still relatively accurate and the dynamics still explore the chaotic attractor. However, the long-time PDFs lose accuracy. Further increases to the noise eventually lead to a catastrophic failure, where emulators were observed to collapse to periodic orbits or fixed points.

\begin{figure}[tb]
\centering
\includegraphics[width=0.8\textwidth]{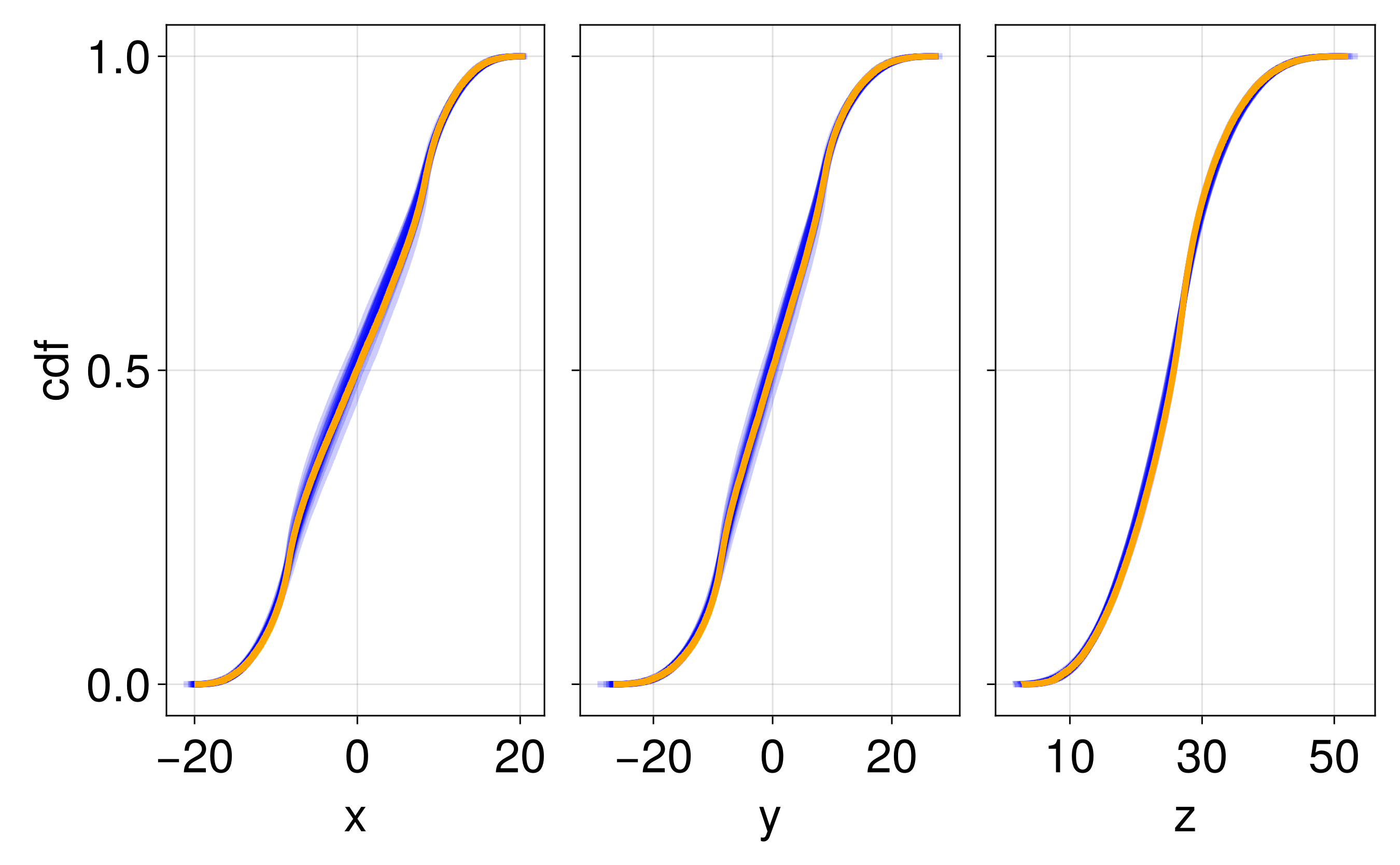}
\caption{Comparison of true marginal empirical CDF (orange) with the CDF of \nn{the RF emulator corresponding to $30$ re-tunings of the hyperparameters (blue) when performing the experiment displayed in Figure \ref{fig:L63}}. The noise covariance is $\Sigma=10^{-4}I$.}
\label{fig:L63_cdfs}
\end{figure}

To inspect the robustness of the optimization methods, Figure \ref{fig:L63_cdfs} shows the difference between the true \nn{cumulative density function (CDF) and emulated CDF corresponding to the tuned RF emulator over 30 independent} runs of the stochastic optimization algorithm. Parameter sweeps also show that the algorithm is robust to the data seeding, to the number of features used for optimization and prediction, and to various modifiers of the optimization algorithm.
Our results are consistent with efficient integrator learning approaches for this chaotic system. \nn{These approaches include} recurrent networks, reservoir computers, or vector auto-regression models \citep{CheFlaSha22} that are often \nn{tuned and fit to} $10^3$ to $10^4$ noiseless points. \nn{In contrast,} feedforward or deep neural networks are frequently trained on more than $10^4$ noiseless points \citep{Dud05,FabHerOua18,SchMes19}. 
Further details about the implementation of this experiment relating to EKI configuration and its convergence are provided in Appendix~\ref{sec:app_app2}. \nn{This appendix also includes supplementary results that explore the effect of feature distribution structure---such as rank and separability---on performance.}

\subsection{Uncertainty Quantification of a Cloud Model}\label{sec:ex_cloud}
The final task is to investigate a parametric uncertainty quantification problem. Here, one has access to a black-box computational simulator \nn{$G$---specifically a parameter-to-data map---that is dependent on several parameters $\theta$, e.g., unknown physical constants. Given a noisy observation $y$, the task is} to estimate the posterior parameter distribution that \nn{explains this measured data}. 
Common challenges in the scientific realization of such problems are that (\emph{i}) the simulator can only be run a relatively few times \nn{(e.g., fewer than $10^3$ times), (\emph{ii})} the parameter-to-data map is likely non-differentiable or noisy, or both, and (\emph{iii}) the parametrizations are not directly observable. One can address these challenges by posing the underlying task as a classical Bayesian inverse problem and using gold-standard sampling methods such as MCMC \citep{Met_etal53,FeaRobShe10}. \nn{Under a Gaussian noise model with covariance matrix $\Gamma$},\footnote{\nn{This matrix $\Gamma$ is inherent to the physical inverse problem and should not be confused with the noise model for the EKI algorithm appearing in \eqref{eq:ekiloss}, which includes a matrix denoted by the same symbol.}} MCMC algorithms draw samples from a density \nn{over $\theta$} proportional to $\exp(-L(\theta,y))$, where the negative log-posterior is \nn{proportional to}
\begin{equation}\label{eq:mcmcloss}
 L(\theta,y)  \defeq \frac{1}{2}\norm[\big]{\Gamma^{-1/2}\bigl(y-G(\theta)\bigr)}^2 +\frac{1}{2} \log\det \Gamma -\log P(\theta).
\end{equation}
In \eqref{eq:mcmcloss}, $P(\theta)$ denotes the probability density function of the prior on $\theta$.
Unfortunately, MCMC sampling algorithms may require $\mathcal{O}(10^3)$ times more model evaluations than those available. Furthermore, non-differentiable and noisy forward maps $G$ may cause rough landscapes that are challenging to sample the posterior from.

One approach to address the challenges of using MCMC is to replace the parameter-to-data forward map with a smooth and fast-to-evaluate surrogate (e.g., a statistical machine learning emulator) and sample using the surrogate instead. For example, when the forward map is random, previous work~\citep{CleGarLanSchStu21,DunGarSchStu21} approximated the forward map by a \nn{tuned GP posterior} $G\sim \mathsf{GP}(G_{\mathrm{ML}},\Gamma_{\mathrm{ML}})$. Then, \eqref{eq:mcmcloss} can be replaced with
\begin{equation}\label{eq:mcmcloss_ml}
 L(\theta,y) \defeq \frac{1}{2}\norm[\big]{\bigl(\Gamma_{\mathrm{ML}}(\theta)\bigr)^{-1/2}\bigl(y-G_{\mathrm{ML}}(\theta)\bigr)}^2 +\frac{1}{2} \log\det \bigl(\Gamma_{\mathrm{ML}}(\theta)\bigr) -\log P(\theta).
\end{equation}
This approximation and the efficient selection of training points around the \nn{regions of high posterior mass are} discussed at length in these studies; the algorithm is known as Calibrate-Emulate-Sample (CES) \citep{CleGarLanSchStu21}. The statistical emulator of the forward map provides a natural smoothing property as it can directly predict the mean of the random process. In the present paper, the goal is to compare the resulting approximate posterior distributions for a scientific inverse problem using GP emulators (GP-CES) and RF emulators (RF-CES). 

The example in this section is taken from the geophysics literature \citep{Lop_eta22}. At its core, the emulation task involves representing a map $G\colon \mathbb{R}^5 \to \mathbb{R}^{50}$. To handle the multi-output problem with GPs, the output space is whitened and $50$ independent $\mathbb{R}^5 \to \mathbb{R}$ maps are \nn{tuned and fitted. The $50$ GPs can be viewed as a single vector-valued GP with a corresponding diagonal matrix-valued kernel.} This approach depends entirely on the existence of the whitening approximation---which assumes a statistical structure on the inputs and outputs---and the training data sample size, while the new proposed approach using vector-valued RFs does not.

In this \nn{scientific application}, the forward map is a one-dimensional vertical column model that represents turbulence and moist convection processes in the atmosphere. \nn{It can also} represent cloud formation. The model is a parametrization of three-dimensional flow and is a type of eddy-diffusivity mass-flux (EDMF) model. It depends on several parameters governing eddy dissipation and diffusion due to turbulence, stability under convection, and the transport of tracers in (entrainment) and out (detrainment) of cloud updrafts. Both the model and experiment setup is detailed at length in the work of \citet{Lop_eta22}. The data is taken to be time-averaged liquid water path (a scalar value measuring the amount of cloud water in the column) from a suite of $50$ three-dimensional cloud-resolving Large Eddy Simulations (LES); each LES simulation represents a different forcing scenario that mimics a range of sites and climates. The observational noise comes from the LES and uses artificial noise inflation to reasonably capture the structural model error. EKI was used to find optimal parameter values. In this experiment, we use an emulator \nn{tuned} with input-output samples \nn{generated by the iterates of the EKI algorithm} within CES to approximate the full posterior distribution.

\begin{figure}[tb]
\centering
\includegraphics[width=0.7\textwidth]{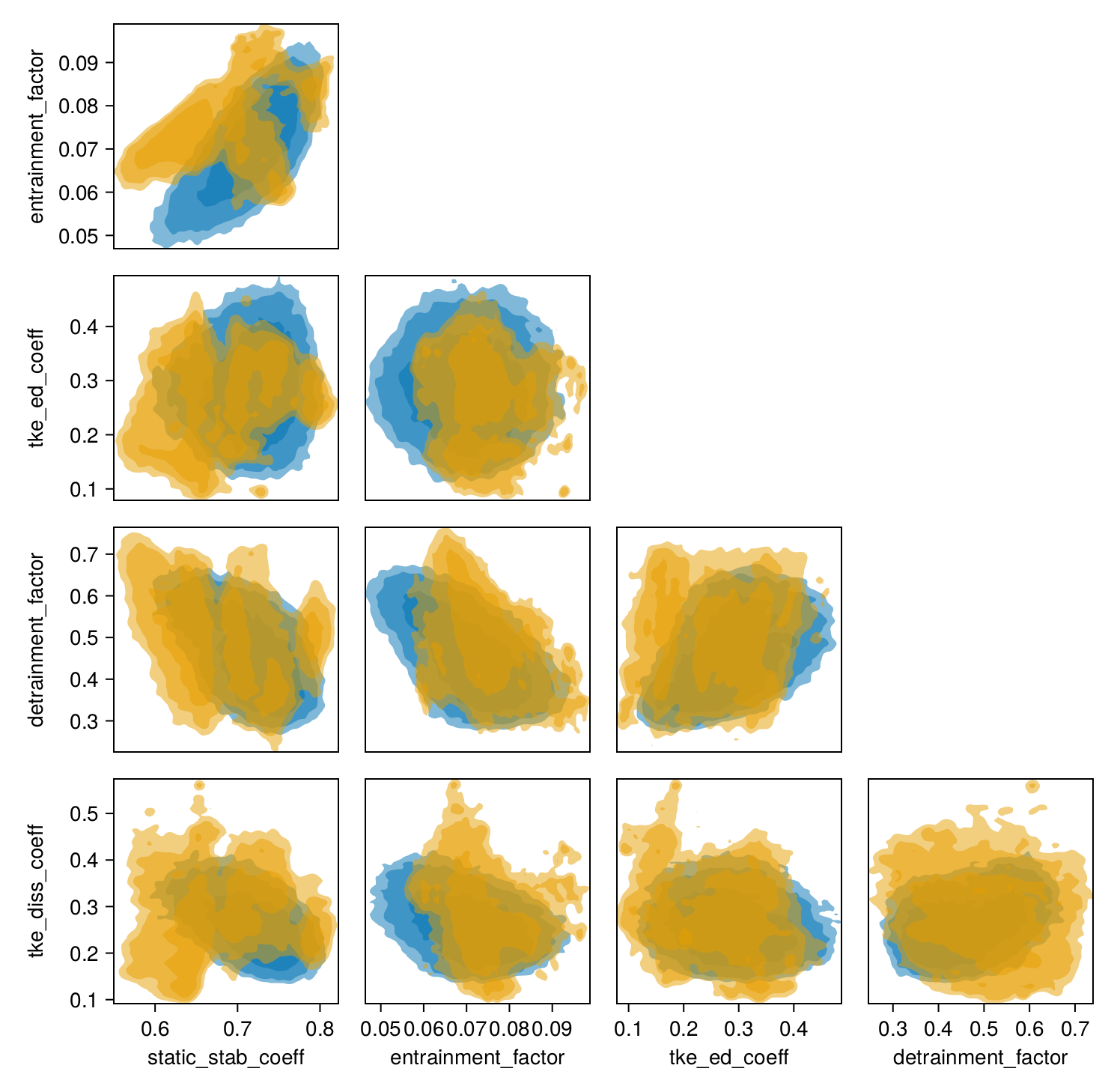}
\caption{Application of CES to the EDMF problem. Displayed are corner-plots showing slices through the joint distribution of five physically meaningful parameters calibrated to data. Orange (blue) contours: the \nn{one, two, and three standard deviation} density of the posterior distribution obtained using GP-CES (RF-CES) after \nn{hyperparameter tuning}.}
\label{fig:EDMF_post}
\end{figure}

Figure~\ref{fig:EDMF_post} displays pairwise slices through the five-dimensional posterior distribution for the geophysical problem. Visualized are the final $10^5$ steps of MCMC using the likelihood calculated from \eqref{eq:mcmcloss_ml}. In orange contours, the GP-CES algorithm is used. In blue contours, the RF-CES algorithm is used. Reasonable qualitative agreement is shown across the support of the \nn{approximate posterior obtained from either approach. However, the GP-CES distribution contain some fine-scale artifacts, while the RF-CES distribution appears smoother overall.} Such artifacts appear to be due to learning small lengthscales in the GP hyperparameter \nn{tuning} algorithm rather than physically meaningful features in the posterior distribution. The GP-CES posterior was also slower to converge; four times \nn{as many samples of burn-in were taken for GP-CES compared to those taken for} RF-CES. Likewise, pairwise correlation between several variables such as entrainment (\texttt{entrainment\_factor}) and the stability coefficient (\texttt{static\_stab\_coeff}) are more clearly seen in the vector-valued RF setting.

Further details of the implementation of this experiment relating to EKI configuration and convergence are provided in Appendix \ref{sec:app_app3}. \nn{In that appendix, we also show numerical results using an untuned RF emulator and a tuned RF emulator with a separable feature distribution. One caveat pertaining to the results of the present section is the finding that the structure of the feature distribution plays a significant role in the shape of the posterior distribution (as does the choice of covariance kernel for GP). The design choices (e.g., covariance structure and rank) we selected were based on $L^2$ errors of the tuned predictors evaluated on a test set, as reported in Table \ref{tab:perf-edmf}. However, these forward map errors were not as reliable at predicting the skill in sampling the posterior distribution in this experiment, as suggested by the supplementary Figure~\ref{fig:app3_uq_prior_and_sep}.
}

\section{Conclusion}\label{sec:conc}
Ensemble Kalman inversion (EKI) and its variants have proven \nn{to be easily implementable and versatile for efficiently solving optimization problems with stochastic objective functions. Such problems arise when tuning hyperparameters in randomized algorithms, and specifically in the model selection problem found in randomized machine learning approaches.} This paper explored a concrete implementation applied to random feature regression for function emulation. The main developments in this work were threefold. First, by viewing random feature emulators as Gaussian processes, the paper used empirical Bayesian arguments to construct an objective function for hyperparameter estimation. Second, this objective function was adapted to an inverse problem framework compatible with EKI. Third, \nn{the work provided structured choices of feature distributions which, when tuned with EKI}, produced emulators that performed comparably to Gaussian processes and displayed robustness to overfitting and feature selection.
% summary of results

\subsection{Discussion}
Random features can offer \nn{satisfactory} approximation with fewer \nn{structural} assumptions and computational requirements than Gaussian processes. As illustrated in Section \ref{sec:ex_cloud}, vector-valued random feature emulators were able to produce qualitatively similar posterior distributions without relying on diagonalization \nn{in the output space---though the framework can take advantage of diagonal structure} when known. As with Gaussian processes, the choice of feature map and distribution is important. In \nn{this paper, the random Fourier features with feature distributions} parametrized by low-rank perturbations (Subsection~\ref{sec:kernel_structure}) were effective across all test cases for a suitably large rank. Further investigation of sensible \nn{feature distribution} selection is left to future work.% benefits of trained RF models

\nn{Performance of random feature regression depends on balancing expressiveness with efficiency in several areas. One must select an appropriate feature distribution family that is both flexible and parameter-efficient for large input and output dimensions. One must propose an optimizer and objective function in order to tune the free parameters of this family efficiently and robustly. The proposed feature distribution families in Subsection~\ref{sec:kernel_structure} were shown to be sufficiently flexible for a wide range of applications, but required a rank that had to be selected offline.
The objective function \eqref{eq:ekiloss} contains a data misfit that scales with $Np$ and the noise factor $\Gamma$ with $(Np)^2$, where $N$ is the size of the training data set and $p$ the output dimension. The proposed adjustment in this paper is to use a cross-validation-inspired partitioning and an offline estimation of a constant $\Gamma$ to reduce these costs. Such changes afforded a feasible online cost of the objective function in our experiments, but we expect that a better balance of accuracy with efficiency can be gained through more principled approximation of $\Gamma$. The proposed EKI optimizer relies on selection of an ensemble size and a prior distribution. A success of EKI is that the ensemble size and parameter dimension are in fact only weakly coupled and newer variants use sampling error correction methods to decouple them further \citep{Morzfeld_etal24}. Such scalability properties are inherited from the ensemble Kalman filtering literature, where assimilators must routinely estimate states of size $\mathcal{O}(10^9)$ with an ensemble size of $\mathcal{O}(10)$. Sensible choice of the prior distribution, here taken to be weakly informative independent distributions in each dimension, is still an open question. Since the proposed framework learns the hyperparameters of the feature distribution, the number of random feature samples $M$ is a free parameter. In practice, it is often advantageous to take $M$ smaller for hyperparameter learning (where accuracy is less important) and take $M$ larger for prediction tasks (where accuracy is the goal). To explore the practical significance of such benefits, comparison with other fixed-feature approaches is an interesting direction for further work, as is elucidating the relationship to large scale hyperparameter transfer methods~\citep{yang2020feature,yang2022tensor}}

\nn{Along related lines, a fruitful area for future development is the design of alternative objective functions for hyperparameter tuning that further improve optimizer performance and accelerate convergence. Objective} functions not motivated from Bayesian arguments exist for kernel methods and are based on cross-validation \citep{OwhYoo19,chen2021consistency,naslidnyk2023comparing}. Such objective functions have proven successful in derivative-based settings for deterministic kernels and could also be recast for random objective functions \nn{such as those required in the present derivative-free framework. The development of theoretical guarantees, like those in \citet{chen2021consistency}, is also of great importance.} Furthermore, since EKI solves an inverse problem, misfits appearing in the \nn{objective function need not be based on directly observed input-output pairs}. This opens the door to \nn{tuning hyperparameters from \emph{indirect} data. Learning from indirect data is useful in many applications where one cannot directly observe the outputs of the model. For example, embedding a machine learning model into the right hand side of a dynamical system will often require online training for stability and performance, yet the input and output of the model is not observable} \citep{AleHasPah24,LevSchStuWu24}.%discussion about objective function

\subsection{Outlook}
\nn{Broadly, the present paper} intends to act as a case study \nn{within} a more general framework for learning distributions that are sampled in randomized algorithms. \nn{Besides the random features studied here, other} machine learning tools that follow a randomization approach \citep{ScaWan17} include Nystr\"{o}m methods \citep{WilSee00,SunZhaZhu15,meanti2022efficient} and reservoir computers \citep{gauthier2018reservoir,BarBolGauGri21,griffith2019forecasting}. Common patterns to tune \nn{the hyperparameters of} such methods often follow the \nn{existing literature on} random features: employing manual calibration, fitting to \nn{fixed realizations of random variables, and restricting to a narrow choice of algorithm- and problem-specific hyperparameters. The proposed framework is a principled and automated alternate strategy.} Moreover, these fields have witnessed other Bayesian optimization methods learn low-dimensional (e.g., fewer than $10$) hyperparameters effectively \citep{griffith2019forecasting, MagRac21, AntBruMarRon23}. \nn{Since EKI is highly scalable when compared with such Gaussian process-based optimizers, it may enable the learning of much higher-dimensional hyperparameters. We remark that recent work also suggests Bayesian optimization methods may be scaled to high input and output spaces \citep{hvarfner2024vanilla,xu2024standard,maddox2021bayesian}.

A parallel literature uses derivative-free evolutionary algorithms to search for deterministic network architectures in a process termed neural evolution and claims superiority over grid search and random search \citep{bergstra2012random} over discrete domains \citep{Rea_etal17,CluLehMiiSta19,BasJiPanXiaYan20_pre}. Such algorithms are also applied at times to continuum hyperparameter domains in low dimensions. All such methods are susceptible to the curse of dimensionality, however. Future work that systematically compares the effectiveness of EKI against Bayesian optimizers, evolutionary algorithms, and random search would be of great value to the field and more decisively determine whether EKI is better suited for higher-dimensional hyperparameters. Altogether, the methodological framework that the present paper develops lays the groundwork for novel automated approaches to hyperparameter calibration on different scales than the existing literature, while simultaneously providing robust and user-friendly software.}

\section*{Code Availability}
The code and examples for this work are available in three open-source software packages written in the Julia programming language. \href{https://github.com/CliMA/RandomFeatures.jl}{\texttt{RandomFeatures.jl v0.3.4}} contains a range of different random feature families, \href{https://github.com/CliMA/EnsembleKalmanProcesses.jl}{\texttt{EnsembleKalmanProcesses.jl v2.0.1}} contains a range of EKI variants and utilities for creating and solving inverse problems, and \href{https://github.com/CliMA/CalibrateEmulateSample.jl}{\texttt{CalibrateEmulateSample.jl v0.6}} contains the code for the application problems presented in this paper.

% Acknowledgments
\acks{\nn{ORAD is supported by Schmidt Sciences, LLC, the National Science Foundation (NSF) under award number AGS-1835860, and the Office of Naval Research (ONR) under award number N00014-23-1-2654.
NHN acknowledges support from NSF award number DMS-2402036, the NSF Graduate
Research Fellowship Program under award number DGE-1745301, the Amazon/Caltech
AI4Science Fellowship, the Air Force Office of Scientific Research under
MURI award number FA9550-20-1-0358 (Machine Learning and Physics-Based Modeling and
Simulation), and the Department of Defense
Vannevar Bush Faculty Fellowship held by Andrew M. Stuart under ONR award
number N00014-22-1-2790. The computations presented in this paper were partially conducted on the Resnick High Performance Computing Center, a facility supported by the Resnick Sustainability Institute at the California Institute of Technology. The authors are grateful to Daniel J. Gauthier for helpful comments on a previous version of the paper and to the two anonymous referees for their valuable feedback.}
}

\appendix

\section{Additional Numerical Experiment Details}\label{app:details}
\nn{In this appendix, we add some additional notes about the experimental configurations to facilitate reproducibility. Appendix~\ref{sec:app_hypprior} discusses hyperparameter priors for the chosen feature distribution parametrizations, while the settings for the EKI algorithm are collected in Appendix~\ref{app:eki_config}. The applications in Section~\ref{sec:applications} were chosen because they require satisfactory performance off of training data. In this appendix, we supplement these scientific applications with auxiliary investigations into the heuristics of running the proposed hyperparameter tuning algorithm. Summarizing the additional results found in Appendices \ref{sec:app_app1}, \ref{sec:app_app2}, and \ref{sec:app_app3}, we find that few features---fewer than training data set size---are sufficient to obtain high-performing tuned hyperparameter, and the structure and rank of the feature distributions are important but can be selected with using a held-out validation data set.}

Finally, we note that all numerical experiments are reproducible at 
\begin{center}
   \href{https://github.com/CliMA/CalibrateEmulateSample.jl}{\texttt{https://github.com/CliMA/CalibrateEmulateSample.jl}}\,.
\end{center}
Users may interact with this code base as a playground to understand our EKI-based algorithm and how its various setup parameters influence performance \nn{in the context of emulator training. Documentation can be found here:
\begin{center}
\href{https://clima.github.io/CalibrateEmulateSample.jl/dev/}{\texttt{https://clima.github.io/CalibrateEmulateSample.jl/dev/}}\,.
\end{center}
Readers should navigate to \texttt{Examples $\to$ Emulator testing}.}
 
\subsection{Hyperparameter Priors}\label{sec:app_hypprior}
To apply EKI, prior probability distributions must be placed on learnable parameters. From a practical perspective, priors serve as a method of introducing information, such as enforcing bounds and setting the initial ensemble span and correlation structure. In particular, for linear problems, the standard EKI update will be confined to the span of the initial ensemble. \nn{Thus, prior specification is a crucial component part of the framework.}

We choose to use a Gaussian prior distribution due to its relationship with a squared-exponential kernel \citep{RahRec07} and find that performance of the algorithm depends strongly on the choice of the prior's covariance structure. Desiring a feature distribution that is flexible yet scales better than the na\"ive $\mathcal{O}((d + p)^2)$ parameters of a Cholesky factored covariance, the low-rank perturbation structure from Section \ref{sec:kernel_structure} was used in all applications:

\vspace{-15pt}

\begin{align*}
\varphi(x;(\varsigma,U, S)) &= \sqrt{\varsigma}\cos(\Xi x + B), \qw\\
(\Xi, B) &\sim \normal\bigl(0,(I + U S U^{\tp})(I + U S U^{\tp})^{\tp}\bigr) \otimes \mathsf{Unif}([0,2\pi]^p).
\end{align*}

\nn{We also compare this class of nonseparable feature distributions with separable feature distributions in multi-output experiments. The class of separable distributions we chose factors both the input and output space, leading to
\begin{align*}
\varphi(x;(\varsigma, V_1,T_1, V_2,T_2)) &= \sqrt{\varsigma}\cos(\Xi x  + B), \qw\\
(\Xi, B) &\sim \mathrm{MatrixNormal}\bigl(0,C_{\mathrm{in}}, C_{\mathrm{out}}\bigr)\otimes \mathsf{Unif}([0,2\pi]^p)\,,\\
C_{\mathrm{in}}&\defeq (I + V_1 T_1 V_1^{\tp})(I + V_1 T_1 V_1^{\tp})^{\tp}\,, \qa\\
C_{\mathrm{out}}&\defeq (I + V_2 T_2 V_2^{\tp})(I + V_2 T_2 V_2^{\tp})^{\tp}.
\end{align*}
This separable kernel places a covariance structure on inputs (subscript $1$) and outputs (subscript $2$) independently and is more restrictive than the nonseparable distribution. However, the separable case has fewer hyperparameters. In the preceding displays, the matrices $S$, $T_1$, and $T_2$ are all diagonal.
}

To enforce positivity of $\varsigma$ and the entries of the diagonal matrices, we endow them with independent log-normal prior distributions with $99\%$ of prior support covering the interval $(10^{-3},10^3)$. For the entries $U_{ij}$, we took independent Gaussian prior distributions with $99\%$ of the prior support covering $(-300,300)$. In the different experiments, the dependence on these prior ranges was quite weak. They were observed however to influence the speed of algorithm convergence when the prior span was overly concentrated or overly spread by several orders of magnitude. 

\subsection{Common EKI Configuration}\label{app:eki_config}
Besides the prior, we also detail the common EKI configuration parameters used across all three application experiments. 
In the experiments, we chose to use the adaptive learning rate scheduler from the work of \citet{IglYan21}. The adaptive value of step size $\Delta t_{n+1}$ provides a large step when bounding Jeffrey's divergence between the distributions represented by the ensemble at times $t_n$ and $t_{n+1}$. This prevents accumulation of errors due to ill-conditioned updates with large $\Delta t$. A small amount of additive white noise (i.e., inflation) was added to the particles at each iteration to prevent degeneracy in the ensemble exploration. \nn{Additionally, the optimizer was more robust under multiple validation partitions $\Omega^\comp$ as described in Subsection~\ref{sec:opt_to_inv}. We chose to stack two partitions of the data during training and took $\mathsf{K}/N = 0.2$ for all experiments except in Subsection \ref{sec:ex_cloud}, where we set $\mathsf{K}/N=0.1$}.
The input and output data was normalized before training. In particular, a whitening matrix was applied so that the input samples had empirical mean zero and unit empirical variance. The $\Gamma=\Gamma(u)$ in Subsection \ref{sec:opt_to_inv} was always estimated at a fixed value of $u=u^\dagger$ taken as the mean of the prior. The output data was also whitened so that the variability matrix $\Gamma$ had identity covariance over the output data. An optimal linear shrinkage estimator was used to ensure a well-conditioned matrix \citep{LedWol04}.

\subsection{Details for Section~\ref{sec:ex_sensitivity}:~\nameref*{sec:ex_sensitivity}}\label{sec:app_app1}
\begin{figure}[tbp]
\centering
\includegraphics[width=0.7\textwidth]{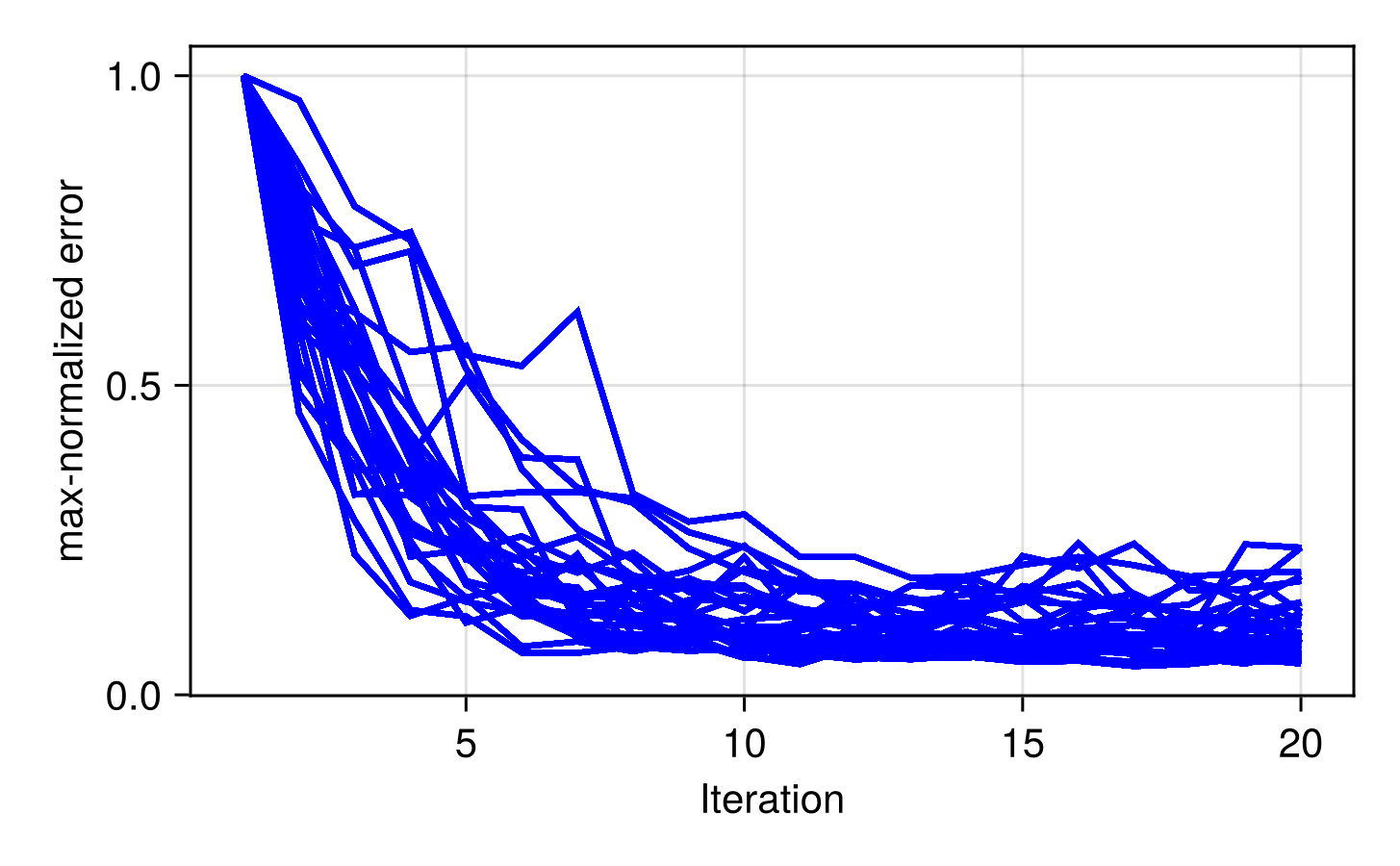}  
\caption{Convergence of the error $n_{\mathrm{iter}}^{-1}\sum_{n=1}^{n_{\mathrm{iter}}} \|\Gamma^{-1/2}(z_n - \cG(u_n))\|^2$ for $20$ different \nn{tuning} experiments with scalar-valued random features \nn{used to learn the Ishigami function for global sensitivity analysis. The initial objective value is normalized to one in the plot}. For this experiment, convergence stagnates at around eight ensemble iterations.}
\label{fig:app1_conv}
\end{figure}

To \nn{tune} the hyperparameters in the Ishigami experiment, the ensemble size was taken to be $J=30$ for the 13 parameters $\#\{\varsigma, \{U_{ij}\}_{ij}, \{D_{ii}\}_i\} = (1, 9, 3)$. The number of features used while optimizing the hyperparameters was taken to be 150. In Figure \ref{fig:app1_conv}, the convergence is plotted over 20 iterations and under 20 restarts of the algorithm with different random samplings from the prior.

To \nn{tune} the hyperparameters in the G-function experiment of dimension $d$, the ensemble size was taken to be $200$ across all dimensional experiments. For the parameters of the rank-$r$ feature distribution, $\#\{\varsigma, \{U_{ij}\}_{ij}, \{S_{ii}\}_i\} = (1, 20r, r)$, we took rank $r=\min(10,d)$. The number of features used while optimizing the hyperparameters was taken to be $M=300$. The optimization was run for 20 iterations with a fixed ensemble size of 200 over all experiments. We do not plot the convergence for all experiments for brevity, but report that the normalized objective value was reduced to \nn{under $10^{-1}$} over the iterations. 

\nn{Table~\ref{tab:gfunc_runtime} displays timings of the tuning and fitting process for the Sobol G-Function against different input dimensions. Both GP and RF methods are compared. Since the algorithms are not reaching the same convergence criteria and performance may depend heavily on implementation details of numerical linear algebra subroutines under-the-hood, such timings are qualitative. However, these results provide evidence of the poor scaling of GP hyperparameter tuning even in moderate dimensions and even with a fixed number of input data points. On the other hand, EKI scales linearly and is also able to take advantage of multithreading---though some overheads are seen here.

\begin{table}[tbp]
    \centering
    \caption{Mean time to train (tune and fit) the machine learning algorithms, averaged over $30$ repeated experiments for learning the Sobol G-function for global sensitivity analysis. Note this comparison runs the default convergence criteria for L-BFGS-B from \texttt{SciKitLearn.jl} and runs EKI for $20$ iterations using $200$ ensemble members. $^\star$The final GP training was canceled after 10 hours. The memory for both tools in serial was 2GB, while EKI in parallel used approximately 6GB.}
    \begin{tabular}{c c c c}   \toprule
    $d$  & GP (L-BFGS-B) & RF (EKI) -- Serial & RF (EKI) -- $8$ Threads \\ \midrule
     3    & 16s & 46s &12s \\ 
     6    & 309s & 65s & 23s  \\ 
     10    & 2653s & 134s & 38s \\ 
     20   & $>$10h$^\star$ & 270s & 67s  \\ \bottomrule
    \end{tabular}    
    \label{tab:gfunc_runtime}
\end{table}
}

\subsubsection{Performance Dependence on Number of Features}\label{app:n_features}
\nn{A useful perk that the proposed hyperparameter tuning framework for RFs can utilize is a decoupling of the number of features $M$ required during tuning from the number of features $M'\neq M$ required during prediction. The timings and performance of the algorithm for the six-dimensional Sobol G-function over varying numbers of features $M$ used in tuning is displayed in Table~\ref{tab:sobol-timings}. The error here represents the $L^2$ difference between the predicted RF mean (with $M'=1000$ features) and the noiseless data, averaged either over the fit inputs (training error) or the other sampling points (test error). In either case, it is observed that increasing the number of features leads to a scaling in time between linear and quadratic, and a reduction of both the training and test error. However, there is clear stagnation in performance with respect to both training and test error when the algorithm exceeds $M=100$ features. This is potentially due to sample error domination or optimization effects.}

\begin{table}[tbp]
\centering
\caption{\nn{The tuned performance for the six-dimensional Sobol G-function emulators as different numbers of features $M$ are taken during tuning. The training and test errors are the average $L^2$ difference between the tuned RF mean (created always with $M'=1000$ features) and noiseless data at training and test points. The final row corresponds to the GP emulator. The time taken for the tuning process is also shown.}}
\begin{tabular}{c c c c}\toprule
    $M$ (When Tuning) & Optimizer Time (Serial) & Training Error & Test Error \\ \midrule
    25 &  7s & 0.014420 & 0.004687\\
 50& 13s& 0.011280 &  0.004069\\
 100 &30s&0.007872&  0.003643 \\
 200 &72s&0.007720&  0.003797\\
 400 &204s&0.006297&   0.003535\\
 800 &700s&0.005756&  0.003660\\ \midrule
 $\infty$ (GP Limit) &  320s & 0.000358 & 0.002827 \\
 
\bottomrule
\end{tabular}
\label{tab:sobol-timings}
\end{table}

% Using a 6D $\to$ 1D regression. Comparing timings and performance with 300 and 600 points of training data, and using the theoretical loss (no cross-validation). We observe the following performance for timing and we compute the $L^2$ difference of the mean to the training data on the training set.
% \begin{table}
% \centering 
% 300 data points, number of features for prediction 800, no CV

% \begin{tabular}{c c c}\toprule
%     number features & optimizer time & training error \\ \midrule
%     25 &  1.6742265721999998 & 1.491967218600323e-7 \\
%     50 & 5.1850888038 & 3.936278983633467e-8\\
%     100 & 12.827569303 & 1.3720619405321185e-8\\
%     200 & 25.722240145199997 & 1.0934971562636165e-8 \\
%     400\orad{ $m>np$} & 73.491480519 & \orad{4.051545516540946e-6} \\
%     800\orad{ $m>np$} &  215.66989169540003 & \orad{7.3294064107778e-6}\\ \bottomrule
% \end{tabular}

% \vspace{10pt}

% 600 data points, number of features for prediction 800, no CV

% \begin{tabular}{c c c}\toprule
%     number features & optimizer time & training error \\ \midrule
%     25 &9.8534599164 & 2.781636055311297e-6
% \\
%     50 &14.755430810000002 &  2.3787867197450326e-7
% \\
%     100 & 27.8037377984 &  6.632769103320949e-8
% \\
%     200 & 48.882027832 &  5.468744273346613e-8
%  \\
%     400 & 125.0634310838 & 8.721860686930762e-8
%  \\
%     800 \orad{$m>np$} & 372.28963181940003& \orad{1.7121867883003227e-5}\\ \bottomrule
% \end{tabular}
% \end{table}

\subsection{Details for Section~\ref{sec:ex_lorenz}:~\nameref*{sec:ex_lorenz}}\label{sec:app_app2}
\begin{figure}[tbp]
\centering
\includegraphics[width=0.7\textwidth]{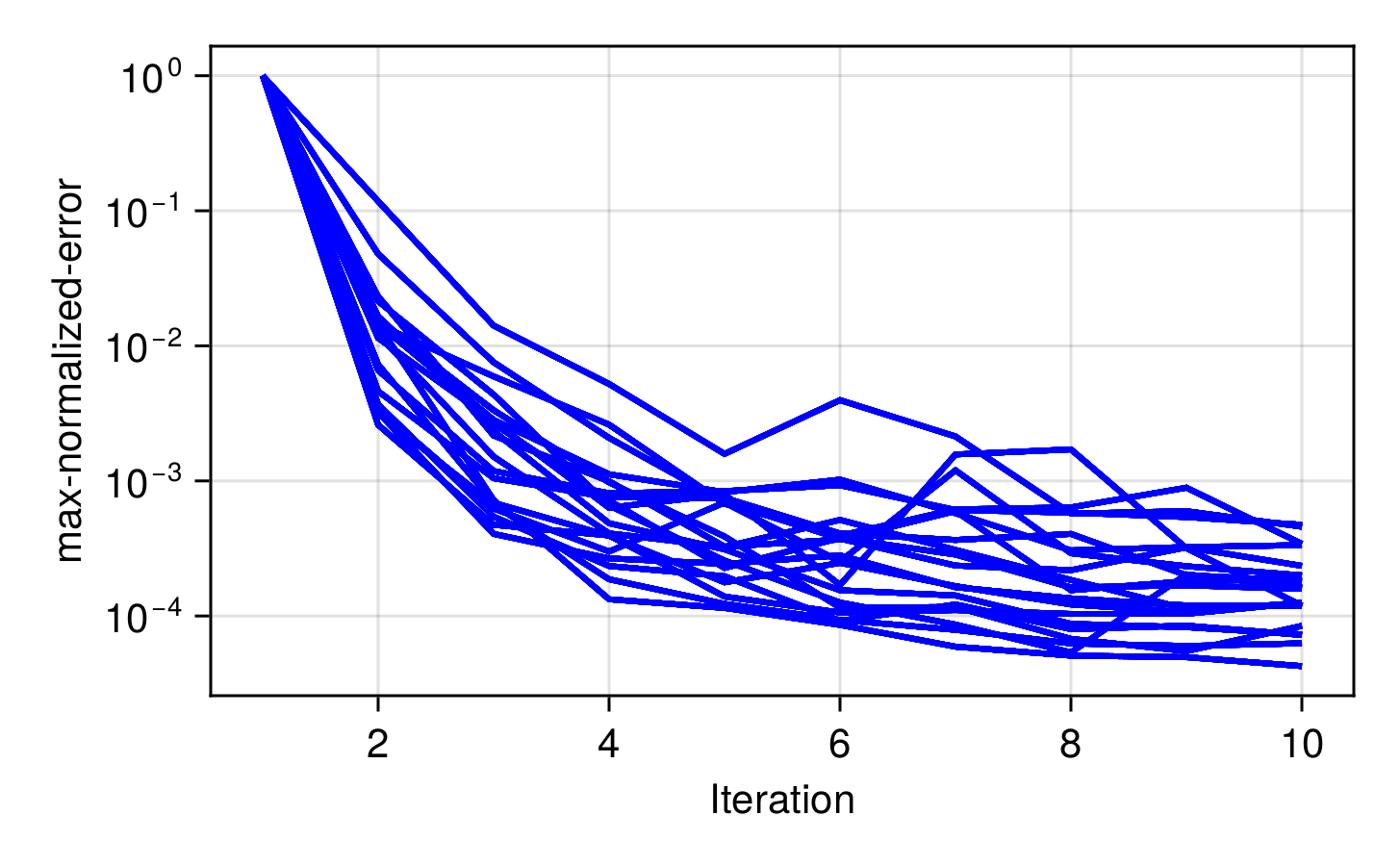}  
\caption{Convergence of the error $n_{\mathrm{iter}}^{-1}\sum_{n=1}^{n_{\mathrm{iter}}} \|\Gamma^{-1/2}(z_n - \cG(u_n))\|^2$ for $20$ independent experiments on the Lorenz 63 system, plotted on a log-scale. The initial objective value is normalized to one. For this experiment, convergence stagnates at around $8$ ensemble iterations. Iterations for this plot contain $42$ ensemble members.}
\label{fig:app2_conv}
\end{figure}

\begin{figure}[tbp]
\includegraphics[width=\textwidth]{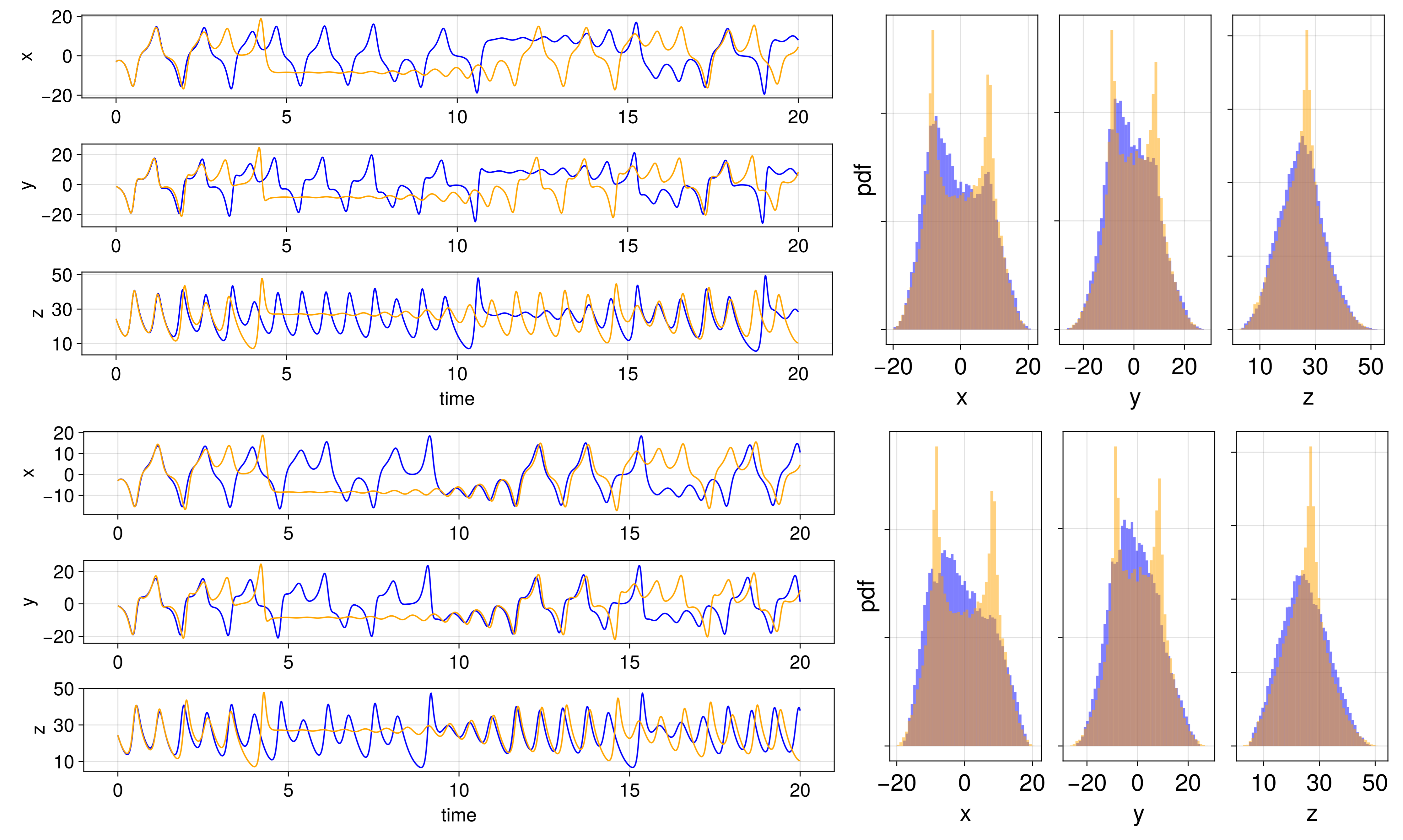} 
\caption{Row one GP and row two RF experiments for the Lorenz 63 system as in Figure \ref{fig:L63} (cf. rows two and three there), but with increased noise level corresponding to noise covariance matrix $\Sigma = 10^{-2}I$.}
\label{fig:L63_highnoise}
\end{figure}

To \nn{tune} the hyperparameters in the \nn{chaotic dynamics} experiment, the EKI ensemble size was taken to be $42$ for the $41$ parameters $\#\{\varsigma, \{U_{ij}\}_{ij}, \{S_{ii}\}_i\} = (1, 36, 4) $ \nn{in the rank-$4$ nonseparable feature distribution}. The number of features used while optimizing the hyperparameters was taken to be $M=200$. The convergence is plotted over $20$ iterations and under $20$ restarts of the algorithm with different random samplings from the prior in Figure~\ref{fig:app2_conv}. 
An experiment run with a larger noise level in the training data is shown in Figure \ref{fig:L63_highnoise}; \nn{this figure should be compared with Figure \ref{fig:L63}.}

\begin{figure}[tbp]
\centering
\includegraphics[width=\textwidth]{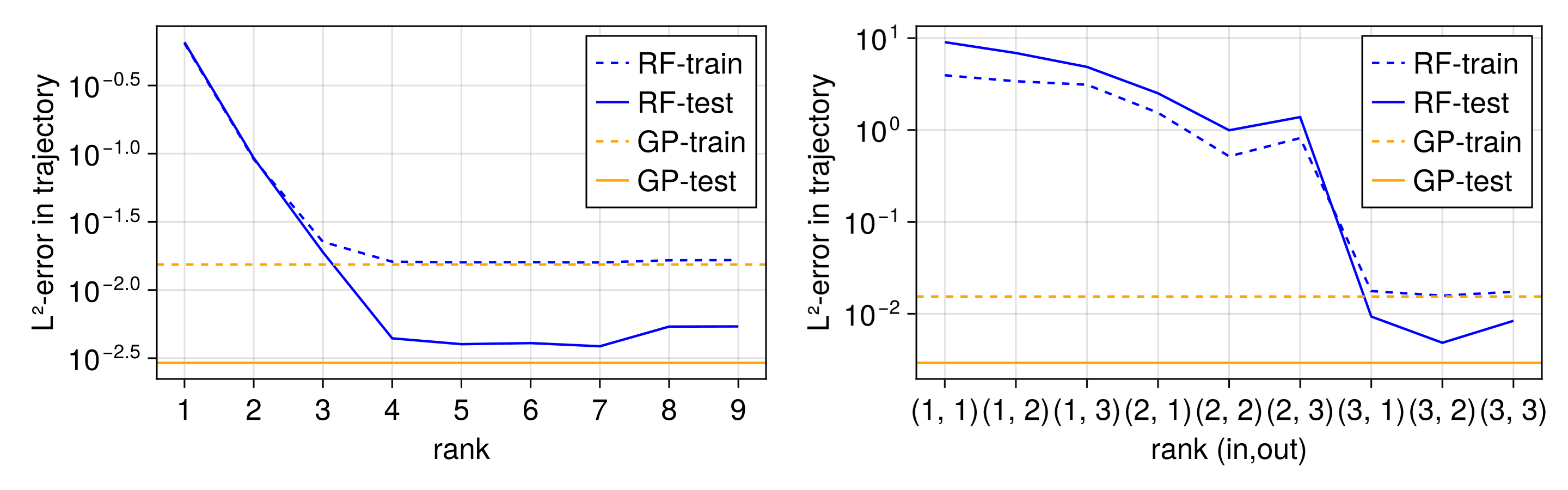} 
\caption{\nn{Results on changing the rank $r=1, \dots, 9$ of the nonseparable feature distribution (left plot) covariance \eqref{eq:rff_cov} and of the separable feature distribution covariance (right plot) with input and output ranks $r=(1,1),\ldots,(3,3)$ for the Lorenz 63 system. On each plot, the two curves represent the $L^2$ error of predicted time steps along the training (dashed) and the test (solid) trajectories. The GP results (independent of rank) are shown for comparison in orange alongside RF results in blue. The vertical axis shows the $L^2$ error of the tuned emulator against the rank on the horizontal axis.
}}
\label{fig:app2-ranktest}
\end{figure}

\nn{
\subsubsection{Performance Dependence on Feature Distribution Structure} \label{app:rank} 
To illustrate the dependence on the rank under optimization, the Lorenz 63 experiment of Section~\ref{sec:ex_lorenz} is repeated across all ranks $1,\ldots,9$ for the nonseparable feature distribution and all ranks $(1,1),\ldots,(3,3)$ for the separable feature distribution, taking all other factors (such as the number of features or optimizer options) fixed. The optimization took an mean time of 114 seconds largely independent of the rank or separability structure. In Figure~\ref{fig:app2-ranktest}, the resulting $L^2$-error of the predicted mean at each time step with both training and test data (and normalized by trajectory length) are plotted against the rank. The plot was obtained from averages over $10$ re-initializations of the optimizer. We include the performance of the GP for reference. For the nonseparable feature distribution structure, the increase in rank shows large benefits to performance on the training data until a plateau occurs at higher ranks. This motivates our choice of taking the rank equal to four in the experiments because increasing the rank quadratically increases the number of tunable parameters. The separable structure is able to achieve similar performance when the input is full rank and poor performance otherwise. Both are able to obtain performance similar to that of the GP reference, as observed in an application to forecast the attractor. We observe that in both experiments, qualitative behavior exhibited on the training trajectory is reflected in the test trajectory.
}

\subsection{Details for Section~\ref{sec:ex_cloud}:~\nameref*{sec:ex_cloud}}\label{sec:app_app3}
\begin{figure}[tbp]
\centering
\includegraphics[width=0.7\textwidth]{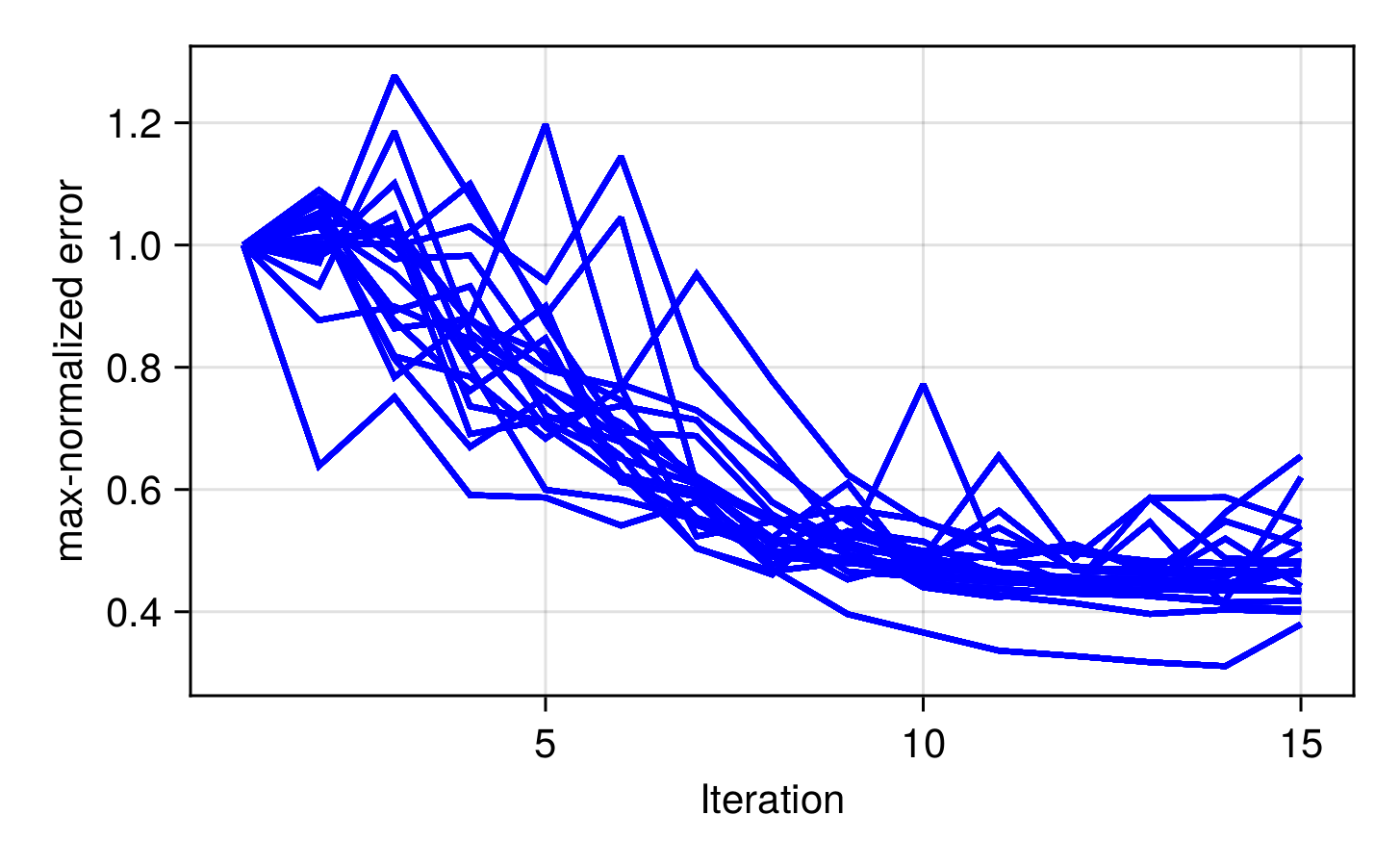}  
\caption{Convergence of the error $n_{\mathrm{iter}}^{-1}\sum_{n=1}^{n_{\mathrm{iter}}} \|\Gamma^{-\frac{1}{2}}(z_n - \cG(u_n))\|^2$ for \nn{tuning the RF emulator used in uncertainty quantification of a cloud model}, normalized to be initially one over $20$ experiments. For this experiment, convergence stagnates at around $12$ ensemble iterations. Iterations for this plot contain $635$ ensemble members.}
\label{fig:app3_conv}
\end{figure}

To \nn{tune} the hyperparameters in the \nn{Bayesian inverse problem experiment involving the cloud model}, the ensemble size was taken to be $J=635$ for the 634 parameters given by $\#\{\varsigma, \{U_{ij}\}_{ij}, \{S_{ii}\}_i\} = (1, 630, 3)$. The number of features used while optimizing the hyperparameters was taken to be $M=200$. \nn{The convergence is plotted over 20 iterations of EKI and under 20 restarts of the algorithm with different random samplings from the prior in Figure \ref{fig:app3_conv}. The priors for the physical parameters in the underlying geophysical inverse problem (not to be confused with the hyperparameter priors for the forward map surrogate) are given in Table \ref{tab:prior_edmf}.}

\begin{table}[tbp]
    \centering
    \caption{Priors for the parameters for the EDMF calibration experiment. Priors are defined by mapping unit variance normal distributions into the desired bounds with a shifted and scaled $\mathrm{logit}$ function.}
    \label{tab:prior_edmf}
    \renewcommand{\arraystretch}{1.2}
    \begin{tabular}{l@{\hspace{10.25mm}}cc}
        \toprule
         Parameter & Bounds & Prior Mean \\ \midrule
        \texttt{entrainment\_factor} & $[0, 1]$ & $0.13$ \\
        \texttt{detrainment\_factor} & $[0, 1]$ & $0.51$ \\
        \texttt{tke\_ed\_coeff} & $[0.01, 1]$ & $0.14$ \\
        \texttt{tke\_diss\_coeff} & $[0.01, 1]$ & $0.22$ \\
        \texttt{static\_stab\_coeff} & $[0.01, 1]$ & $0.40$ \\
        \bottomrule
    \end{tabular}
\end{table}

\begin{figure}[tbp]
    \centering
    \includegraphics[width=\textwidth]{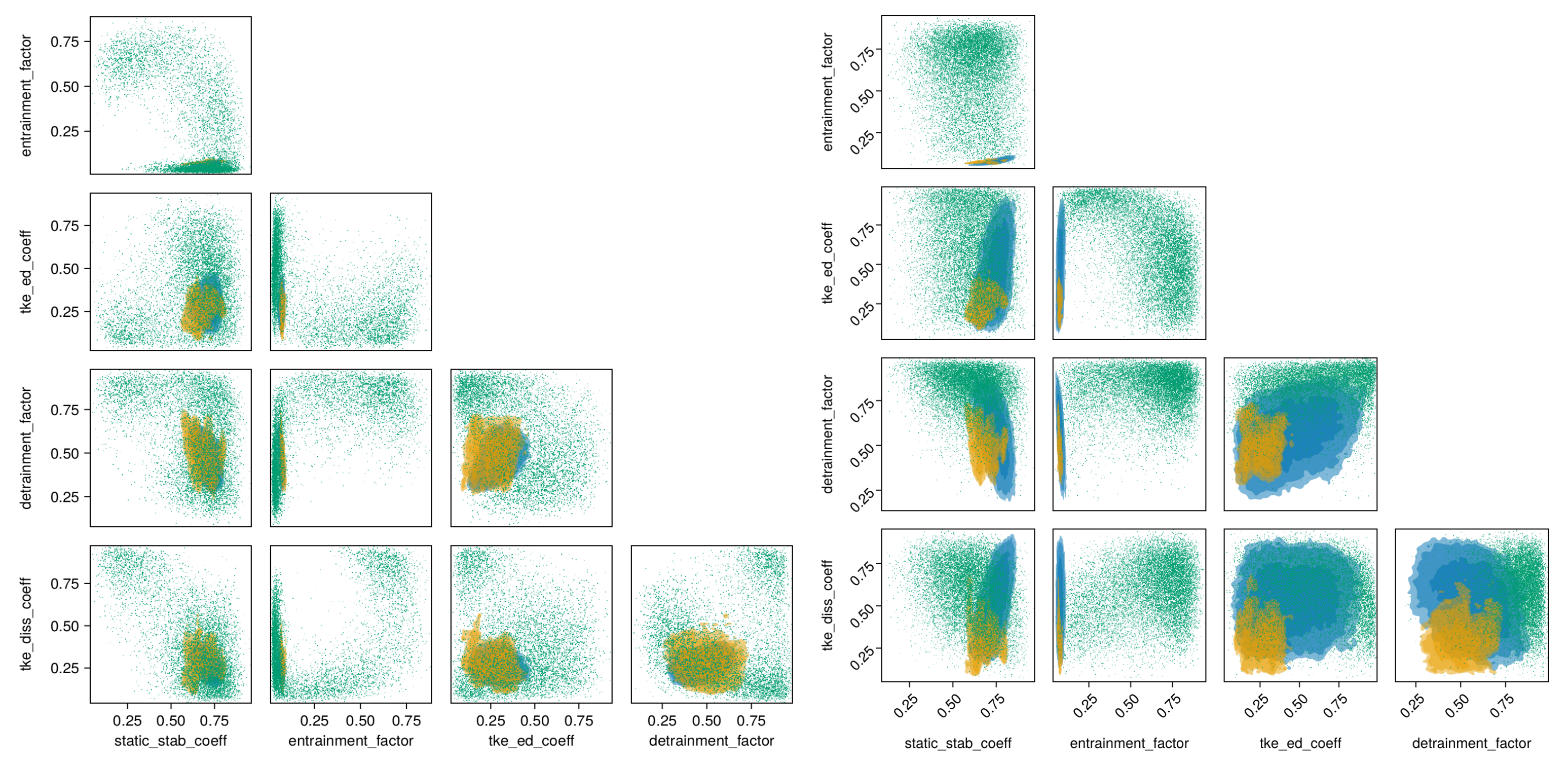}
    \caption{\nn{Additional plots for the  uncertainty quantification of a cloud model. The corner plots display marginal posterior distributions resulting from emulators using GP-CES (yellow), RF-CES (blue), and RF-CES without tuning (green scatter). Left: rank-$1$ nonseparable feature distribution in RF-CES, as shown in Figure \ref{fig:EDMF_post}. Right: separable feature distribution in RF-CES with rank (5,1).}
    }
    \label{fig:app3_uq_prior_and_sep}
\end{figure}

For the MCMC algorithm, we use a simple random walk Metropolis method. The step size was tuned in each experiment to achieve an acceptance rate of approximately $25\%$. The MCMC method is initialized in a region of high posterior mass and the GP selected a smaller step size than the RF did by a factor of \nn{approximately three}. We ran the algorithm until the posteriors appeared to converge, for a total of $2\times10^5$ steps for RF and $5\times10^5$ steps for GP. \nn{Table \ref{tab:perf-edmf} compares the time taken for tuning and prediction and also displays training and test errors with different feature distributions or with GPs. The ranks chosen for the RF method were based on having low test error (as done in Appendix \ref{app:rank}). The test set was held out of the final iteration of data from the training point selection algorithm \citep{CleGarLanSchStu21}; by construction, test accuracy is thus evaluated on a region of high posterior mass that the MCMC samples will explore. Table \ref{tab:perf-edmf} shows more consistent performance on training and test error with RF, while the GP results show a much lower error on training data than on test data, indicating possible overfitting.

\begin{table}[tbp]
    \centering
    \caption{\nn{Performance of the different emulators on the accelerated uncertainty quantification test problem, measured as average $L^2$-error over the training data and over a held-out test set, under 10 restarts of the algorithm. Reported timings are for the tuning procedure, and the average time to perform one MCMC likelihood calculation. The rank of the separable or nonseparable distribution are indicated in parentheses in column one.}}
     \label{tab:perf-edmf}
    \resizebox{\textwidth}{!}{%
        \begin{tabular}{l@{\hspace{7.25mm}}cccc}
            \toprule
            Emulator & Tuning Time & Training Error & Test Error & MCMC Step Time\\ \midrule
            GP (L-BFGS-B) & 260s & 4.85 & 15.08 & 7.8ms\\
            RF (untuned), rank $1$ & --- & 15.80 & 12.61 & 3.0ms (8 threads)\\
            RF (EKI), rank $(5,1)$ & 504s (8-threads) & 16.77 &  14.05 & 3.8ms (8 threads)\\
            RF (EKI), rank $1$ & 565s (8-threads) & 12.81 & 11.70 & 3.0ms (8 threads) \\ \bottomrule
         \end{tabular}
     }
\end{table}

In Figure \ref{fig:app3_uq_prior_and_sep}, the panels show the performance of RF-CES posteriors based on a separable feature distribution and on an untuned RF emulator, in addition to the emulator with nonseparable feature distribution seen in Subsection~\ref{sec:ex_cloud} and Figure \ref{fig:EDMF_post} in the main text. As expected, the posterior corresponding to the untuned emulator performs poorly; its samples cover much of the geophysical parameter prior support; see Table \ref{tab:prior_edmf}. However, this is somewhat surprising in light of Table \ref{tab:perf-edmf} because the untuned RF emulator obtained low test error there. This implies that low emulator test error for the forward map does not automatically translate to an accurate posterior for the inverse problem. The RF emulator with separable feature distribution also produces a posterior with too wide of a support compared to the GP-CES reference posterior. Finally, we observe that although the GP emulator obtained large forward map test error compared to that of our chosen nonseparable RF emulator, the resulting posterior distributions remained similar as seen in Figure \ref{fig:EDMF_post}.}

\vskip 0.2in
\bibliography{references}

\begin{thebibliography}{88}
\providecommand{\natexlab}[1]{#1}
\providecommand{\url}[1]{\texttt{#1}}
\expandafter\ifx\csname urlstyle\endcsname\relax
  \providecommand{\doi}[1]{doi: #1}\else
  \providecommand{\doi}{doi: \begingroup \urlstyle{rm}\Url}\fi

\bibitem[Ambikasaran et~al.(2016)Ambikasaran, O'Neil, and
  Singh]{AmbONeSin16_pre}
S.~Ambikasaran, M.~O'Neil, and K.~R. Singh.
\newblock Fast symmetric factorization of hierarchical matrices with
  applications.
\newblock \emph{preprint arXiv.1405.0223}, 2016.

\bibitem[Antonik et~al.(2023)Antonik, Marsal, Brunner, and
  Rontani]{AntBruMarRon23}
P.~Antonik, N.~Marsal, D.~Brunner, and D.~Rontani.
\newblock Bayesian optimisation of large-scale photonic reservoir computers.
\newblock \emph{Cognitive Computation}, pages 1--9, 2023.

\bibitem[Archer et~al.(1997)Archer, Saltelli, and Sobol]{ArcSalSob97}
G.~E.~B. Archer, A.~Saltelli, and I.~M. Sobol.
\newblock Sensitivity measures, {ANOVA}-like techniques and the use of
  bootstrap.
\newblock \emph{Journal of Statistical Computation and Simulation}, 58\penalty0
  (2):\penalty0 99--120, 1997.

\bibitem[Bergstra and Bengio(2012)]{bergstra2012random}
J.~Bergstra and Y.~Bengio.
\newblock Random search for hyper-parameter optimization.
\newblock \emph{Journal of machine learning research}, 13\penalty0 (2), 2012.

\bibitem[Bishop and Nasrabadi(2006)]{bishop2006pattern}
C.~M. Bishop and N.~M. Nasrabadi.
\newblock \emph{Pattern recognition and machine learning}.
\newblock Springer, 2006.

\bibitem[Bollt(2021)]{bollt2021explaining}
E.~Bollt.
\newblock {On explaining the surprising success of reservoir computing
  forecaster of chaos? The universal machine learning dynamical system with
  contrast to VAR and DMD}.
\newblock \emph{Chaos: An Interdisciplinary Journal of Nonlinear Science},
  31\penalty0 (1):\penalty0 013108, 2021.

\bibitem[B{\"o}ttcher(2023)]{Bot23_pre}
L.~B{\"o}ttcher.
\newblock {Gradient-free training of neural ODEs for system identification and
  control using ensemble Kalman inversion}.
\newblock \emph{preprint arXiv:2307.07882}, 2023.

\bibitem[Brault et~al.(2016)Brault, Heinonen, and Buc]{brault2016random}
R.~Brault, M.~Heinonen, and F.~Buc.
\newblock Random {Fourier} features for operator-valued kernels.
\newblock In \emph{Asian Conference on Machine Learning}, pages 110--125. PMLR,
  2016.

\bibitem[Calvello et~al.(2025)Calvello, Reich, and Stuart]{CalReiStu22}
E.~Calvello, S.~Reich, and A.~M. Stuart.
\newblock {Ensemble Kalman Methods: A Mean Field Perspective}.
\newblock \emph{Acta Numerica}, 2025.

\bibitem[Chattopadhyay et~al.(2020)Chattopadhyay, Hassanzadeh, and
  Subramanian]{ChaHasSub20}
A.~Chattopadhyay, P.~Hassanzadeh, and D.~Subramanian.
\newblock {Data-driven predictions of a multiscale Lorenz 96 chaotic system
  using machine-learning methods: Reservoir computing, artificial neural
  network, and long short-term memory network}.
\newblock \emph{Nonlinear Processes in Geophysics}, 27\penalty0 (3):\penalty0
  373--389, 2020.

\bibitem[Chen and Oliver(2012)]{CheOli12}
Y.~Chen and D.~S. Oliver.
\newblock Ensemble randomized maximum likelihood method as an iterative
  ensemble smoother.
\newblock \emph{Mathematical Geosciences}, 44:\penalty0 1--26, 2012.

\bibitem[Chen et~al.(2021)Chen, Owhadi, and Stuart]{chen2021consistency}
Y.~Chen, H.~Owhadi, and A.~M. Stuart.
\newblock Consistency of empirical {Bayes} and kernel flow for hierarchical
  parameter estimation.
\newblock \emph{Mathematics of Computation}, 90\penalty0 (332):\penalty0
  2527--2578, 2021.

\bibitem[Cleary et~al.(2021)Cleary, Garbuno-Inigo, Lan, Schneider, and
  Stuart]{CleGarLanSchStu21}
E.~Cleary, A.~Garbuno-Inigo, S.~Lan, T.~Schneider, and A.~M. Stuart.
\newblock Calibrate, emulate, sample.
\newblock \emph{Journal of Computational Physics}, 424:\penalty0 109716, 2021.
\newblock ISSN 0021-9991.

\bibitem[Dudul(2005)]{Dud05}
S.~V. Dudul.
\newblock Prediction of a {Lorenz} chaotic attractor using two-layer perceptron
  neural network.
\newblock \emph{Applied Soft Computing}, 5\penalty0 (4):\penalty0 333--355,
  2005.
\newblock ISSN 1568-4946.

\bibitem[Dunbar et~al.(2022{\natexlab{a}})Dunbar, Duncan, Stuart, and
  Wolfram]{DunDunStuWol22}
O.~R. Dunbar, A.~B. Duncan, A.~M. Stuart, and M.-T. Wolfram.
\newblock Ensemble inference methods for models with noisy and expensive
  likelihoods.
\newblock \emph{SIAM Journal on Applied Dynamical Systems}, 21\penalty0
  (2):\penalty0 1539--1572, 2022{\natexlab{a}}.

\bibitem[Dunbar et~al.(2021)Dunbar, Garbuno-Inigo, Schneider, and
  Stuart]{DunGarSchStu21}
O.~R.~A. Dunbar, A.~Garbuno-Inigo, T.~Schneider, and A.~M. Stuart.
\newblock {Calibration and Uncertainty Quantification of Convective Parameters
  in an Idealized GCM}.
\newblock \emph{Journal of Advances in Modeling Earth Systems}, 13\penalty0
  (9):\penalty0 e2020MS002454, 2021.

\bibitem[Dunbar et~al.(2022{\natexlab{b}})Dunbar, Lopez-Gomez, Garbuno-Iñigo,
  Huang, Bach, and Wu]{BacDunGarHuaLopWu22}
O.~R.~A. Dunbar, I.~Lopez-Gomez, A.~Garbuno-Iñigo, D.~Z. Huang, E.~Bach, and
  J.-l. Wu.
\newblock {EnsembleKalmanProcesses.jl: Derivative-free ensemble-based model
  calibration}.
\newblock \emph{Journal of Open Source Software}, 7\penalty0 (80):\penalty0
  4869, 2022{\natexlab{b}}.

\bibitem[Dunbar et~al.(2024)Dunbar, Bieli, Garbuno-Iñigo, Howland, Souza,
  Mansfield, Wagner, and Efrat-Henrici]{Dun_etal24}
O.~R.~A. Dunbar, M.~Bieli, A.~Garbuno-Iñigo, M.~Howland, A.~N. Souza, L.~A.
  Mansfield, G.~L. Wagner, and N.~Efrat-Henrici.
\newblock {CalibrateEmulateSample.jl: Accelerated Parametric Uncertainty
  Quantification}.
\newblock \emph{Journal of Open Source Software}, 9\penalty0 (97):\penalty0
  6372, 2024.

\bibitem[Dutilleul(1999)]{dutilleul1999mle}
P.~Dutilleul.
\newblock {The MLE algorithm for the matrix normal distribution}.
\newblock \emph{Journal of Statistical Computation and Simulation}, 64\penalty0
  (2):\penalty0 105--123, 1999.

\bibitem[Fablet et~al.(2018)Fablet, Ouala, and Herzet]{FabHerOua18}
R.~Fablet, S.~Ouala, and C.~Herzet.
\newblock Bilinear residual neural network for the identification and
  forecasting of geophysical dynamics.
\newblock In \emph{2018 26th European Signal Processing Conference (EUSIPCO)},
  pages 1477--1481, 2018.

\bibitem[Falk et~al.(2022)Falk, Cilibert, and Pontil]{falk2022implicit}
J.~I.~T. Falk, C.~Cilibert, and M.~Pontil.
\newblock Implicit kernel meta-learning using kernel integral forms.
\newblock In \emph{Uncertainty in Artificial Intelligence}, pages 652--662.
  PMLR, 2022.

\bibitem[Garbuno-Inigo et~al.(2020{\natexlab{a}})Garbuno-Inigo, Hoffmann, Li,
  and Stuart]{GarHofLiStu20}
A.~Garbuno-Inigo, F.~Hoffmann, W.~Li, and A.~M. Stuart.
\newblock {Interacting Langevin diffusions: Gradient structure and ensemble
  Kalman sampler}.
\newblock \emph{SIAM Journal on Applied Dynamical Systems}, 19\penalty0
  (1):\penalty0 412--441, 2020{\natexlab{a}}.

\bibitem[Garbuno-Inigo et~al.(2020{\natexlab{b}})Garbuno-Inigo, N\"usken, and
  Reich]{GarNusRei20}
A.~Garbuno-Inigo, N.~N\"usken, and S.~Reich.
\newblock {Affine invariant interacting Langevin dynamics for Bayesian
  inference}.
\newblock \emph{SIAM Journal on Applied Dynamical Systems}, 19\penalty0
  (3):\penalty0 1633--1658, 2020{\natexlab{b}}.

\bibitem[Gauthier(2018)]{gauthier2018reservoir}
D.~J. Gauthier.
\newblock Reservoir computing: Harnessing a universal dynamical system.
\newblock \emph{SIAM News}, 51\penalty0 (2):\penalty0 12, 2018.

\bibitem[Gauthier et~al.(2021)Gauthier, Bollt, Griffith, and
  Barbosa]{BarBolGauGri21}
D.~J. Gauthier, E.~Bollt, A.~Griffith, and W.~A. Barbosa.
\newblock Next generation reservoir computing.
\newblock \emph{Nature Communications}, 12\penalty0 (1):\penalty0 5564, 2021.

\bibitem[Gelman et~al.(2017)Gelman, Simpson, and Betancourt]{BetGelSim17}
A.~Gelman, D.~Simpson, and M.~Betancourt.
\newblock The prior can often only be understood in the context of the
  likelihood.
\newblock \emph{Entropy}, 19\penalty0 (10), 2017.
\newblock ISSN 1099-4300.

\bibitem[Griffith et~al.(2019)Griffith, Pomerance, and
  Gauthier]{griffith2019forecasting}
A.~Griffith, A.~Pomerance, and D.~J. Gauthier.
\newblock Forecasting chaotic systems with very low connectivity reservoir
  computers.
\newblock \emph{Chaos: An Interdisciplinary Journal of Nonlinear Science},
  29\penalty0 (12):\penalty0 123108, 2019.

\bibitem[Hamid et~al.(2014)Hamid, Xiao, Gittens, and DeCoste]{DeCGitHamXia14}
R.~Hamid, Y.~Xiao, A.~Gittens, and D.~DeCoste.
\newblock Compact random feature maps.
\newblock In E.~P. Xing and T.~Jebara, editors, \emph{Proceedings of the 31st
  International Conference on Machine Learning}, volume~32 of \emph{Proceedings
  of Machine Learning Research}, pages 19--27, Bejing, China, 22--24 Jun 2014.
  PMLR.

\bibitem[Hensman et~al.(2015)Hensman, Matthews, and
  Ghahramani]{hensman2015scalable}
J.~Hensman, A.~Matthews, and Z.~Ghahramani.
\newblock Scalable variational {Gaussian} process classification.
\newblock In \emph{Artificial Intelligence and Statistics}, pages 351--360.
  PMLR, 2015.

\bibitem[Hensman et~al.(2018)Hensman, Durrande, and
  Solin]{hensman2018variational}
J.~Hensman, N.~Durrande, and A.~Solin.
\newblock {Variational Fourier features for Gaussian processes}.
\newblock \emph{Journal of Machine Learning Research}, 18\penalty0
  (151):\penalty0 1--52, 2018.

\bibitem[Huang et~al.(2022)Huang, Huang, Reich, and Stuart]{HuaHuaReiStu22}
D.~Z. Huang, J.~Huang, S.~Reich, and A.~M. Stuart.
\newblock Efficient derivative-free {Bayesian} inference for large-scale
  inverse problems.
\newblock \emph{Inverse Problems}, 38\penalty0 (12):\penalty0 125006, oct 2022.

\bibitem[Huang et~al.(2006)Huang, Zhu, and Siew]{huang2006extreme}
G.-B. Huang, Q.-Y. Zhu, and C.-K. Siew.
\newblock Extreme learning machine: theory and applications.
\newblock \emph{Neurocomputing}, 70\penalty0 (1-3):\penalty0 489--501, 2006.

\bibitem[Hvarfner et~al.(2024)Hvarfner, Hellsten, and
  Nardi]{hvarfner2024vanilla}
C.~Hvarfner, E.~O. Hellsten, and L.~Nardi.
\newblock Vanilla {Bayesian} optimization performs great in high dimension.
\newblock \emph{preprint arXiv:2402.02229}, 2024.

\bibitem[Iglesias and Yang(2021)]{IglYan21}
M.~Iglesias and Y.~Yang.
\newblock {Adaptive regularisation for ensemble Kalman inversion}.
\newblock \emph{Inverse Problems}, 37\penalty0 (2):\penalty0 025008, 2021.

\bibitem[Iglesias et~al.(2013)Iglesias, Law, and Stuart]{IglLawStu13}
M.~A. Iglesias, K.~J. Law, and A.~M. Stuart.
\newblock Ensemble kalman methods for inverse problems.
\newblock \emph{Inverse Problems}, 29\penalty0 (4):\penalty0 045001, 2013.

\bibitem[Ishigami and Homma(1990)]{HomIsh90}
T.~Ishigami and T.~Homma.
\newblock An importance quantification technique in uncertainty analysis for
  computer models.
\newblock In \emph{First International Symposium on Uncertainty Modeling and
  Analysis}, pages 398--403. IEEE, 1990.

\bibitem[Kadri et~al.(2016)Kadri, Duflos, Preux, Canu, Rakotomamonjy, and
  Audiffren]{kadri2016operator}
H.~Kadri, E.~Duflos, P.~Preux, S.~Canu, A.~Rakotomamonjy, and J.~Audiffren.
\newblock Operator-valued kernels for learning from functional response data.
\newblock \emph{Journal of Machine Learning Research}, 17\penalty0
  (20):\penalty0 1--54, 2016.

\bibitem[Kammonen et~al.(2020)Kammonen, Kiessling, Plech{\'a}{\v{c}}, Sandberg,
  and Szepessy]{kammonen2020adaptive}
A.~Kammonen, J.~Kiessling, P.~Plech{\'a}{\v{c}}, M.~Sandberg, and A.~Szepessy.
\newblock {Adaptive random Fourier features with Metropolis sampling}.
\newblock \emph{Foundations of Data Science}, 2\penalty0 (3):\penalty0
  309--332, 2020.

\bibitem[Kammonen et~al.(2023)Kammonen, Kiessling, Plech{\'a}{\v{c}}, Sandberg,
  Szepessy, and Tempone]{kammonen2023smaller}
A.~Kammonen, J.~Kiessling, P.~Plech{\'a}{\v{c}}, M.~Sandberg, A.~Szepessy, and
  R.~Tempone.
\newblock Smaller generalization error derived for a deep residual neural
  network compared with shallow networks.
\newblock \emph{IMA Journal of Numerical Analysis}, 43\penalty0 (5):\penalty0
  2585--2632, 2023.

\bibitem[Kanagawa et~al.(2018)Kanagawa, Hennig, Sejdinovic, and
  Sriperumbudur]{HenKanSejSri18_pre}
M.~Kanagawa, P.~Hennig, D.~Sejdinovic, and B.~K. Sriperumbudur.
\newblock {Gaussian processes and kernel methods: A review on connections and
  equivalences}.
\newblock \emph{preprint arXiv:1807.02582}, 2018.

\bibitem[Kovachki and Stuart(2019)]{KovStu19}
N.~B. Kovachki and A.~M. Stuart.
\newblock Ensemble {Kalman} inversion: a derivative-free technique for machine
  learning tasks.
\newblock \emph{Inverse Problems}, 35\penalty0 (9):\penalty0 095005, 2019.

\bibitem[Lanthaler and Nelsen(2023)]{lanthaler2023error}
S.~Lanthaler and N.~H. Nelsen.
\newblock Error bounds for learning with vector-valued random features.
\newblock In A.~Oh, T.~Neumann, A.~Globerson, K.~Saenko, M.~Hardt, and
  S.~Levine, editors, \emph{Advances in Neural Information Processing Systems},
  volume~36, pages 71834--71861. Curran Associates, Inc., 2023.

\bibitem[Le et~al.(2013)Le, Sarl{\'o}s, Smola, et~al.]{LeSarSmo13}
Q.~Le, T.~Sarl{\'o}s, A.~Smola, et~al.
\newblock Fastfood-approximating kernel expansions in loglinear time.
\newblock In \emph{Proceedings of the International Conference on Machine
  Learning}, volume~85, 2013.

\bibitem[Ledoit and Wolf(2004)]{LedWol04}
O.~Ledoit and M.~Wolf.
\newblock A well-conditioned estimator for large-dimensional covariance
  matrices.
\newblock \emph{Journal of Multivariate Analysis}, 88\penalty0 (2):\penalty0
  365--411, 2004.

\bibitem[Li et~al.(2019)Li, Chang, Mroueh, Yang, and Poczos]{ChaLiMroPocYan19}
C.-L. Li, W.-C. Chang, Y.~Mroueh, Y.~Yang, and B.~Poczos.
\newblock Implicit kernel learning.
\newblock In K.~Chaudhuri and M.~Sugiyama, editors, \emph{Proceedings of the
  Twenty-Second International Conference on Artificial Intelligence and
  Statistics}, volume~89 of \emph{Proceedings of Machine Learning Research},
  pages 2007--2016. PMLR, 16--18 Apr 2019.

\bibitem[Liu et~al.(2022)Liu, Huang, Chen, and Suykens]{CheHuaLiuSuy22}
F.~Liu, X.~Huang, Y.~Chen, and J.~A.~K. Suykens.
\newblock Random features for kernel approximation: A survey on algorithms,
  theory, and beyond.
\newblock \emph{IEEE Transactions on Pattern Analysis and Machine
  Intelligence}, 44\penalty0 (10):\penalty0 7128--7148, 2022.

\bibitem[Lopez-Gomez et~al.(2022)Lopez-Gomez, Christopoulos, Langeland~Ervik,
  Dunbar, Cohen, and Schneider]{Lop_eta22}
I.~Lopez-Gomez, C.~Christopoulos, H.~L. Langeland~Ervik, O.~R.~A. Dunbar,
  Y.~Cohen, and T.~Schneider.
\newblock {Training Physics-Based Machine-Learning Parameterizations With
  Gradient-Free Ensemble Kalman Methods}.
\newblock \emph{Journal of Advances in Modeling Earth Systems}, 14\penalty0
  (8):\penalty0 e2022MS003105, 2022.

\bibitem[Lorenz(1963)]{Lor63}
E.~N. Lorenz.
\newblock Deterministic nonperiodic flow.
\newblock \emph{Journal of Atmospheric Sciences}, 20\penalty0 (2):\penalty0
  130--141, 1963.

\bibitem[Maddox et~al.(2021)Maddox, Balandat, Wilson, and
  Bakshy]{maddox2021bayesian}
W.~J. Maddox, M.~Balandat, A.~G. Wilson, and E.~Bakshy.
\newblock Bayesian optimization with high-dimensional outputs.
\newblock In \emph{Advances in Neural Information Processing Systems},
  volume~34, pages 19274--19287, 2021.

\bibitem[Meanti et~al.(2022)Meanti, Carratino, De~Vito, and
  Rosasco]{meanti2022efficient}
G.~Meanti, L.~Carratino, E.~De~Vito, and L.~Rosasco.
\newblock Efficient hyperparameter tuning for large scale kernel ridge
  regression.
\newblock In \emph{International Conference on Artificial Intelligence and
  Statistics}, pages 6554--6572. PMLR, 2022.

\bibitem[Metropolis et~al.(1953)Metropolis, Rosenbluth, Rosenbluth, Teller, and
  Teller]{Met_etal53}
N.~Metropolis, A.~W. Rosenbluth, M.~N. Rosenbluth, A.~H. Teller, and E.~Teller.
\newblock Equation of state calculations by fast computing machines.
\newblock \emph{The Journal of Chemical Physics}, 21\penalty0 (6):\penalty0
  1087--1092, 1953.

\bibitem[Micchelli and Pontil(2004)]{micchelli2004kernels}
C.~Micchelli and M.~Pontil.
\newblock Kernels for multi--task learning.
\newblock In \emph{Advances in Neural Information Processing Systems},
  volume~17, 2004.

\bibitem[Micchelli and Pontil(2005)]{micchelli2005learning}
C.~A. Micchelli and M.~Pontil.
\newblock On learning vector-valued functions.
\newblock \emph{Neural Computation}, 17\penalty0 (1):\penalty0 177--204, 2005.

\bibitem[Naslidnyk et~al.(2023)Naslidnyk, Kanagawa, Karvonen, and
  Mahsereci]{naslidnyk2023comparing}
M.~Naslidnyk, M.~Kanagawa, T.~Karvonen, and M.~Mahsereci.
\newblock {Comparing Scale Parameter Estimators for Gaussian Process
  Regression: Cross Validation and Maximum Likelihood}.
\newblock \emph{preprint arXiv:2307.07466}, 2023.

\bibitem[Nelsen and Stuart(2021)]{nelsen2021random}
N.~H. Nelsen and A.~M. Stuart.
\newblock {The random feature model for input-output maps between Banach
  spaces}.
\newblock \emph{SIAM Journal on Scientific Computing}, 43\penalty0
  (5):\penalty0 A3212--A3243, 2021.

\bibitem[Owhadi(2022)]{owhadi2022ideas}
H.~Owhadi.
\newblock {Do ideas have shape? Idea registration as the continuous limit of
  artificial neural networks}.
\newblock \emph{Physica D: Nonlinear Phenomena}, art. 133592, 2022.

\bibitem[Owhadi and Yoo(2019)]{OwhYoo19}
H.~Owhadi and G.~R. Yoo.
\newblock Kernel flows: From learning kernels from data into the abyss.
\newblock \emph{Journal of Computational Physics}, 389:\penalty0 22--47, 2019.
\newblock ISSN 0021-9991.

\bibitem[Pahlavan et~al.(2024)Pahlavan, Hassanzadeh, and
  Alexander]{AleHasPah24}
H.~A. Pahlavan, P.~Hassanzadeh, and M.~J. Alexander.
\newblock {Explainable Offline-Online Training of Neural Networks for
  Parameterizations: A 1D Gravity Wave-QBO Testbed in the Small-Data Regime}.
\newblock \emph{Geophysical Research Letters}, 51\penalty0 (2):\penalty0
  e2023GL106324, 2024.

\bibitem[Racca and Magri(2021)]{MagRac21}
A.~Racca and L.~Magri.
\newblock Robust optimization and validation of echo state networks for
  learning chaotic dynamics.
\newblock \emph{Neural Networks}, 142:\penalty0 252--268, 2021.
\newblock ISSN 0893-6080.

\bibitem[Rahimi and Recht(2007)]{RahRec07}
A.~Rahimi and B.~Recht.
\newblock Random features for large-scale kernel machines.
\newblock In \emph{Advances in Neural Information Processing Systems},
  volume~20. Citeseer, 2007.

\bibitem[Rahimi and Recht(2008{\natexlab{a}})]{RahRec08}
A.~Rahimi and B.~Recht.
\newblock Uniform approximation of functions with random bases.
\newblock In \emph{2008 46th Annual Allerton Conference on Communication,
  Control, and Computing}, pages 555--561. IEEE, 2008{\natexlab{a}}.

\bibitem[Rahimi and Recht(2008{\natexlab{b}})]{rahimi2008weighted}
A.~Rahimi and B.~Recht.
\newblock Weighted sums of random kitchen sinks: Replacing minimization with
  randomization in learning.
\newblock In \emph{Advances in Neural Information Processing Systems},
  volume~21, 2008{\natexlab{b}}.

\bibitem[Real et~al.(2017)Real, Moore, Selle, Saxena, Suematsu, Tan, Le, and
  Kurakin]{Rea_etal17}
E.~Real, S.~Moore, A.~Selle, S.~Saxena, Y.~L. Suematsu, J.~Tan, Q.~V. Le, and
  A.~Kurakin.
\newblock Large-scale evolution of image classifiers.
\newblock In \emph{International conference on machine learning}, pages
  2902--2911. PMLR, 2017.

\bibitem[Rudi and Rosasco(2017)]{rudi2017generalization}
A.~Rudi and L.~Rosasco.
\newblock Generalization properties of learning with random features.
\newblock \emph{Advances in Neural Information Processing Systems}, 30, 2017.

\bibitem[Saltelli(2002)]{Sal02}
A.~Saltelli.
\newblock Sensitivity analysis for importance assessment.
\newblock \emph{Risk Analysis}, 22\penalty0 (3):\penalty0 579--590, 2002.

\bibitem[Saltelli et~al.(2010)Saltelli, Annoni, Azzini, Campolongo, Ratto, and
  Tarantola]{AnnAzzCamRatSalTar10}
A.~Saltelli, P.~Annoni, I.~Azzini, F.~Campolongo, M.~Ratto, and S.~Tarantola.
\newblock Variance based sensitivity analysis of model output: Design and
  estimator for the total sensitivity index.
\newblock \emph{Computer Physics Communications}, 181\penalty0 (2):\penalty0
  259--270, 2010.
\newblock ISSN 0010-4655.

\bibitem[Sanz-Alonso et~al.(2023)Sanz-Alonso, Stuart, and
  Taeb]{sanz2023inverse}
D.~Sanz-Alonso, A.~M. Stuart, and A.~Taeb.
\newblock \emph{Inverse problems and data assimilation}, volume 107.
\newblock Cambridge University Press, 2023.

\bibitem[Scardapane and Wang(2017)]{ScaWan17}
S.~Scardapane and D.~Wang.
\newblock Randomness in neural networks: an overview.
\newblock \emph{WIREs Data Mining and Knowledge Discovery}, 7\penalty0
  (2):\penalty0 e1200, 2017.

\bibitem[Scher and Messori(2019)]{SchMes19}
S.~Scher and G.~Messori.
\newblock Generalization properties of feed-forward neural networks trained on
  {Lorenz} systems.
\newblock \emph{Nonlinear Processes in Geophysics}, 26\penalty0 (4):\penalty0
  381--399, 2019.

\bibitem[Schillings and Stuart(2017)]{SchStu17}
C.~Schillings and A.~M. Stuart.
\newblock {Analysis of the ensemble Kalman filter for inverse problems}.
\newblock \emph{SIAM Journal on Numerical Analysis}, 55\penalty0 (3):\penalty0
  1264--1290, 2017.

\bibitem[Shahi et~al.(2022)Shahi, Fenton, and Cherry]{CheFlaSha22}
S.~Shahi, F.~H. Fenton, and E.~M. Cherry.
\newblock Prediction of chaotic time series using recurrent neural networks and
  reservoir computing techniques: A comparative study.
\newblock \emph{Machine Learning with Applications}, 8:\penalty0 100300, 2022.
\newblock ISSN 2666-8270.

\bibitem[Sherlock et~al.(2010)Sherlock, Fearnhead, and Roberts]{FeaRobShe10}
C.~Sherlock, P.~Fearnhead, and G.~O. Roberts.
\newblock {The Random Walk Metropolis: Linking Theory and Practice Through a
  Case Study}.
\newblock \emph{Statistical Science}, 25\penalty0 (2):\penalty0 172--190, 2010.
\newblock ISSN 08834237.

\bibitem[Sinha and Duchi(2016)]{DucSin16}
A.~Sinha and J.~C. Duchi.
\newblock Learning kernels with random features.
\newblock In \emph{Advances in Neural Information Processing Systems},
  volume~29, 2016.

\bibitem[Sobol and Levitan(1999)]{LevSob99}
I.~M. Sobol and Y.~L. Levitan.
\newblock On the use of variance reducing multipliers in {Monte Carlo}
  computations of a global sensitivity index.
\newblock \emph{Computer Physics Communications}, 117\penalty0 (1):\penalty0
  52--61, 1999.

\bibitem[Stanley et~al.(2019)Stanley, Clune, Lehman, and
  Miikkulainen]{CluLehMiiSta19}
K.~O. Stanley, J.~Clune, J.~Lehman, and R.~Miikkulainen.
\newblock Designing neural networks through neuroevolution.
\newblock \emph{Nature Machine Intelligence}, 1\penalty0 (1):\penalty0 24--35,
  2019.

\bibitem[Stuart(2010)]{stuart2010inverse}
A.~M. Stuart.
\newblock {Inverse problems: a Bayesian perspective}.
\newblock \emph{Acta Numerica}, 19:\penalty0 451--559, 2010.

\bibitem[Sun et~al.(2015)Sun, Zhao, and Zhu]{SunZhaZhu15}
S.~Sun, J.~Zhao, and J.~Zhu.
\newblock {A review of Nystr\"om methods for large-scale machine learning}.
\newblock \emph{Information Fusion}, 26:\penalty0 36--48, 2015.
\newblock ISSN 1566-2535.

\bibitem[Titsias(2009)]{titsias2009variational}
M.~Titsias.
\newblock Variational learning of inducing variables in sparse {Gaussian}
  processes.
\newblock In \emph{Artificial Intelligence and Statistics}, pages 567--574.
  PMLR, 2009.

\bibitem[Vishny et~al.(2024)Vishny, Morzfeld, Gwirtz, Bach, Dunbar, and
  Hodyss]{Morzfeld_etal24}
D.~Vishny, M.~Morzfeld, K.~Gwirtz, E.~Bach, O.~R.~A. Dunbar, and D.~Hodyss.
\newblock High-dimensional covariance estimation from a small number of
  samples.
\newblock \emph{Journal of Advances in Modeling Earth Systems}, 16\penalty0
  (9):\penalty0 e2024MS004417, 2024.

\bibitem[Williams and Seeger(2000)]{WilSee00}
C.~Williams and M.~Seeger.
\newblock {Using the Nystr{\"o}m method to speed up kernel machines}.
\newblock In \emph{Advances in Neural Information Processing Systems},
  volume~13, 2000.

\bibitem[Williams and Rasmussen(2006)]{RasWil06}
C.~K. Williams and C.~E. Rasmussen.
\newblock \emph{Gaussian processes for machine learning}, volume~2.
\newblock MIT Press, Cambridge, MA, 2006.

\bibitem[Wu et~al.(2024)Wu, Levine, Schneider, and Stuart]{LevSchStuWu24}
J.-L. Wu, M.~E. Levine, T.~Schneider, and A.~Stuart.
\newblock Learning about structural errors in models of complex dynamical
  systems.
\newblock \emph{Journal of Computational Physics}, art. 113157, 2024.

\bibitem[Xiao et~al.(2016)Xiao, Wu, Wang, Sun, and Roy]{XiaSunRoyWanWu16}
H.~Xiao, J.-L. Wu, J.-X. Wang, R.~Sun, and C.~J. Roy.
\newblock {Quantifying and reducing model-form uncertainties in
  Reynolds-averaged Navier–Stokes simulations: A data-driven,
  physics-informed Bayesian approach}.
\newblock \emph{Journal of Computational Physics}, 324:\penalty0 115--136,
  2016.
\newblock ISSN 0021-9991.

\bibitem[Xiao et~al.(2020)Xiao, Yan, Basodi, Ji, and Pan]{BasJiPanXiaYan20_pre}
X.~Xiao, M.~Yan, S.~Basodi, C.~Ji, and Y.~Pan.
\newblock Efficient hyperparameter optimization in deep learning using a
  variable length genetic algorithm.
\newblock \emph{preprint arXiv:2006.12703}, 2020.

\bibitem[Xu et~al.(2024)Xu, Wang, Phillips, and Zhe]{xu2024standard}
Z.~Xu, H.~Wang, J.~M. Phillips, and S.~Zhe.
\newblock Standard {Gaussian} process can be excellent for high-dimensional
  {Bayesian} optimization.
\newblock \emph{preprint arXiv:2402.02746}, 2024.

\bibitem[Yang and Hu(2020)]{yang2020feature}
G.~Yang and E.~J. Hu.
\newblock Feature learning in infinite-width neural networks.
\newblock \emph{preprint arXiv:2011.14522}, 2020.

\bibitem[Yang et~al.(2022)Yang, Hu, Babuschkin, Sidor, Liu, Farhi, Ryder,
  Pachocki, Chen, and Gao]{yang2022tensor}
G.~Yang, E.~J. Hu, I.~Babuschkin, S.~Sidor, X.~Liu, D.~Farhi, N.~Ryder,
  J.~Pachocki, W.~Chen, and J.~Gao.
\newblock Tensor programs {V}: Tuning large neural networks via zero-shot
  hyperparameter transfer.
\newblock \emph{preprint arXiv:2203.03466}, 2022.

\bibitem[Yang et~al.(2015)Yang, Wilson, Smola, and Song]{SmoSonWilZic15}
Z.~Yang, A.~Wilson, A.~Smola, and L.~Song.
\newblock {A la Carte -- Learning Fast Kernels}.
\newblock In G.~Lebanon and S.~V.~N. Vishwanathan, editors, \emph{Proceedings
  of the Eighteenth International Conference on Artificial Intelligence and
  Statistics}, volume~38 of \emph{Proceedings of Machine Learning Research},
  pages 1098--1106, San Diego, California, USA, 09--12 May 2015. PMLR.

\end{thebibliography}

\end{document}